%% file: arxiv.tex
\documentclass[journal]{IEEEtran}
\usepackage[noadjust]{cite}
\usepackage{booktabs} 
\usepackage[utf8]{inputenc} 
\usepackage[T1]{fontenc}    
\usepackage[bookmarks=false]{hyperref}
\usepackage{url}            
\usepackage{booktabs}       
\usepackage{amsfonts}       
\usepackage{nicefrac}       
\usepackage{microtype}      
\usepackage{amsthm}
\usepackage{times}
\usepackage{epsfig}
\usepackage{graphicx}
\usepackage{amsmath}
\usepackage{amssymb}
\usepackage{extarrows}
\usepackage{wrapfig}
\usepackage{bm}
\usepackage{subcaption}
\usepackage{todonotes}
\usepackage{tcolorbox}
\usepackage{tikz}
\usepackage{tikz-3dplot}
\usepackage{tikzscale}
\usetikzlibrary{arrows,shapes,chains,matrix,positioning}
\usetikzlibrary{scopes,decorations,shadows,backgrounds,fit}
\usetikzlibrary{decorations.pathreplacing,calc,3d,quotes}
\usepackage{pgfmath}
\usepackage{threeparttable}
\usepackage{color}
\usepackage{array}
\usepackage{mdwmath}
\usepackage{mdwtab}
\usepackage{eqparbox}
\usepackage{fixltx2e}
\usepackage{stfloats}
\usepackage{diagbox}
\usepackage{verbatim}
\usepackage{multirow}
\usepackage{mathtools}
\usepackage{breqn}
\usepackage{algorithmic}
\usepackage[ruled, vlined, linesnumbered]{algorithm2e}
\usepackage{listings}
\usepackage{color} 
\usepackage{multicol}
\usepackage{siunitx}
\usepackage{stackengine}
\usepackage{pdfpages}

\definecolor{mygreen}{RGB}{28,172,0} 
\definecolor{mylilas}{RGB}{170,55,241}
\definecolor{overviewred}{RGB}{244,177,131}
\definecolor{overviewblue}{RGB}{157,195,230}
\definecolor{overviewgreen}{RGB}{169,209,142}
\definecolor{revision_color}{RGB}{0,0,255}
\definecolor{algo_comment}{RGB}{79,79,79}

\def\etal{{et al. }}

\def\st{\mathop{\rm subject\ to}}
\def\argmin{\mathop{\rm argmin}\limits}
\def\ourmethod{NSGANetV1}

\theoremstyle{definition}

\theoremstyle{remark}

\def\argmin{\mathop{\rm argmin}\limits}

\def\Minimize{\mathop{\rm minimize}\limits}

\def\st{\mathop{\rm subject\ to}}

\def\*#1{\mathbf{#1}}
\usepackage[symbol]{footmisc}

\def\sota{state-of-the-art}

\usepackage{xcolor}
\definecolor{color_blue}{RGB}{223,234,246}
\definecolor{color_yellow}{RGB}{250,231,163}
\definecolor{color_green}{RGB}{217,255,219}
\definecolor{color_red}{RGB}{245,217,196}

\def\HiLiBlue{\leavevmode\rlap{\hbox to \hsize{\color{color_blue}\leaders\hrule height .8\baselineskip depth .5ex\hfill}}}
\def\HiLiYellow{\leavevmode\rlap{\hbox to \hsize{\color{color_yellow}\leaders\hrule height .8\baselineskip depth .5ex\hfill}}}
\def\HiLiGreen{\leavevmode\rlap{\hbox to \hsize{\color{color_green}\leaders\hrule height .8\baselineskip depth .5ex\hfill}}}
\def\HiLiRed{\leavevmode\rlap{\hbox to \hsize{\color{color_red}\leaders\hrule height .8\baselineskip depth .5ex\hfill}}}

\ifCLASSINFOpdf
\else
\fi

\begin{document}
\title{Multi-Objective Evolutionary Design of Deep Convolutional Neural Networks for Image Classification}

\author{Zhichao Lu,~\IEEEmembership{Student Member,~IEEE,} Ian Whalen,
        Yashesh Dhebar,
        Kalyanmoy Deb,~\IEEEmembership{Fellow,~IEEE,} \\
        Erik Goodman, Wolfgang Banzhaf and Vishnu Naresh Boddeti~\IEEEmembership{Member,~IEEE}
\thanks{The authors are with Michigan State University, East Lansing, MI, 48824 USA, Corresponding author's e-mail: (luzhicha@msu.edu).}
}

\maketitle

\begin{abstract}
Convolutional neural networks (CNNs) are the backbones of deep learning paradigms for numerous vision tasks. Early advancements in CNN architectures are primarily driven by human expertise and by elaborate design {processes}. Recently, neural architecture search was proposed with the aim of automating the network design process and generating task-dependent architectures. While existing approaches have achieved competitive performance in image classification, they are not well suited to problems where the computational budget is limited for two reasons: (1) the obtained architectures are either solely optimized for classification performance, or only for one {deployment scenario}; (2) the search process requires vast computational resources in most approaches. To overcome these limitations, we propose an evolutionary algorithm for searching neural architectures under multiple objectives, such as classification performance and {floating point operations (FLOPs)}. The proposed method addresses the first shortcoming by populating a set of architectures to approximate the entire Pareto frontier through genetic operations that recombine and modify architectural components progressively. Our approach improves computational efficiency by carefully down-scaling the architectures during the search as well as reinforcing the patterns commonly shared among past successful architectures through Bayesian {model learning}. The integration of these two main contributions allows an efficient design of architectures that are competitive and in {most} cases outperform both manually and automatically designed architectures on benchmark image classification datasets: CIFAR, ImageNet and human chest X-ray. The flexibility provided from simultaneously obtaining multiple architecture choices for different compute requirements further differentiates our approach from other methods in the literature. Code is available at \url{https://github.com/mikelzc1990/nsganetv1}.
\end{abstract}


\IEEEpeerreviewmaketitle

\input{introduction.tex}
\input{related-work.tex}
\input{approach.tex}

\input{experiments.tex}

\input{ablation.tex}

\input{application.tex}
\input{conclusion.tex}

\section*{Acknowledgements}
This material is based in part upon work supported by the National Science Foundation under Cooperative Agreement No. DBI-0939454. Any opinions, findings, and conclusions or recommendations expressed in this material are those of the author(s) and do not necessarily reflect the views of the National Science Foundation.

\ifCLASSOPTIONcaptionsoff
  \newpage
\fi

\bibliography{egbib}
\bibliographystyle{IEEEtran}

\appendix
\input{appendix}
\end{document}

%% file: introduction.tex
\section{Introduction\label{sec:intro}}
Deep convolutional neural networks (CNNs) have been overwhelmingly successful in a variety of computer-vision-related tasks like object classification, detection, and segmentation. One of the main driving forces behind this success is the introduction of many CNN architectures, including GoogLeNet \cite{googlenet}, ResNet \cite{resnet}, DenseNet \cite{densenet}, etc., in the context of object classification. Concurrently, architecture designs, such as ShuffleNet \cite{zhang2018shufflenet}, MobileNet \cite{sandler2018mobilenetv2}, LBCNN \cite{lbcnn}, etc., have been developed with the goal of enabling real-world deployment of high-performance models on resource-constrained devices. These developments are the fruits of years of painstaking efforts and human ingenuity.

Neural architecture search (NAS), on the other hand, presents a promising path to alleviate this painful process by posing the design of CNN architectures as an optimization problem. By altering the architectural components in an algorithmic fashion, novel CNNs can be discovered that exhibit improved performance metrics on representative datasets. The huge surge in research and applications of NAS indicates the tremendous academic and industrial interest NAS has attracted, as teams seek to stake out some of this territory. It is now well recognized that designing bespoke neural network architectures for various tasks is one of the most challenging and practically beneficial components of the entire Deep Neural Network (DNN) development process, and is a fundamental step toward automated machine learning.

Early methods for NAS relied on Reinforcement Learning (RL) to navigate and search for architectures with high performance. A major limitation of these approaches \cite{zoph2016,nasnet2018} is the steep computational requirement for the search process itself, often requiring weeks of wall clock time on hundreds of Graphics Processing Unit (GPU) cards. Recent \emph{relaxation}-based methods \cite{liu2018darts,xie2018snas,Dong_2019_CVPR,wu2019fbnet} seek to improve the computational efficiency of NAS approaches by approximating the connectivity between different layers in the CNN architectures by real-valued variables that are learned (optimized) through gradient descent together with the weights. However, such relaxation-based NAS methods suffer from excessive GPU memory requirements during search, resulting in constraints on the size of the search space (e.g., reduced layer operation choices).

In addition to being accurate in prediction, real-world applications demand that NAS methods find network architectures that are also efficient in computation---e.g., have low power consumption in mobile applications and low latency in autonomous driving applications. It has been a common observation that the predictive performance continuously  improves as the complexity (i.e., \# of layers, channels, etc.) of the network architectures increases \cite{resnet,densenet,nasnet2018,tan2019efficientnet}. This alludes to the competing nature of trying to simultaneously maximize predictive performance and minimize network complexity, thereby necessitating multi-objective optimization. Despite recent advances in RL and relaxation-based NAS methods, they are still not readily applicable for multi-objective NAS.

{Among the many different NAS methods being continually proposed, Evolutionary Algorithms (EAs) are getting a plethora of attention,} due to their population-based nature and flexibility in encoding.  They offer a viable alternative to conventional machine learning (ML)-oriented approaches, especially under the scope of multi-objective NAS. An EA, in general, is an iterative process in which individuals in a population are made gradually better by applying variations to selected individuals and/or recombining parts of multiple individuals. Despite the ease of  extending them to handle multiple objectives, most existing EA-based NAS methods \cite{genetic-cnn,real2017largescale,liu2018hierarchical,real2019regularized,ae-cnn,ae-cnn-e2epp} are still single-objective driven. 

In this paper, we present \ourmethod{}, a multi-objective evolutionary algorithm for NAS, {extending on an earlier proof-of-principle method \cite{lu2019nsga},} to address the aforementioned limitations of current approaches. The key contributions followed by the extensions made in this paper are summarized below:
\begin{enumerate}
    
    \item \ourmethod{} populates a set of architectures to approximate the entire Pareto front in one run through customized genetic operations that recombine and modify architectural components  progressively. \ourmethod{} improves computational efficiency by carefully down-scaling the architectures during the search as well as reinforcing the emerging patterns shared among past  successful architectures through a Bayesian Network based distribution estimation operator. Empirically, the obtained architectures, in most cases, outperform both manually and other automatically designed architectures on various datasets.
    
    
    \item {By obtaining a set of architectures in one run, \ourmethod{} allows designers to choose a suitable network \emph{a-posteriori} as opposed to a pre-defined preference weighting of objectives prior to the search. Further post-optimal analysis of the set of non-dominated architectures often reveals valuable design principles, which is another benefit of posing NAS as a multi-objective optimization problem, as is done in \ourmethod{}}.
    
    \item {From an algorithmic perspective, we extend our previous work \cite{lu2019nsga} in a number of ways: (i) an expanded search space to include five more layer operations and one more option that controls the width of the network, (ii) improved encoding, mutation and crossover operators accompanying the modified search space, and (iii) a more thorough lower-level optimization process for weight learning, resulting in better and more reliable performance.}
    
    \item {From an evaluation perspective, we extend our previous work \cite{lu2019nsga} in two different ways: (i) adding three more tasks, including medical imaging, robustness to adversarial attacks, and car key-point estimation; and (ii) evaluating the searched architectures on five new datasets, including, ImageNet, ImageNet-V2, CIFAR-10.1, corrupted CIFAR-10 and corrupted CIFAR-100.}

\end{enumerate}

The remainder of this paper is organized as follows. Section~\ref{sec:related-work} introduces and summarizes related literature. In Section~\ref{sec:approach}, we provide a detailed description of the main components of our approach. We describe the experimental setup to validate our approach along with a discussion of the results in Section~\ref{sec:exp}, followed by further analysis and an application study in Sections~\ref{sec:analysis} and \ref{sec:appl}, respectively. Finally, we conclude with a summary of our findings and comment on possible future directions in Section~\ref{sec:conclusion}.

%% file: related-work.tex
\begin{figure*}[t]
    \centering
    \includegraphics[width=0.8\textwidth{}]{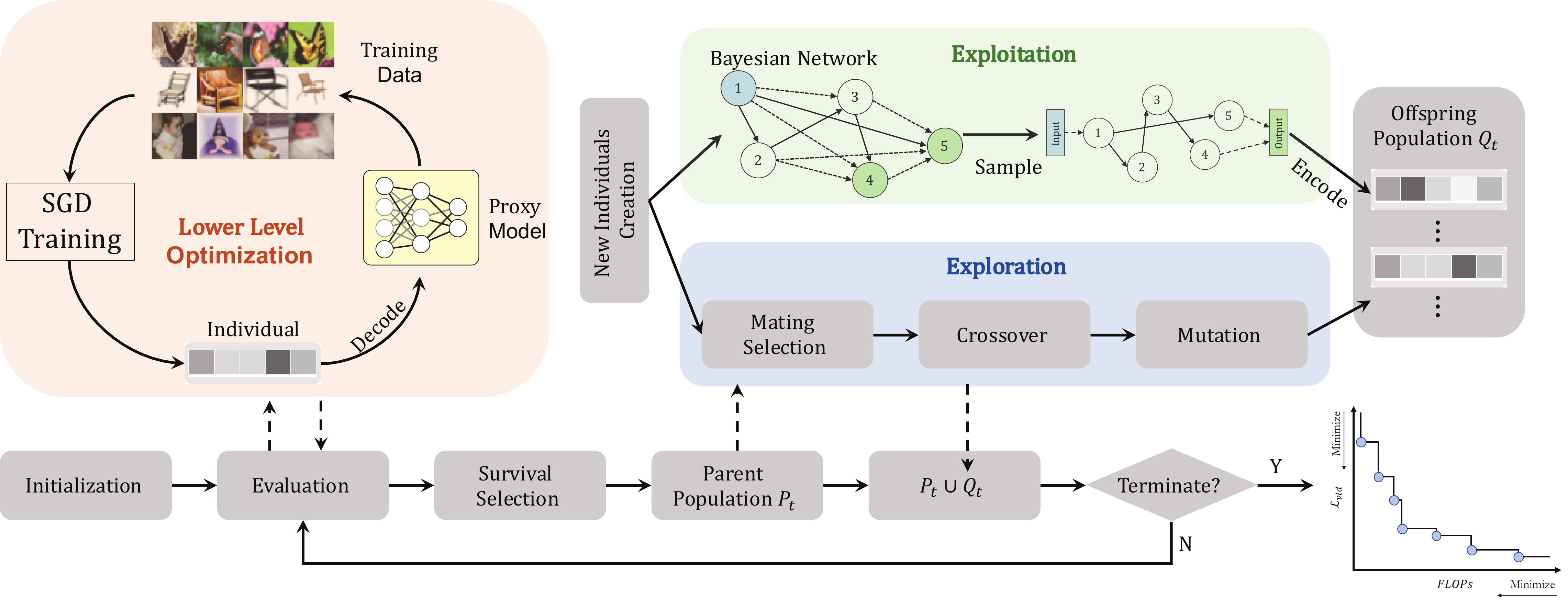}
    \caption{\textbf{Overview:} Given a dataset and objectives, \ourmethod{} designs a set of custom architectures spanning the trading-off front. \ourmethod{} estimate the performance of an architecture through its \emph{proxy model}, optimized by Stochastic Gradient Descent (SGD) in the lower-level. The search proceeds in \emph{exploration} via genetic operations, followed by \emph{exploitation} via distribution estimation. See Algorithm~\ref{algo:framework} for pseudocode and colors are in correspondence.}
    \label{fig:approach_overview}
    \vspace{-1em}
\end{figure*}

\section{Related Work\label{sec:related-work}}
Recent years have witnessed growing interest in NAS. The promise of being able to automatically and efficiently search for task-dependent network architectures is particularly appealing as deep neural networks are widely deployed in diverse applications and computational environments. { Early methods \cite{yao1999evolving,stanley2002evolving} made efforts to simultaneously evolve the topology of neural networks along with weights and hyperparameters. These methods perform competitively with hand-crafted networks on control tasks with shallow fully connected networks. 
In the following, we present studies related to deep convolutional neural networks for image classification.  Readers are referred to the supplementary materials for a more detailed review of the topic.
}

\vspace{3pt}
\noindent\textbf{Evolutionary NAS:} Designing neural networks through evolution has been a topic of interest for a long time. Recent evolutionary approaches focus on evolving solely the topology while leaving the learning of weights to gradient descent algorithms, and using hyper-parameter settings that are manually tuned. Xie and Yuille's work of Genetic CNN \cite{genetic-cnn} is one of the early studies that shows the promise of using EAs for NAS. Real et al. \cite{real2017largescale} introduce perhaps the first truly large scale application of a simple EA to NAS. The extension of this method presented in \cite{real2019regularized}, called AmoebaNet, provides the first large scale comparison of EA and RL methods. Their EA, using an age-based selection similar to \cite{alphs}, has demonstrated faster convergence to an accurate network when compared to RL and random search. Concurrently, Liu et al. \cite{liu2018hierarchical} evolve a hierarchical representation that allows non-modular layer structures to emerge. Despite the impressive improvements achieved on various datasets, these EA methods are extremely computationally inefficient, e.g., one run of the regularized evolution method \cite{real2019regularized} takes one week on 450 GPU cards.

Concurrently, another streamlining of EA methods for use in budgeted NAS has emerged. Suganuma et al. \cite{cgp-cnn} use Cartesian genetic programming to assemble an architecture from existing modular blocks (e.g., Residual blocks). Sun et al. in \cite{ae-cnn-e2epp} use a random forest as an offline surrogate model to predict the performance of architectures, partially eliminating the lower-level optimization via gradient descent. The reported results yield 3$\times$ savings in wall clock time with similar classification performance when compared to their previous works \cite{ae-cnn,cnn-ga}. However, results reported from these budgeted EA methods are far from state-of-the-art and only demonstrated on small-scale datasets---i.e., CIFAR-10 and CIFAR-100.

\vspace{3pt}
\noindent\textbf{Multi-objective NAS:} {In this work, the term \emph{multi-objective NAS} refers to methods that simultaneously approximate the entire set of efficient trade-off architectures in one run \cite{jin2008pareto,zhu2019multi}.} Kim et al. \cite{kim2017nemo} presented NEMO, one of the earliest evolutionary multi-objective approaches to evolve CNN architectures. NEMO uses NSGA-II \cite{deb2002fast} to maximize classification performance and inference time of a network and searches over the space of the number of output channels from each layer within a restricted space of seven different architectures. DPP-Net \cite{dong2018dpp}, an extension from \cite{liu2018progressive}, progressively expands networks from simple structures and only trains the top-K (based on Pareto-optimality) networks that are predicted to be promising by a surrogate model. Elsken et al. \cite{elsken2018efficient} present the LEMONADE method, which is formulated to develop networks with high predictive performance and lower resource constraints. LEMONADE reduces compute requirements through custom-designed approximate network morphisms \cite{wei2016morhpisms}, which allow newly generated networks to share parameters with their forerunners, obviating the need to train new networks from scratch. However, LEMONADE still requires nearly 100 GPU-days to search on the CIFAR datasets \cite{cifar10}.

\vspace{3pt}
{
\noindent\textbf{Search Efficiency:} The main computation bottleneck of NAS resides in the lower-level optimization of learning the weights for evaluating the performance of architectures. One such evaluation typically requires hours to finish. To improve the practical utility of search under a constrained computational budget, NAS methods commonly advocate for substitute measurements without a full-blown lower-level optimization. A widely-used approach proceeds as follows: it reduces the depth (number of layers) and the width (number of channels) of the intended architecture to create a small-scale network---i.e., a \emph{proxy model}. Proxy models require an order of magnitude less computation time (typically, minutes) to perform lower-level optimization, and the performance of proxy models is then used as surrogate measurements to guide the search. However, most existing NAS work \cite{nasnet2018,liu2018hierarchical,zhong2017blockqnn,real2019regularized,liu2018progressive} follows simple heuristics to construct the proxy model, resulting in low correlation in prediction. For instance, NASNet \cite{nasnet2018} has an additional re-ranking stage that trains the top 250 architectures for 300 epochs each (takes more than a year on a single GPU card) before picking the best one, and the reported NASNet-A model was originally ranked 70th among the top 250 according to the performance measured at proxy model scale. Similarly, AmoebaNet \cite{real2019regularized} relies on evaluation of duplicate architectures to gauge representative performance, leading to 27K models being evaluated during search.

In this work, we focus on both the efficiency and the reliability aspects of the proxy model; through a series of systematic studies in a controlled setting, we empirically establish the trade-off between the correlation of proxy performance to true performance and the speed-up in estimation. We then implement a suitable setting that is specific to our search space and dataset.}

%% file: approach.tex
\section{Proposed Approach\label{sec:approach}}
{Practical applications of NAS can rarely be considered from the point of view of a single objective of maximizing performance; rather, they must be evaluated from at least one additional, conflicting objective that is specific to the deployment scenario. In this work, we approach the problem of designing high-performance architectures with diverse complexities for different deployment scenarios as a multi-objective bilevel optimization problem}\footnote{See the supplement, Section~III, for more about bilevel optimization.} \cite{eichfelder2010multiobjective}. We mathematically formulate the problem as,
{
\begin{equation}
\begin{aligned}
\Minimize & \hspace{3mm} \bm{F}(\bm{x}) = \big(f_1(\bm{x}; \bm{w}^*(\bm{x})), f_2(\bm{x})\big)^T, \\
\st  & \hspace{3mm} \bm{w}^*(\bm{x}) \in \argmin~\mathcal{L}(\bm{w};\bm{x}), \\
     & \hspace{3mm} \bm{x} \in \mathbf{\Omega}_{x}, \hspace{3mm} \bm{w} \in \mathbf{\Omega}_{w},
\end{aligned}
\label{def:bi_obj_nas}
\end{equation}

\noindent where $\bm{\Omega}_x = \Pi_{i=1}^{n}[a_i, b_i] \subseteq \mathbb{Z}^n$ is the architecture decision space, where $a_i$, $b_i$ are the lower and upper bounds, $\bm{x} = (x_1, \ldots, x_n)^T \in \bm{\Omega}_x$ is a candidate architecture, and the lower-level variable $\bm{w} \in \bm{\Omega}_w$ denotes its associated weights. The upper-level objective vector $\bm{F}$ comprises of the classification error ($f_1$) on the validation data $\mathcal{D}_{vld}$, and the complexity ($f_2$) of the network architecture. The lower level objective $\mathcal{L}(\bm{w};\bm{x})$ is the cross-entropy loss on the training data $\mathcal{D}_{trn}$.}


\begin{algorithm}[t]
\SetAlgoLined
\SetKwInOut{Input}{Input}
\SetKwInOut{Output}{Output}
\SetKwFor{For}{for}{do}{end for}
\footnotesize
{
\Input{Complexity objective $\tilde{f}$ (see Eq.~(\ref{def:bi_obj_nas})), Max. number of generations $G$, Population size $K$, Crossover probability $p_{c}$, Mutation probability $p_{m}$, {The starting generation of exploitation $\tau$}.}
    $g$ $\leftarrow$ 0 // initialize a generation counter.\\
    {$\rho \leftarrow 1$ // initialize the control parameter for \emph{exploration}.}\\
    $\mathcal{A}$ $\leftarrow$ initialize an empty archive to keep track of evaluated archs.\\
    $P \leftarrow$ initialize the parent population by uniform sampling.\\
    // compute accuracy through lower-level optimization in Algo.~\ref{algo:back-prop}.\\
    \HiLiRed $f \leftarrow$ \emph{Evaluate}$(P)$ \\
    // calculate domination rank and crowding distance.\\
    $[F_1, F_2, \ldots] \leftarrow$ \emph{NondominatedSort}$(f, \tilde{f}(P))$ \\
    $dist \leftarrow$ \emph{CrowdingDistance}$(F_1, F_2, \ldots)$\\
    \While{$g < G$}{
        $k$ $\leftarrow$ 0 // initialize an individual counter.\\
        $Q \leftarrow \emptyset$ // offspring population. \\
        \While{$k < K$}{
            {// one offspring is created in each iteration k.} \\
            \uIf{$rand()$ $< \rho$}{
                {// choose two parents for mating.}\\
                \HiLiBlue $p \leftarrow$ \emph{BinaryTournamentSelection}$(P, [F_1, F_2, \ldots], dist)$ \\
                \HiLiBlue $q \leftarrow$ \emph{Crossover}($p, p_{c}$) \\
                \HiLiBlue $q \leftarrow$ \emph{Mutation}($q, p_{m}$) \\
            }
            \Else{
                // estimate the distribution of the Pareto set. \\
                \HiLiGreen $BN \leftarrow$ construct a Bayesian Network from $\mathcal{A}$. \\
                \HiLiGreen $q \leftarrow$ {sample an offspring from $BN$.} \\
            }
            $Q \leftarrow Q \cup q$; $k$ $\leftarrow$ $k + 1$ \\
        }
        \HiLiRed $f^{\prime} \leftarrow$ \emph{Evaluate}$(Q)$ // see line 5.\\
        $[F_1, F_2, \ldots] \leftarrow$ \emph{NondominatedSort}$(f \cup f^{\prime}, \tilde{f}(P) \cup \tilde{f}(Q))$ \\
        $dist \leftarrow$ \emph{CrowdingDistance}$(F_1, F_2, \ldots)$\\
        {// survive the top-$K$ archs to next generation following the environmental selection procedures outlined in \cite{deb2002fast}.} \\
        $P \leftarrow$ \emph{Selection}$(P \cup Q, [F_1, F_2, \ldots], dist, K)$\\
        $g$ $\leftarrow$ $g + 1$; $\mathcal{A} \leftarrow \mathcal{A} \cup Q$ \\
        {
        \uIf{$g = \tau$}{$\rho$ $\leftarrow$ 0.75 // assign 25\% of the offspring to be created by BN.}
        \uElseIf{$g > \tau$}{update $\rho$ according to Eq.~(\ref{def:adapt_bn}).}
        \Else{$\rho$ $\leftarrow$ 1 // remain unchanged from initial value.}
        }
    }
\textbf{Return} parent population $P$.
\caption{General framework of \ourmethod{}\label{algo:framework}}}
\end{algorithm}

Our proposed algorithm, \ourmethod{}, is an iterative process in which initial architectures are made gradually better as a group, called a \emph{population}. In every iteration, a group of \emph{offspring} (i.e., new architectures) is created by applying variations through crossover and mutation to the more promising of the architectures already found, also known as \emph{parents}, from the population. Every  member in the population (including both parents and offspring) compete for survival and reproduction (becoming a parent) in each iteration. The initial population may be generated randomly or guided by prior-knowledge, i.e., seeding the past successful architectures directly into the initial population. Subsequent to initialization, \ourmethod{} conducts the search in two sequential stages: (i) \emph{exploration}, with the goal of discovering diverse ways to construct architectures, and (ii) \emph{exploitation} that reinforces the emerging patterns commonly shared among the architectures successful during exploration.
A set of architectures representing efficient trade-offs between network performance and complexity is obtained at the end of evolution, through genetic operators and a Bayesian-model-based learning procedure. A flowchart and a pseudocode outlining the overall approach are shown in Fig.~\ref{fig:approach_overview} and Algorithm~\ref{algo:framework}, respectively. In the remainder of this section, we provide a detailed description of the aforementioned components in Sections \ref{sec:search_space} - \ref{sec:reproduction}.

\subsection{Search Space and Encoding}\label{sec:search_space}
The search for optimal network architectures can be performed over many different search spaces. The generality of the chosen search space has a major influence on the quality of results that are even possible. Most existing evolutionary NAS approaches \cite{genetic-cnn,ae-cnn-e2epp,cgp-cnn,elsken2018efficient} search only one aspect of the architecture space---e.g., the connections and/or hyper-parameters. In contrast, \ourmethod{} searches over both operations and connections---the search space is thus more comprehensive, including most of the previous successful architectures designed both by human experts and algorithmically.

Modern CNN architectures are often composed of an outer structure (\emph{network-level}) design where the width (i.e., \# of channels), the depth (i.e., \# of layers) and the spatial resolution changes (i.e., locations of pooling layers) are decided; and an inner structure (\emph{block-level}) design where the layer-wise connections and computations are specified, e.g., Inception block \cite{googlenet}, ResNet block \cite{resnet}, and DenseNet block \cite{densenet}, etc. As seen in the CNN literature, the network-level decisions are mostly hand-tuned based on meta-heuristics from prior knowledge and the task at hand, as is the case in this work. For block-level design, we adopt the one used in \cite{nasnet2018,liu2018progressive,real2019regularized,liu2018darts} to be consistent with previous work.

\begin{figure*}[t]
	\centering
	\begin{subfigure}[t]{.22\textwidth}
		\centering
		\includegraphics[width=0.95\textwidth]{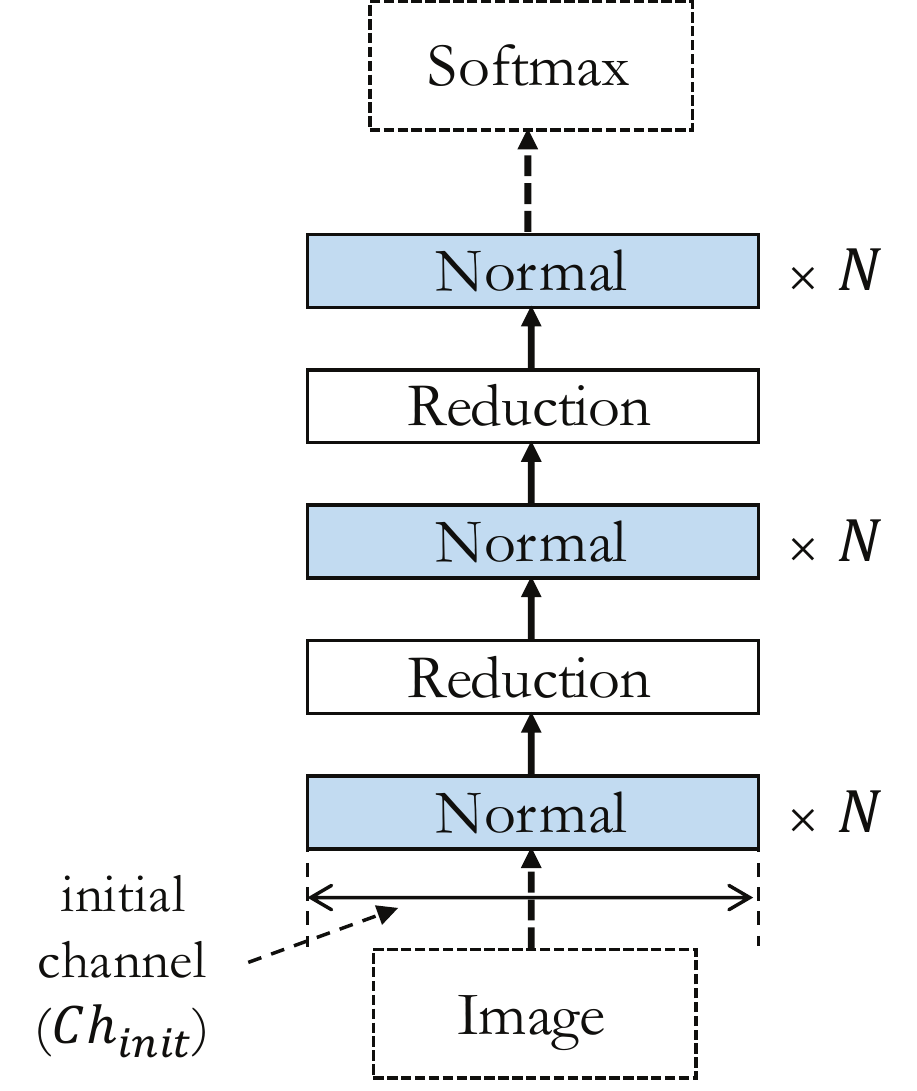}
		\caption{\label{fig:search_space_structure}}
	\end{subfigure}
	\begin{subfigure}[t]{.20\textwidth}
		\centering
		\includegraphics[width=0.95\textwidth]{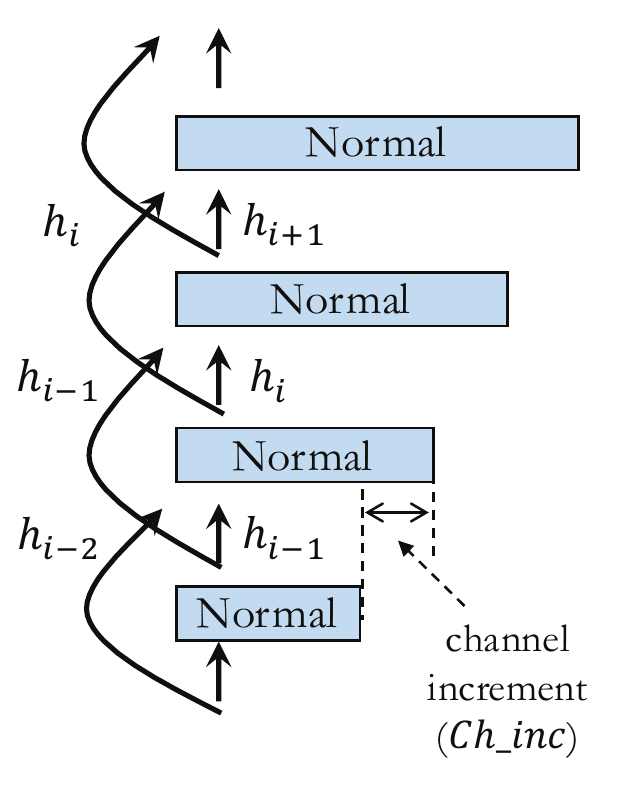}
		\caption{\label{fig:search_space_connection}}
	\end{subfigure}
	\begin{subfigure}[t]{.33\textwidth}
		\centering
		\includegraphics[width=\textwidth]{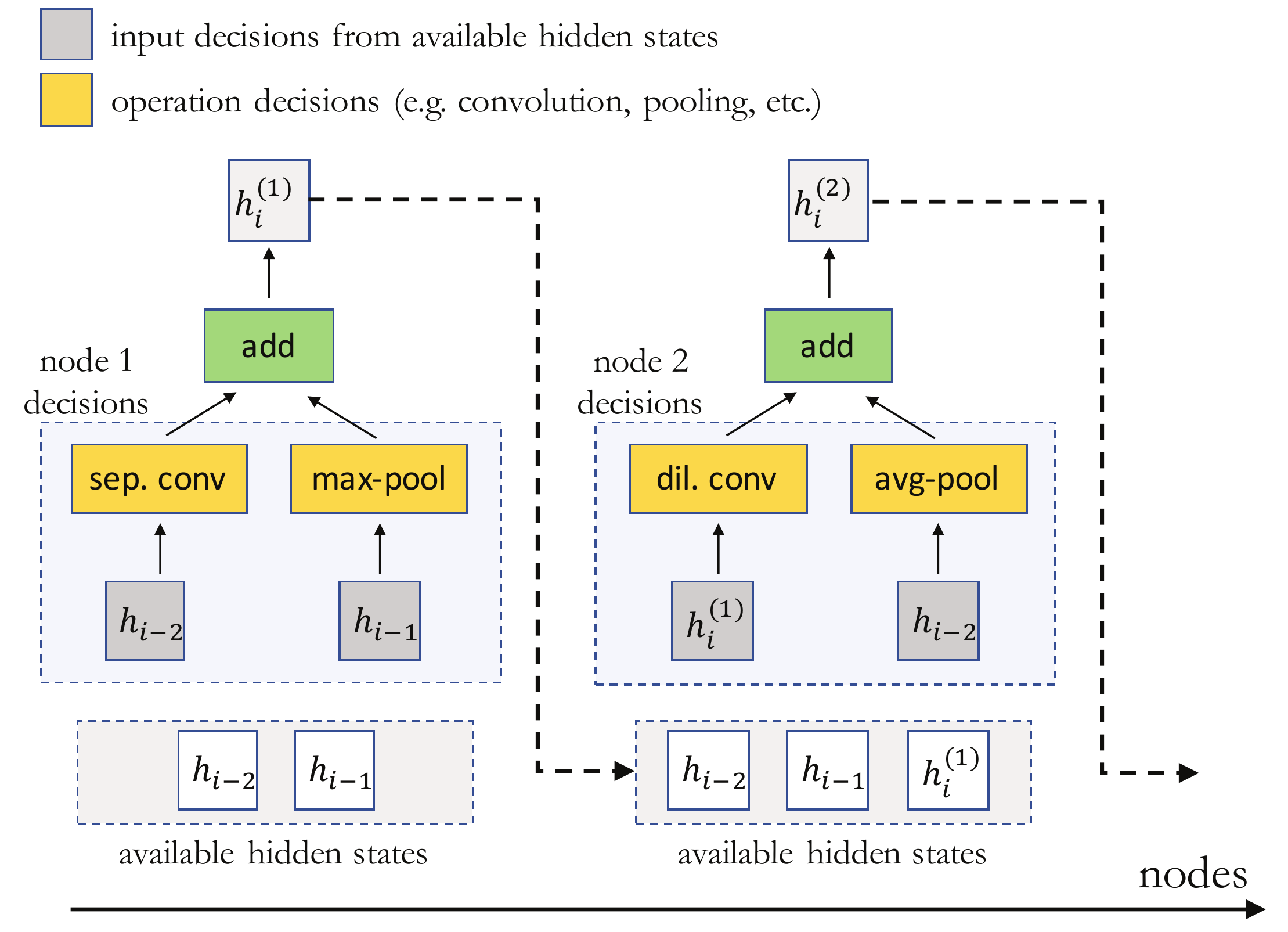}
		\caption{\label{fig:search_space_block}}
	\end{subfigure}
	\begin{subfigure}[t]{.23\textwidth}
		\centering
		\includegraphics[width=0.65\textwidth]{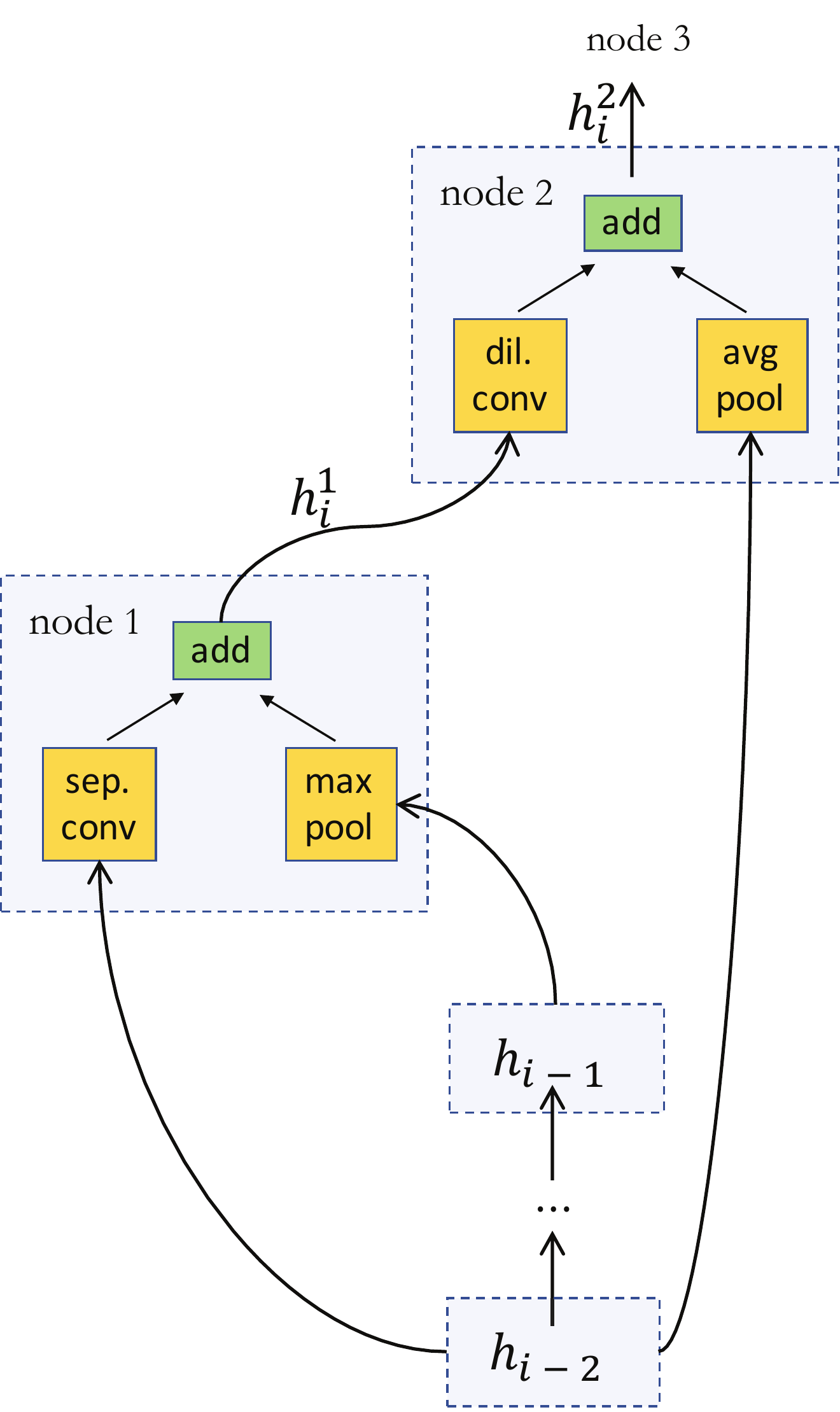}
		\caption{\label{fig:search_space_block_phenome}}
	\end{subfigure}
	\caption{Schematic of the \ourmethod{} search space motivated from \cite{nasnet2018}: (a) An architecture is composed of stacked blocks. (b) The number of channels in each block is gradually increased with depth of the network. (c) Each block is composed of five nodes, where each node is a two-branched computation applied to outputs from either previous blocks or previous nodes within the same block. (d) A graphical visualization of (c).\label{fig:search_space}}
	\vspace{-1em}
\end{figure*}

A \emph{block} is a small convolutional module, typically repeated multiple times to form the entire neural network. To construct scalable architectures for images of different resolutions, we use two types of blocks to process intermediate information: (1) the \emph{Normal} block, a block type that returns information of the same spatial resolution; and (2) the \emph{Reduction} block,  another block type that returns information with spatial resolution halved by using a stride of two. See Fig.~\ref{fig:search_space_structure} for a pictorial illustration.

We use directed acyclic graphs (DAGs) consisting of five nodes to construct both types of blocks (a Reduction block uses a stride of two). Each \emph{node} is a two-branched structure, mapping two inputs to one output. For each node in block $i$, we need to pick two inputs from among the output of the previous block $h_{i-1}$, the output of the previous-previous block $h_{i-2}$, and the set of hidden states created in any previous nodes of block $i$. For pairs of inputs chosen, we choose a computation operation from among the following options, collected based on their prevalence in the CNN literature:
\vspace{-2mm}
\begin{multicols}{2}
\footnotesize{
\begin{itemize}
    \item identity
    \item 3x3 max pooling
    \item 3x3 average pooling
    \item squeeze-and-excitation \cite{hu2018squeeze}
    \item 3x3 local binary conv \cite{lbcnn}
    \item 5x5 local binary conv \cite{lbcnn}
    \item 3x3 dilated convolution
    \item 5x5 dilated convolution
    \item 3x3 depthwise-separable conv
    \item 5x5 depthwise-separable conv
    \item 7x7 depthwise-separable conv
    \item 1x7 then 7x1 convolution
\end{itemize}}
\end{multicols}
\vspace{-2mm}

The results computed from both branches are then added together to create a new hidden state, which is available for subsequent nodes in the same block. See Fig.~\ref{fig:search_space_connection}-\ref{fig:search_space_block_phenome} for pictorial illustrations. {The search space we consider in this paper is an expanded version of the \emph{micro search space} used in our previous work \cite{lu2019nsga}. Specifically, the current search space (i) search for a factor that gradually increments the channel size of each block with depth (see Fig.~\ref{fig:search_space_connection}) as opposed to sharply doubling the channel size when down-sampling. (ii) considers an expanded set of primitive operations to include both more recent advanced layer primitives such as squeeze-and-excitation \cite{hu2018squeeze} and more parameter-efficient layer primitives like local binary convolution \cite{lbcnn}}.

With the above-mentioned search space, there are in total 20 decisions to constitute a block structure---i.e., choose two pairs of input and operation for each node, and repeat for five nodes. The resulting number of combinations for a block structure is:
$$\mathcal{B} = \big((n + 1)!\big)^2 \cdot (n\_ops)^{2n}$$
where $n$ denotes the number of nodes, $n\_ops$ denotes the number of considered operations. Therefore, with one Normal block and one Reduction block with five nodes in each, the overall size of the encoded search space is approximately \num{E+33}.

\vspace{-2mm}
\subsection{Performance Estimation Strategy}\label{sec:performance_estimation}
To guide \ourmethod{} towards finding more accurate and efficient architectures, we consider two metrics as objectives, namely, classification accuracy and architecture complexity. Assessing the classification accuracy of an architecture during search requires another optimization to first identify the optimal values of the associated weights via Stochastic Gradient Descent (SGD; Algorithm~\ref{algo:back-prop}). Even though there exist well-established gradient descent algorithms to efficiently solve this optimization, repeatedly executing such an algorithm for every candidate architecture renders the overall process computationally very prohibitive. {Therefore, to overcome this computational bottleneck, we carefully (using a series of ablation studies) down-scale the architectures to create their proxy models \cite{nasnet2018,real2019regularized}, which can be optimized efficiently in the lower-level through SGD. Their performances become surrogate measurements to select architectures during search. Details are provided in Section~\ref{sec:hyper-param-analysis}.}

\begin{algorithm}[t]
\SetAlgoLined
\SetKwInOut{Input}{Input}
\SetKwInOut{Output}{Output}
\SetKwFor{For}{for}{do}{end for}
\footnotesize
\Input{The architecture $\alpha$, training data $\mathcal{D}_{trn}$, \\
validation data $\mathcal{D}_{vld}$, number of epochs $T$, \\
weight decay $\lambda$, initial learning rate $\eta_{max}$.}
 $\omega$ $\leftarrow$ Randomly initialize the weights in $\alpha$\;
 $t$ $\leftarrow$ 0\;
 \While{$t < T$}{
  $\eta$ $\leftarrow$ $\frac{1}{2}\eta_{max}\big(1 + \cos{\left(\frac{t}{T}\pi\right)}\big)$\;
  \For{each data-batch in $\mathcal{D}_{trn}$}{
    $\mathcal{L}$ $\leftarrow$ Cross-entropy loss on the \emph{data-batch}\;
    $\nabla\omega$ $\leftarrow$ Compute the gradient by $\partial \mathcal{L} / \partial \omega$\;
    $\omega$ $\leftarrow$ $(1 - \lambda)\omega - \eta\nabla\omega$\;
  }
  $t$ $\leftarrow$ $t + 1$\;
 }
 $acc$ $\leftarrow$ Compute accuracy of $\alpha(\omega)$ on $\mathcal{D}_{vld}$\;
\textbf{Return} the classification accuracy $acc$.
 \caption{Performance Evaluation of a CNN\label{algo:back-prop}}
\end{algorithm}


A number of metrics can serve as proxies for complexity, including: the number of active nodes, number of active connections between the nodes, number of parameters, inference time and {number of floating-point operations (FLOPs) needed to execute the forward pass of a given architecture.} Our initial experiments considered each of these metrics in turn. We concluded from extensive experimentation that inference time cannot be estimated reliably due to differences and inconsistencies in the computing environment, GPU manufacturer, ambient temperature, etc. {Similarly, the number of parameters, active connections or active nodes only relate to one aspect of the complexity.} In contrast, we found an estimate of FLOPs to be a more accurate and reliable proxy for network complexity. Therefore, classification accuracy and FLOPs serve as our choice of twin objectives to be traded off for selecting architectures. To simultaneously compare and select architectures based on these two objectives, we use the non-dominated ranking and the ``crowded-ness" concepts proposed in \cite{deb2002fast}.

\vspace{-2mm}
{
\subsection{Creation of New Generation}\label{sec:reproduction}
\noindent\textbf{Exploration:}} Given a population of architectures, parents are selected from the population with a fitness bias. This choice is dictated by two observations, (1) offspring created around better parents are expected to have higher fitness on average than those created around worse parents, with the assumption of some level of gradualism in the solution space; (2) occasionally (although not usually), offspring perform better than their parents, through inheriting useful traits from both parents. Because of this, one might demand that the best architecture in the population should always be chosen as one of the parents. However, the deterministic and greedy nature of that approach would likely lead to premature convergence due to loss of diversity in the population \cite{leung1997degree}. {To address this problem, we use binary tournament selection \cite{miller1995genetic} to promote parent architectures in a stochastic fashion. At each iteration, binary tournament selection randomly picks two architectures from the population, then the one favored by the {multi-objective} selection criterion described in Section~\ref{sec:performance_estimation} becomes one of the parents. This process is repeated to select a second parent architecture; the two parent architectures then undergo a crossover operation.}

\begin{figure}[!tbh]
    \centering
    \includegraphics[width=0.4\textwidth{}]{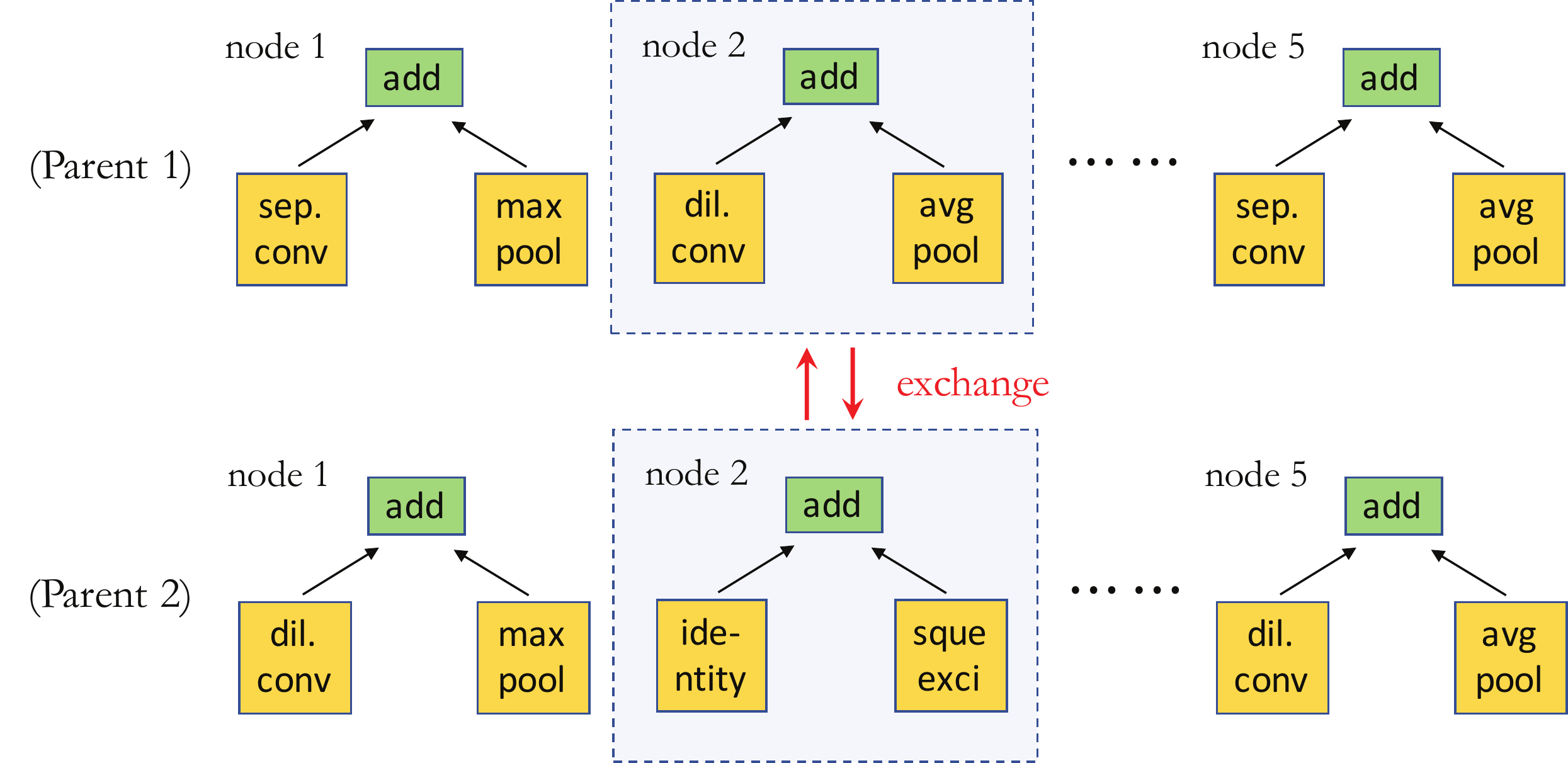}
    \caption{Illustration of node-level crossover.}
    \label{fig:crossover}
    \vspace{-1em}
\end{figure}

In \ourmethod{}, we use two types of crossover (with equal probability of being chosen) to efficiently exchange sub-structures between two parent architectures. {The first type is at the block level, in which the offspring architectures are created by recombining the Normal block from the first parent with the Reduction block from the other parent and vice versa}. The second type is at the node level, where a node from one parent is randomly chosen and exchanged with another node at the same position from the other parent. We apply the node-level crossover to both Normal and Reduction blocks. Fig.~\ref{fig:crossover} illustrates an example of node-level crossover. {Note that two offspring architectures are generated after each crossover operation, and only one of them (randomly chosen) is added to the offspring population. In each generation, an offspring population of the same size as the parent population is generated.}

\begin{figure}[!htb]
    \centering
    \includegraphics[width=0.48\textwidth{}]{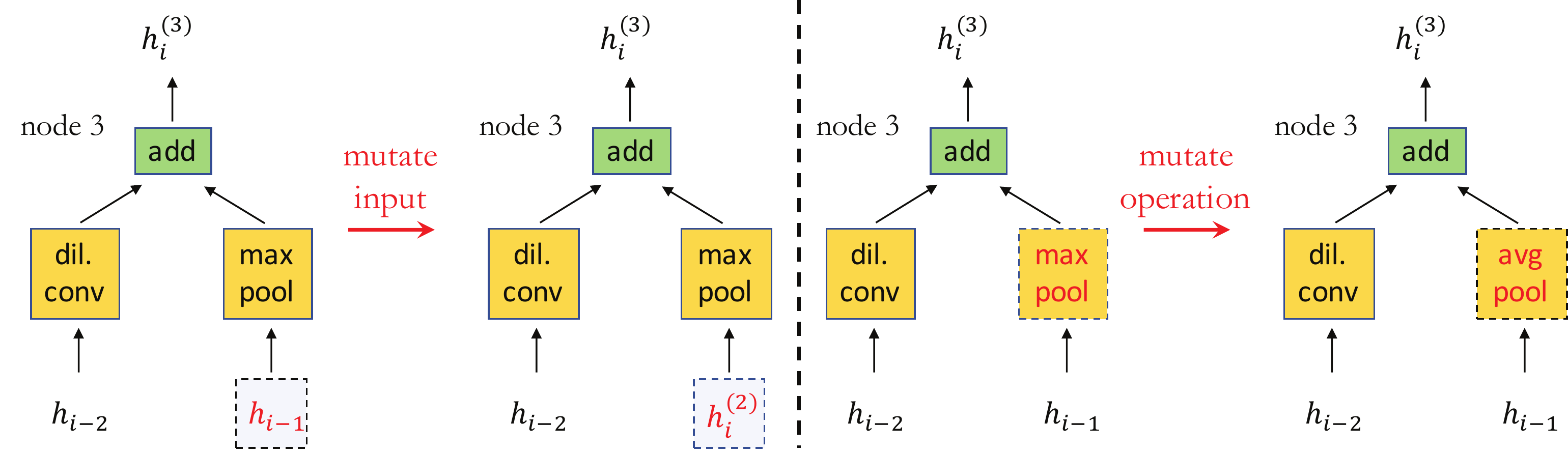}
    \caption{Input and Operation Mutation: Dashed line boxes with red color highlight the mutation. $h_{i-2}$ and $h_{i-1}$ are outputs from previous-previous and previous blocks, respectively. $h_{i}^{(3)}$ indicates output from node 3 of the current block.}
    \label{fig:mutation}
    \vspace{-1em}
\end{figure}

To enhance the diversity of the population and the ability to escape from local attractors, we use a discretized version of the polynomial mutation (PM) operator \cite{deb1995simulated} subsequent to crossover. We allow mutation to be applied on both the input hidden states and the choice of operations. Figure~\ref{fig:mutation} shows an example of each type of mutation using the parent-centric PM operator, in which the offspring are intentionally created around the parents in the decision space.
In association with PM, we sort our discrete encoding of input hidden states chronologically and choice of operations in ascending order of computational complexity. In the context of neural architecture, this step results in the mutated input hidden states in offspring architectures to more likely be close to the input hidden states in parent architectures in a chronological manner. For example, $h_i^{(2)}$ is more likely to be mutated to $h_i^{(1)}$ than to $h_{i-2}$ by PM. A similar logic is applied in case of mutation on layer operations.

\begin{figure}[!tbh]
    \centering
    \includegraphics[width=0.48\textwidth{}]{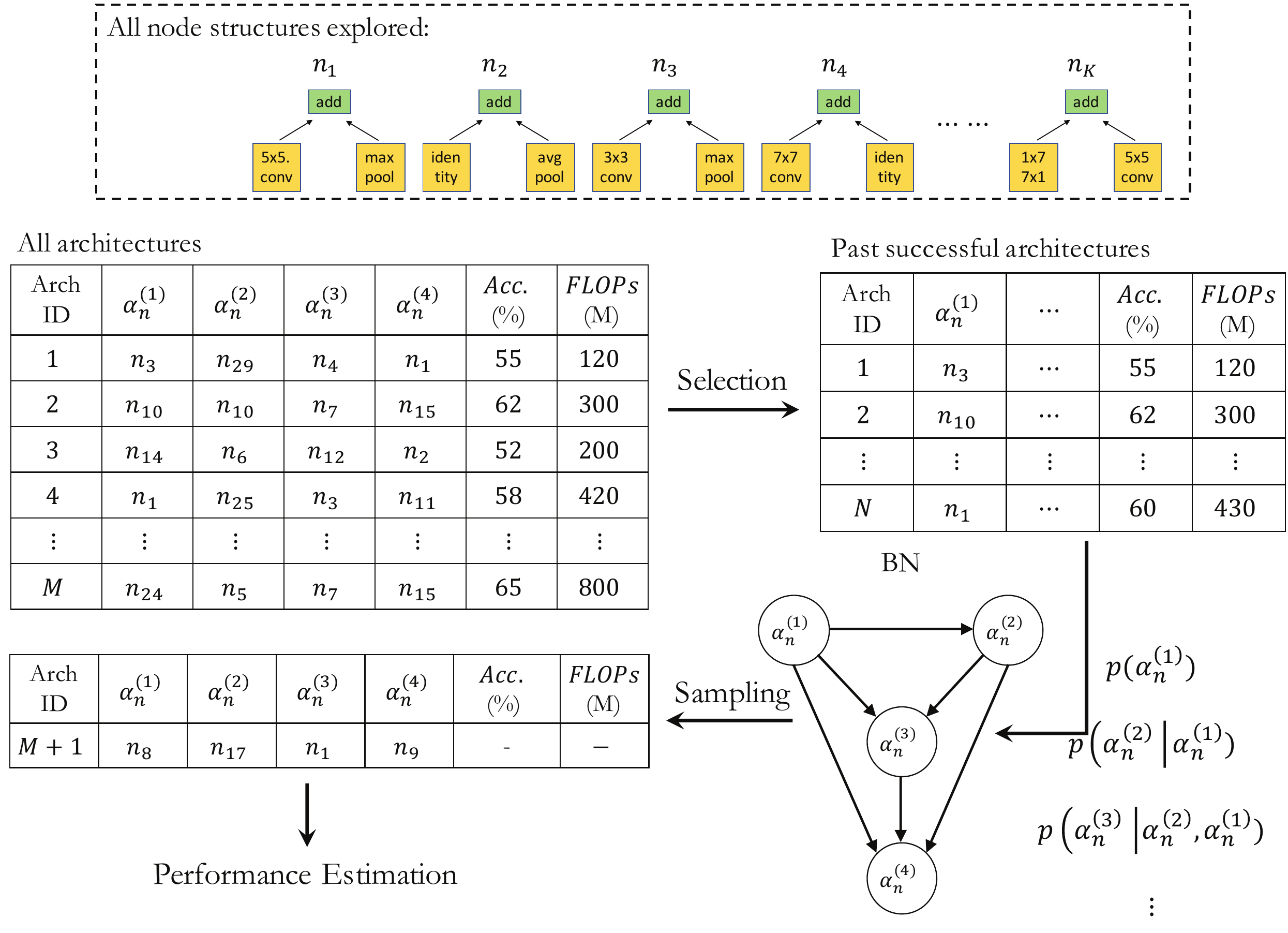}
    \caption{Illustrative example of BN-based exploitation step in \ourmethod{}: given past successful architectures, we construct a BN relating the dependencies between the four nodes inside the Normal block. A new architecture is then sampled from this BN and proceeds forward for performance estimation.}
    \label{fig:BN}
\end{figure}

\vspace{3pt}
\noindent\textbf{Exploitation:} After a sufficient number of architectures has been explored (consuming 2/3 of the total computational budget; {i.e., $\tau$ in Algorithm~\ref{algo:framework}}), we start to enhance the exploitation aspect of the search. The key idea is to reinforce and reuse the patterns commonly shared among past successful architectures. {We use the Bayesian Network (BN) \cite{pelikan1999boa} as the probabilistic model to estimate the distribution of the Pareto set (of architectures).}
{In the context of our search space and encoding, this translates to learning the correlations among the operations and connections of nodes within a block. Our exploitation step uses a subset (top-100 architectures selected based on domination rank and crowding distance \cite{deb2002fast}) of the past evaluated architectures to guide the final part of the search.} More specifically, say we are designing a Normal block with three nodes, namely ${\alpha_n^{(1)}}$, ${\alpha_n^{(2)}}$, and ${\alpha_n^{(3)}}$. We would like to know the relationship among these three nodes. For this purpose, we construct a BN relating these variables, modeling the probability of Normal blocks beginning with a particular node ${\alpha_n^{(1)}}$, the probability that ${\alpha_n^{(2)}}$ follows ${\alpha_n^{(1)}}$, and the probability that ${\alpha_n^{(3)}}$ follows ${\alpha_n^{(2)}}$ and ${\alpha_n^{(1)}}$. In other words, we estimate the conditional distributions $p\left({\alpha_n^{(1)}}\right)$, $p\left({\alpha_n^{(2)}} | {\alpha_n^{(1)}}\right)$, and $p\left({\alpha_n^{(3)}} | {\alpha_n^{(2)}}, {\alpha_n^{(1)}}\right)$ by using the population history, and update these estimates during the exploitation process. New offspring architectures are created by sampling from this BN. A pictorial illustration of this process is provided in Fig.~\ref{fig:BN}. This BN-based exploitation strategy is used in addition to the genetic operators, where we initially (i.e., at the beginning of exploitation) assign 25\% of the offspring {(line 34 in Algorithm~\ref{algo:framework})} to be created by BN and we update this probability adaptively {(line 36 in Algorithm~\ref{algo:framework})}. {To be more specific, we calculate the probabilities of using genetic operators and sampling from the BN model at generation $t$ based on the survival rates of offspring created using them in the previous generation, following the softmax function:
\begin{equation}
\label{def:adapt_bn}
\rho_t^{(i)} = \frac{\mbox{exp}(s_{t-1}^{(i)})}{\sum_{i=1}^2\mbox{exp}(s_{t-1}^{(i)})}
\end{equation}
where $\rho_t^{(i)}$ are the probabilities of using genetic operators ($i = 1$) and sampling from the learned BN model ($i = 2$); and $s_{t-1}^{(i)}$ are the survival rates of the offspring created by genetic operators ($i = 1$) and the learned BN model ($i = 2$) at the previous generation $t-1$. {Note that $\rho_t^{(i=1)}$ corresponds to $\rho$ in Algorithm~\ref{algo:framework}}.}




%% file: experiments.tex
\section{Experimental Setup and Results\label{sec:exp}}
In this section, we will evaluate the efficacy of \ourmethod{} on multiple benchmark image classification datasets.

\vspace{-2mm}
\subsection{Baselines}
To demonstrate the effectiveness of the proposed algorithm, we compare the non-dominated architectures achieved at the conclusion of \ourmethod{}'s evolution with architectures reported by various peer methods published in top-tier venues. The chosen peer methods can be broadly categorized into three groups: architectures manually designed by human experts, non-EA- (mainly RL or relaxation)-based, and EA-based. Human engineered architectures include ResNet \cite{resnet}, ResNeXt \cite{xie2017aggregated}, and DenseNet \cite{densenet}, etc. The second and third groups range from earlier methods \cite{zoph2016,real2017largescale,genetic-cnn} that are oriented towards ``proof-of-concept" for NAS, to more recent methods \cite{nasnet2018,real2019regularized,liu2018darts,cai2018proxylessnas}, many of which improve state-of-the-art results on various computer vision benchmarks at the time they were published. The effectiveness of the different architectures is judged on both classification accuracy and computational complexity. For comparison on classification accuracy, three widely used natural object classification benchmark datasets are considered, namely, CIFAR-10, CIFAR-100 and ImageNet. More details and a gallery of examples from these three datasets are provided in Fig.~2 in supplementary materials under Section~V.

\vspace{-2mm}
\subsection{Implementation Details\label{sec:implementation_detail}}
Motivated by efficiency and practicality considerations most existing NAS methods, including \cite{nasnet2018,real2019regularized,pmlr-v80-pham18a,ae-cnn},
carry out the search process on the CIFAR-10 dataset. However, as we demonstrate through ablation studies in  Section~\ref{sec:hyper-param-analysis} the CIFAR-100 provides a more reliable measure of an architecture's efficacy in comparison to CIFAR-10. Based on this observation, in contrast to existing approaches, we use the more challenging CIFAR-100 dataset for the search process. Furthermore, we split the original CIFAR-100 training set (80\%-20\%) to create a training and validation set to prevent over-fitting to the training set and improve the generalizability. We emphasize that the original testing set is \emph{never} used to guide the selection of architectures in any form during the search.

The search itself is repeated five times with different initial random seeds. We select and report the performance of the median run as measured by hypervolume (HV). Such a procedure ensures the reproducibility of our NAS experiments and mitigates the concerns that have arisen in recent NAS studies \cite{li2019random,xie2019exploring}. We use the standard SGD algorithm for learning the associated weights for each architecture. Other hyper-parameter settings related to the search space, the gradient descent training and the search strategy are summarized in Table~\ref{tab:hyper-params}. We provide analysis aimed at justifying some of the hyper-parameter choices in Section~\ref{sec:hyper-param-analysis}. All experiments are performed on 8 Nvidia 2080Ti GPU cards.

{
Our post-search training settings largely follow \cite{liu2018darts}: We extend the number of epochs to 600 with a batch size of 96 to thoroughly re-train the selected models from scratch. We also incorporate a data pre-processing technique \emph{cutout} \cite{cutout}, and a regularization technique \emph{scheduled path dropout} introduced in \cite{nasnet2018}. In addition, to further improve the training process, an auxiliary head classifier \cite{googlenet} is appended to the architecture at approximately 2/3 depth (right after the second resolution-reduction operation). The loss from this auxiliary head classifier, scaled by a constant factor 0.4, is aggregated with the loss from the original architecture before back-propagation during training.}

\begin{table}[t]
\centering
\caption{Summary of Hyper-parameter Settings.}
\label{tab:hyper-params}
\resizebox{0.45\textwidth}{!}{%
\begin{tabular}{@{ }l|l|c@{ }}
\toprule
\multicolumn{1}{l|}{\textbf{Categories}} & \textbf{Parameters} & \textbf{Settings} \\ \midrule
\multirow{3}{*}{search space} & \# of initial channels ($Ch_{init}$) & 32 \\
 & \# of channel increments ($Ch_{inc}$) & 6 \\
 & \# of repetitions of Normal blocks ($\mathcal{N}$) & 4/5/6 \\ \midrule
\multirow{4}{*}{gradient descent} & batch size & 128 \\
 & weight decay ($L_{2}$ regularization) & 5.00E-04 \\
 & epochs & 36/600 \\
 & learning rate schedule & Cosine Annealing \cite{loshchilov2016sgdr}\\ \midrule
\multicolumn{1}{l|}{\multirow{4}{*}{search strategy}} & population size & 40 \\
\multicolumn{1}{l|}{} & \# of generations & 30 \\
\multicolumn{1}{l|}{} & crossover probability & 0.9 \\
\multicolumn{1}{l|}{} & mutation probability & 0.1 \\ \bottomrule
\end{tabular}%
}
\end{table}

\vspace{-2mm}
\subsection{Effectiveness of \ourmethod{}\label{sec:effectiveness}}
\begin{figure*}[t]
	\centering
	\begin{subfigure}[t]{.48\textwidth}
		\centering
		\includegraphics[width=0.98\textwidth]{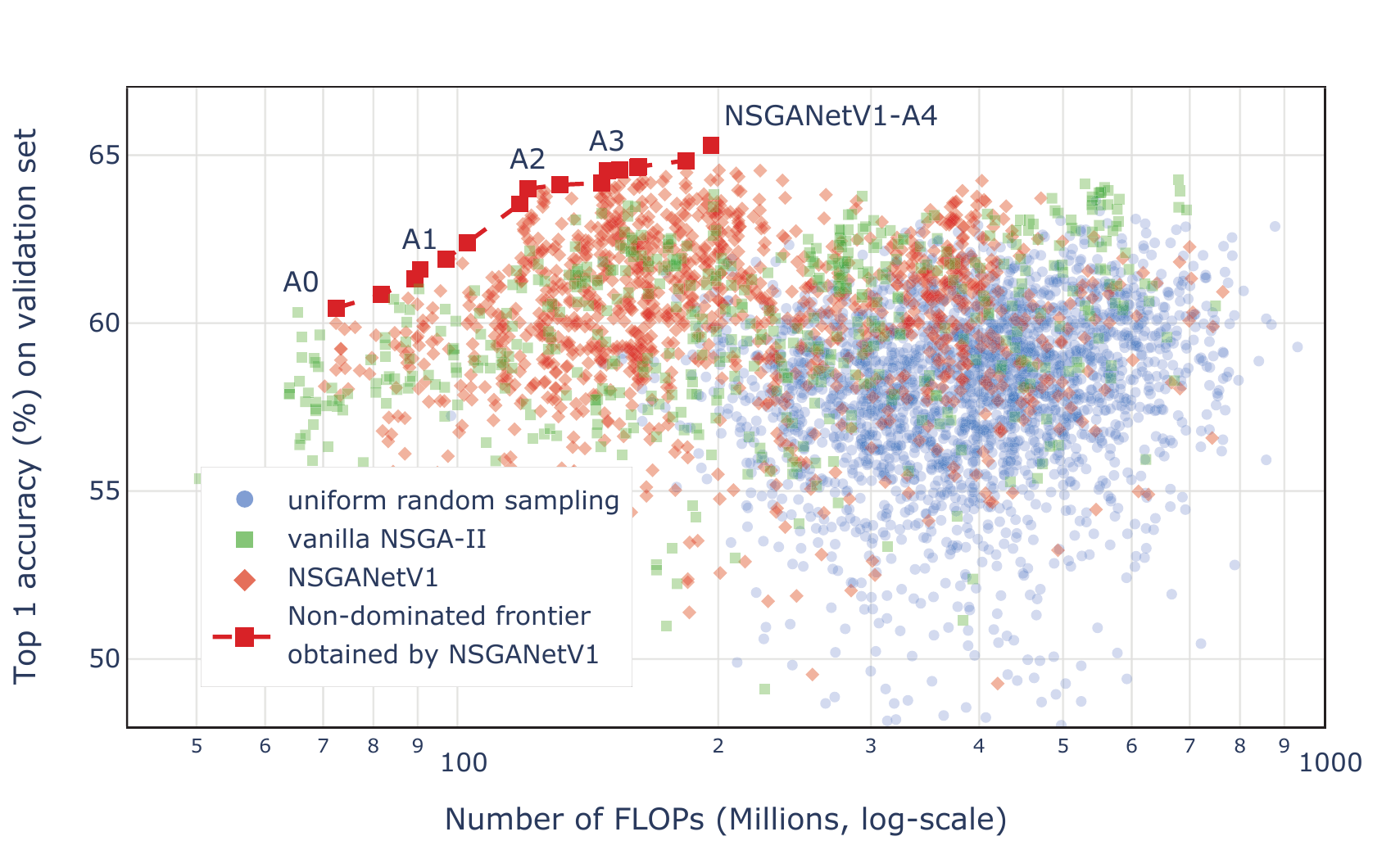}
		\caption{During search\label{fig:nsganet_nd_frontier}}
	\end{subfigure} \hfill
	\begin{subfigure}[t]{.48\textwidth}
		\centering
		\includegraphics[width=0.98\textwidth]{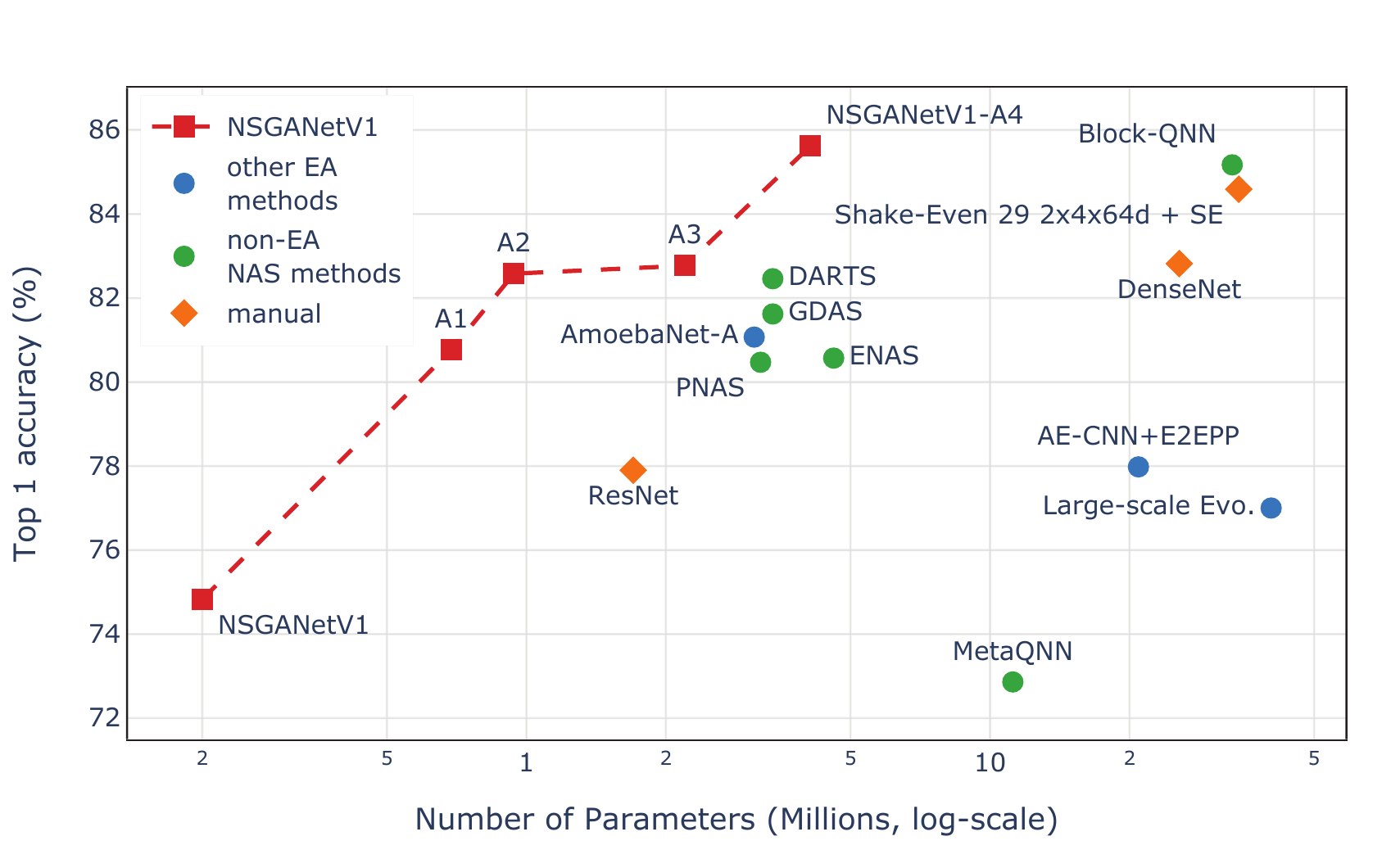}
		\caption{Post search\label{fig:cifar100}}
	\end{subfigure}
	\caption{(a) Accuracy vs. FLOPs of all architectures generated by \ourmethod{} during the course of evolution on CIFAR-100. A subset of non-dominated architectures (see text), named \ourmethod{}-A0 to A4, are re-trained thoroughly and compared with other peer methods in (b).
	\label{fig:overall_results}}
	\vspace{-1em}
\end{figure*}

We first present the objective space distribution of all architectures generated by \ourmethod{} during the course of evolution on CIFAR-100, in Fig.~\ref{fig:nsganet_nd_frontier}. We include architectures generated by the original NSGA-II algorithm and uniform random sampling as references for comparison. Details of these two methods are provided in Section~\ref{sec:efficiency}. From the set of non-dominated solutions (outlined by red box markers in Fig.~\ref{fig:nsganet_nd_frontier}), we select five architectures based on the ratio of the gain on accuracy over the sacrifice on FLOPs. For reference purposes, we name these five architectures as \ourmethod{}-A0 to -A4 in ascending FLOPs order. See Fig.~1 in the supplementary materials for a visualization of the searched architectures.

{For comparison with other peer methods, we follow the training procedure in \cite{liu2018darts} and re-train the weights of \ourmethod{}-A0 to -A4 on CIFAR-100, following the steps outlined in Section~\ref{sec:implementation_detail}}. We would like to mention that since most existing approaches do not report the number of FLOPs for the architectures used on the CIFAR-100 dataset, we instead compare their computational complexity through number of parameters to prevent potential discrepancies from re-implementation. Fig.~\ref{fig:cifar100} shows the post-search architecture comparisons, \ourmethod{}-A0 to A4, i.e., the algorithms derived in this paper, jointly dominate all other considered peer methods with a clear margin. More specifically, \ourmethod{}-A1 is more accurate than peer EA method, AE-CNN-E2EPP \cite{ae-cnn-e2epp}, while being {\textbf{30x more efficient}} in network parameters; {\ourmethod{}-A2 achieves better performance than AmoebaNet \cite{real2019regularized} and NSGA-Net \cite{lu2019nsga} with \textbf{3x fewer parameters}}. Furthermore, \ourmethod{}-A4 exceeds the classification accuracy of \emph{Shake-Even 29 2x4x64d + SE} \cite{hu2018squeeze} using {\textbf{8x fewer parameters}}. More comparisons can be found in Table~\ref{tab:cifar-100}.

Following the practice adopted in most previous approaches \cite{nasnet2018,real2019regularized,liu2018darts,pmlr-v80-pham18a,liu2018progressive}, we measure the transferability of the obtained architectures by allowing the architectures evolved on one dataset (CIFAR-100 in this case) to be inherited and used on other datasets, by retraining the weights from scratch on the new dataset--in our case, on CIFAR-10 and ImageNet.

\begin{figure*}[t]
    \centering
    \begin{subfigure}[t]{.48\textwidth}
		\centering
		\includegraphics[width=0.98\textwidth]{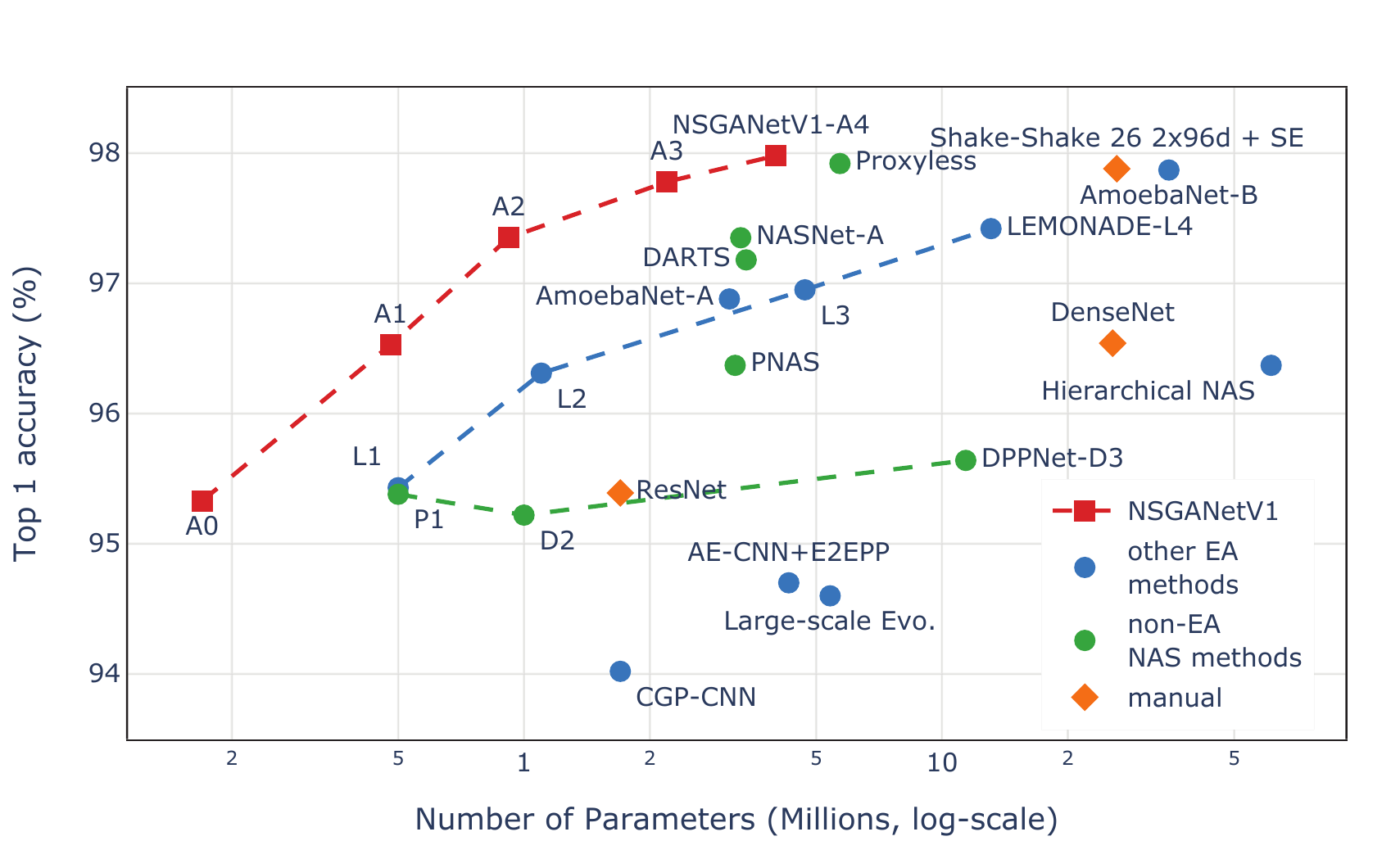}
		\caption{CIFAR-10 \label{fig:cifar10}}
	\end{subfigure} \hfill
	\begin{subfigure}[t]{.48\textwidth}
		\centering
		\includegraphics[width=0.98\textwidth]{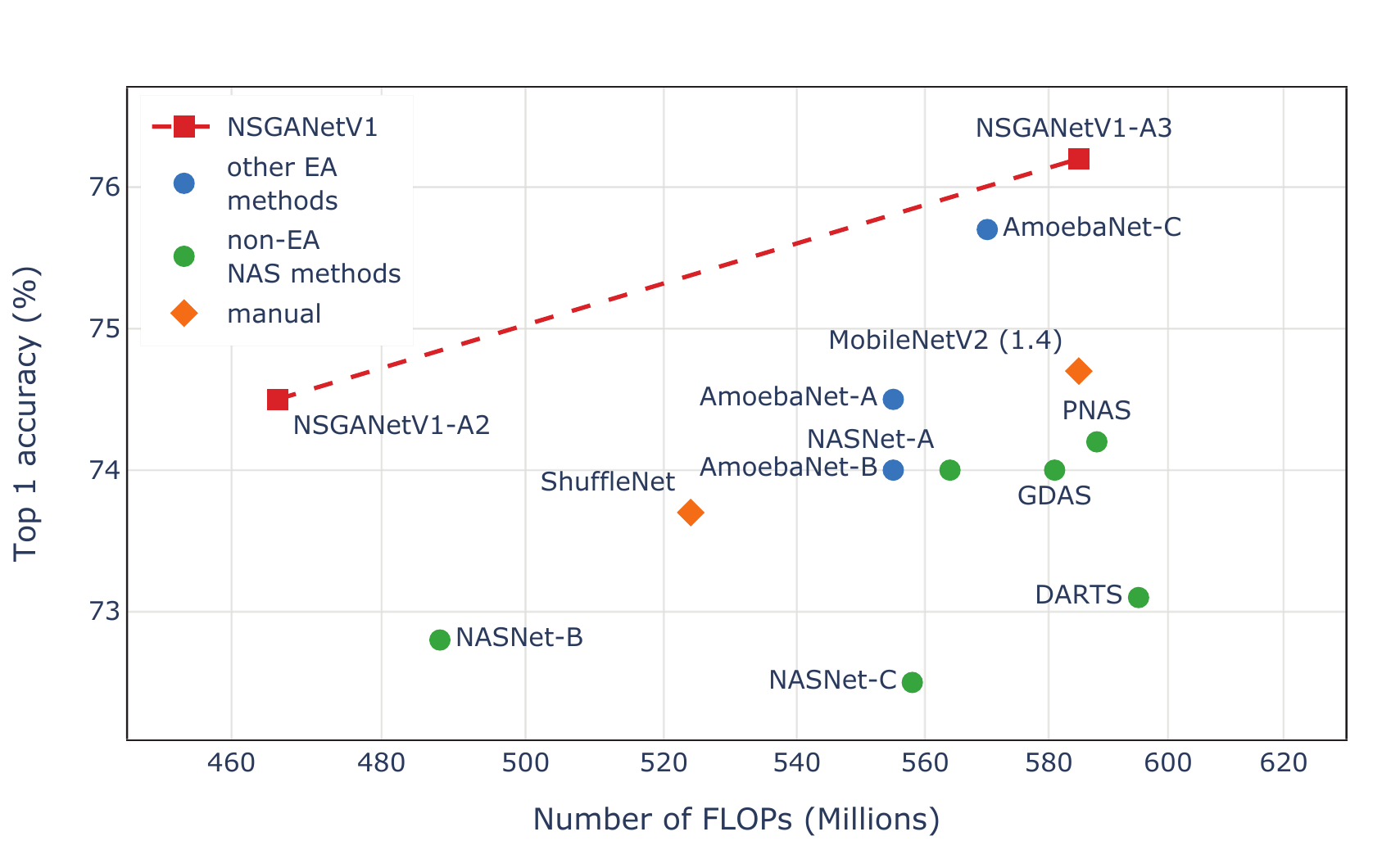}
		\caption{ImageNet \label{fig:imagenet}}
	\end{subfigure}
    \caption{Transferability of the \ourmethod{} architectures to (a) CIFAR-10, and (b) ImageNet. We compare Top-1 Accuracy \emph{vs.} Computational Complexity. Architectures joined by dashed lines are from multi-objective algorithms.}
    \label{fig:cifar10_imagenet}
\end{figure*}

The effectiveness of \ourmethod{} is further validated by the transferred performance on the CIFAR-10 dataset. As we show in Figs.~\ref{fig:cifar10}, the trade-off frontier established by \ourmethod{}-A0 to -A4 completely dominates the frontiers obtained by the peer EMO methods, both DPP-Net \cite{dong2018dpp} and LEMONADE \cite{elsken2018efficient}, as well as those obtained with other single-objective peer methods. More specifically, \ourmethod{}-A0 uses {\textbf{27x fewer parameters}} and achieves higher classification accuracy than Large-scale Evo. \cite{real2017largescale}. \ourmethod{}-A1 outperforms Hierarchical NAS \cite{liu2018hierarchical} and DenseNet \cite{densenet} in classification, while {\textbf{saving 122x}} and {\textbf{51x}} in {\textbf{parameters}}. {\ourmethod{}-A2 uses \textbf{4x less parameters} to achieve similar performance as compared to NSGA-Net \cite{lu2019nsga}.} Furthermore, \ourmethod{}-A4 exceeds previous \sota{} results reported by Proxyless NAS \cite{cai2018proxylessnas} while being {\textbf{1.4x more compact}}. Refer to Table~\ref{tab:cifar-10} for more comparisons.

\begin{table*}[!htbp]
\caption{Comparison between \ourmethod{} and other baseline methods. \ourmethod{} architectures are obtained by searching on CIFAR-100. \ourmethod{} results on CIFAR-10 and ImageNet are obtained by re-training the weights with images from their respective datasets. Ratio-to-\ourmethod{} indicates the resulting savings on \#Params and \#FLOPs. The search cost is compared in GPU-days, calculated by multiplying the number of GPU cards deployed with the execution time in days.}
\label{tab:cifar}
    \begin{subtable}[h]{0.48\textwidth}
        \centering
        \caption{CIFAR-10}
        \label{tab:cifar-10}
        \resizebox{0.98\textwidth}{!}{%
        \begin{tabular}{@{ }l|cc|ccc@{ }}
        \toprule
        \multirow{2}{*}{\textbf{Architecture}} & \multirow{2}{*}{\textbf{\begin{tabular}[c]{@{}c@{}}Search\\ Method\end{tabular}}} & \multirow{2}{*}{\textbf{\begin{tabular}[c]{@{}c@{}}GPU-\\ Days\end{tabular}}} & \multirow{2}{*}{\textbf{Top-1 Acc.}} & \multirow{2}{*}{\textbf{\#Params}} & \multirow{2}{*}{\textbf{\begin{tabular}[c]{@{}c@{}}Ratio-to-\\ \ourmethod{}\end{tabular}}} \\
         &  &  &  &  &  \\ \midrule
        \textbf{\ourmethod{}-A0} & \textbf{EA} & \textbf{27} & \textbf{95.33\%} & \textbf{0.2M} & \textbf{1x} \\
        CGP-CNN \cite{cgp-cnn} & EA & 27 & 94.02\% & 1.7M & 8.5x \\
        Large-scale Evo. \cite{real2017largescale} & EA & 2,750 & 94.60\% & 5.4M & 27x \\
        AE-CNN+E2EPP \cite{ae-cnn-e2epp} & EA & 7 & 94.70\% & 4.3M & 21x \\
        ResNet \cite{resnet} & manual & - & 95.39\% & 1.7M & 8.5x \\ \midrule
        \textbf{\ourmethod{}-A1} & \textbf{EA} & \textbf{27} & \textbf{96.51\%} & \textbf{0.5M} & \textbf{1x} \\
        Hierarchical NAS \cite{liu2018hierarchical} & EA & 300 & 96.37\% & 61.3M & 122x \\
        PNAS \cite{liu2018progressive} & SMBO & 150 & 96.37\% & 3.2M & 6.4x \\
        DenseNet \cite{densenet} & manual & - & 96.54\% & 25.6M & 51x \\ \midrule
        \textbf{\ourmethod{}-A2} & \textbf{EA} & \textbf{27} & \textbf{97.35\%} & \textbf{0.9M} & \textbf{1x} \\
        {CNN-GA \cite{cnn-ga}} & {EA} & {35} & {96.78\%} & {2.9M} & {3.2x} \\
        AmoebaNet-A \cite{real2019regularized} & EA & 3,150 & 96.88\% & 3.1M & 3.4x \\
        DARTS \cite{liu2018darts} & relaxation & 1 & 97.18\% & 3.4M & 3.8x \\
        {NSGA-Net \cite{lu2019nsga}} & {EA} & {4} & {97.25\%} & {3.3M} & {3.7x} \\\midrule
        \textbf{\ourmethod{}-A3} & \textbf{EA} & \textbf{27} & \textbf{97.78\%} & \textbf{2.2M} & \textbf{1x} \\
        NASNet-A \cite{nasnet2018} & RL & 1,575 & 97.35\% & 3.3M & 1.5x \\
        LEMONADE \cite{elsken2018efficient} & EA & 90 & 97.42\% & 13.1M & 6.0x \\ \midrule
        \textbf{\ourmethod{}-A4} & \textbf{EA} & \textbf{27} & \textbf{97.98\%} & \textbf{4.0M} & \textbf{1x} \\
        AmoebaNet-B \cite{real2019regularized} & EA & 3,150 & 97.87\% & 34.9M & 8.7x \\
        Proxyless NAS \cite{cai2018proxylessnas} & RL & 1,500 & 97.92\% & 5.7 M & 1.4x \\ \bottomrule
        \end{tabular}}
    \end{subtable}
    \hfill
    \begin{subtable}[h]{0.48\textwidth}
        \centering
        \caption{CIFAR-100}
        \label{tab:cifar-100}
        \resizebox{0.98\textwidth}{!}{%
        \begin{tabular}{@{ }l|cc|ccc@{ }}
        \toprule
        \multirow{2}{*}{\textbf{Architecture}} & \multirow{2}{*}{\textbf{\begin{tabular}[c]{@{}c@{}}Search\\ Method\end{tabular}}} & \multirow{2}{*}{\textbf{\begin{tabular}[c]{@{}c@{}}GPU-\\ Days\end{tabular}}} & \multirow{2}{*}{\textbf{Top-1 Acc.}} & \multirow{2}{*}{\textbf{\#Params}} & \multirow{2}{*}{\textbf{\begin{tabular}[c]{@{}c@{}}Ratio-to-\\ \ourmethod{}\end{tabular}}} \\
         &  &  &  &  &  \\ \midrule
        \textbf{\ourmethod{}-A0} & \textbf{EA} & \textbf{27} & \textbf{74.83\%} & \textbf{0.2M} & \textbf{1x} \\
        Genetic CNN \cite{genetic-cnn} & EA & 17 & 70.95\% & - & - \\
        MetaQNN \cite{baker2017metaqnn} & RL & 90 & 72.86\% & 11.2M & 56x \\ \midrule
        \textbf{\ourmethod{}-A1} & \textbf{EA} & \textbf{27} & \textbf{80.77\%} & \textbf{0.7M} & \textbf{1x} \\
        Large-scale Evo. \cite{real2017largescale} & EA & 2,750 & 77.00\% & 40.4M & 58x \\
        ResNet \cite{resnet} & manual & - & 77.90\% & 1.7M & 2.4x \\
        AE-CNN+E2EPP \cite{ae-cnn-e2epp} & EA & 10 & 77.98\% & 20.9M & 30x \\
        {NSGA-Net \cite{lu2019nsga}} & {EA} & {8} & {79.26\%} & {3.3M} & {4.7x} \\
        {CNN-GA \cite{cnn-ga}} & {EA} & {40} & {79.47\%} & {4.1M} & {5.9x} \\
        PNAS \cite{liu2018progressive} & SMBO & 150 & 80.47\% & 3.2M & 4.6x \\
        ENAS \cite{pmlr-v80-pham18a} & RL & 0.5 & 80.57\% & 4.6M & 6.6x \\ \midrule
        \textbf{\ourmethod{}-A2} & \textbf{EA} & \textbf{27} & \textbf{82.58\%} & \textbf{0.9M} & \textbf{1x} \\
        AmoebaNet-A \cite{real2019regularized} & EA & 3,150 & 81.07\% & 3.1M & 3.4x \\
        GDAS \cite{Dong_2019_CVPR} & relaxation & 0.2 & 81.62\% & 3.4M & 3.8x \\
        DARTS \cite{liu2018darts} & relaxation & 1 & 82.46\% & 3.4M & 3.8x \\ \midrule
        \textbf{\ourmethod{}-A3} & \textbf{EA} & \textbf{27} & \textbf{82.77\%} & \textbf{2.2M} & \textbf{1x} \\ \midrule
        \textbf{\ourmethod{}-A4} & \textbf{EA} & \textbf{27} & \textbf{85.62\%} & \textbf{4.1M} & \textbf{1x} \\
        DenseNet \cite{densenet} & manual & - & 82.82\% & 25.6M & 6.2x \\
        SENet \cite{hu2018squeeze} & manual & - & 84.59\% & 34.4M & 8.4x \\
        Block-QNN \cite{zhong2017blockqnn} & RL & 32 & 85.17\% & 33.3M & 8.1x \\ \bottomrule
        \end{tabular}}
     \end{subtable} \\
     \centering{
     \begin{subtable}[h]{0.92\textwidth}
        \vspace{4mm}
        \caption{ImageNet}
        \label{tab:imagenet}
        \resizebox{0.95\textwidth}{!}{%
        \begin{tabular}{@{ }l|cc|cc|cc|cc@{ }}
        \toprule
        \textbf{Architecture} & \multicolumn{1}{l}{\textbf{Search Method}} & \multicolumn{1}{l|}{\textbf{GPU-Days}} & \multicolumn{1}{l}{\textbf{Top-1 Acc.}} & \multicolumn{1}{l|}{\textbf{Top-5 Acc.}} & \multicolumn{1}{l}{\textbf{\#Params}} & \multicolumn{1}{l|}{\textbf{Ratio-to-\ourmethod{}}} & \multicolumn{1}{l}{\textbf{\#FLOPs}} & \multicolumn{1}{l}{\textbf{Ratio-to-\ourmethod{}}} \\ \midrule
        \textbf{\ourmethod{}-A1} & \textbf{EA} & \textbf{27} & 70.9\% & 90.0\% & \textbf{3.0M} & \textbf{1x} & \textbf{270M} & \textbf{1x} \\
        MobileNet-V2 \cite{sandler2018mobilenetv2} & manual & - & \textbf{72.0\%} & \textbf{91.0\%} & 3.4M & 1.1x & 300M & 1.1x \\ \midrule
        \textbf{\ourmethod{}-A2} & \textbf{EA} & \textbf{27} & \textbf{74.5\%} & \textbf{92.0\%} & \textbf{4.1M} & \textbf{1x} & \textbf{466M} & \textbf{1x} \\
        ShuffleNet \cite{zhang2018shufflenet} & manual & - & 73.7\% & - & 5.4M & 1.3x & 524M & 1.1x \\
        NASNet-A \cite{nasnet2018} & RL & 1,575 & 74.0\% & 91.3\% & 5.3M & 1.3x & 564M & 1.2x \\
        PNAS \cite{liu2018progressive} & SMBO & 150 & 74.2\% & 91.9\% & 5.1M & 1.2x & 588M & 1.3x \\
        AmoebaNet-A \cite{real2019regularized} & EA & 3,150 & 74.5\% & 92.0\% & 5.1M & 1.2x & 555M & 1.2x \\
        DARTS \cite{liu2018darts} & relaxation & 1 & 73.1\% & 91.0\% & 4.9M & 1.2x & 595M & 1.3x \\ \midrule
        \textbf{\ourmethod{}-A3} & \textbf{EA} & \textbf{27} & \textbf{76.2\%} & \textbf{93.0\%} & \textbf{5.0M} & \textbf{1x} & 585M & 1x \\
        MobileNetV2 (1.4) \cite{sandler2018mobilenetv2} & manual & - & 74.7\% & 92.5\% & 6.06M & 1.2x & 582M & 1x \\
        AmoebaNet-C \cite{real2019regularized} & EA & 3,150 & 75.7\% & 92.4\% & 6.4M & 1.3x & \textbf{570M} & \textbf{1x} \\ \bottomrule
        \end{tabular}}
     \end{subtable}}
      \begin{minipage}{.9\linewidth}
      \begin{threeparttable}
        \begin{tablenotes}
        \scriptsize{
        \item[\textdagger] SMBO stands for sequential model-based optimization. SENet is the abbreviation for Shake-Even 29 2x4x64d + SE.
        \item[\textdaggerdbl] The CIFAR-100 accuracy and \#params for ENAS \cite{pmlr-v80-pham18a} and DARTS \cite{liu2018darts} are from \cite{Dong_2019_CVPR}. \#Params for AE-CNN+E2EPP are from \cite{ae-cnn}.
        }
        \end{tablenotes}
      \end{threeparttable}
    \end{minipage}
\end{table*}

For transfer performance comparison on the ImageNet dataset, we follow previous work \cite{nasnet2018,liu2018darts,pmlr-v80-pham18a,sandler2018mobilenetv2} and use the ImageNet-mobile setting, i.e., the setting where number of FLOPs is less than 600M. The \ourmethod{}-A0 is too simple for the ImageNet dataset and \ourmethod{}-A4 exceeds the 600M FLOPs threshold for the mobile setting, so we provide results only for \ourmethod{}-A1, -A2 and -A3. Fig.~\ref{fig:imagenet} compares the objective space with the other peer methods. Clearly, \ourmethod{} can achieve a better trade-off between the objectives. \ourmethod{}-A2 dominates a wide range of peer methods including ShuffleNet \cite{zhang2018shufflenet} by human experts, NASNet-A \cite{nasnet2018} by RL, DARTS \cite{liu2018darts} by relaxation-based methods, and AmoebaNet-A \cite{real2019regularized} by EA. Moreover, \ourmethod{}-A3 surpasses previous \sota{} performance reported by MobileNet-V2 \cite{sandler2018mobilenetv2} and AmoebaNet-C \cite{real2019regularized} on mobile-setting with a marginal overhead in FLOPs (1\% - 3\%).

\vspace{-2mm}
\subsection{Efficiency of \ourmethod{}\label{sec:efficiency}}
\begin{figure*}[t]
	\centering
	\begin{subfigure}[t]{.48\textwidth}
		\centering
		\includegraphics[width=0.98\textwidth]{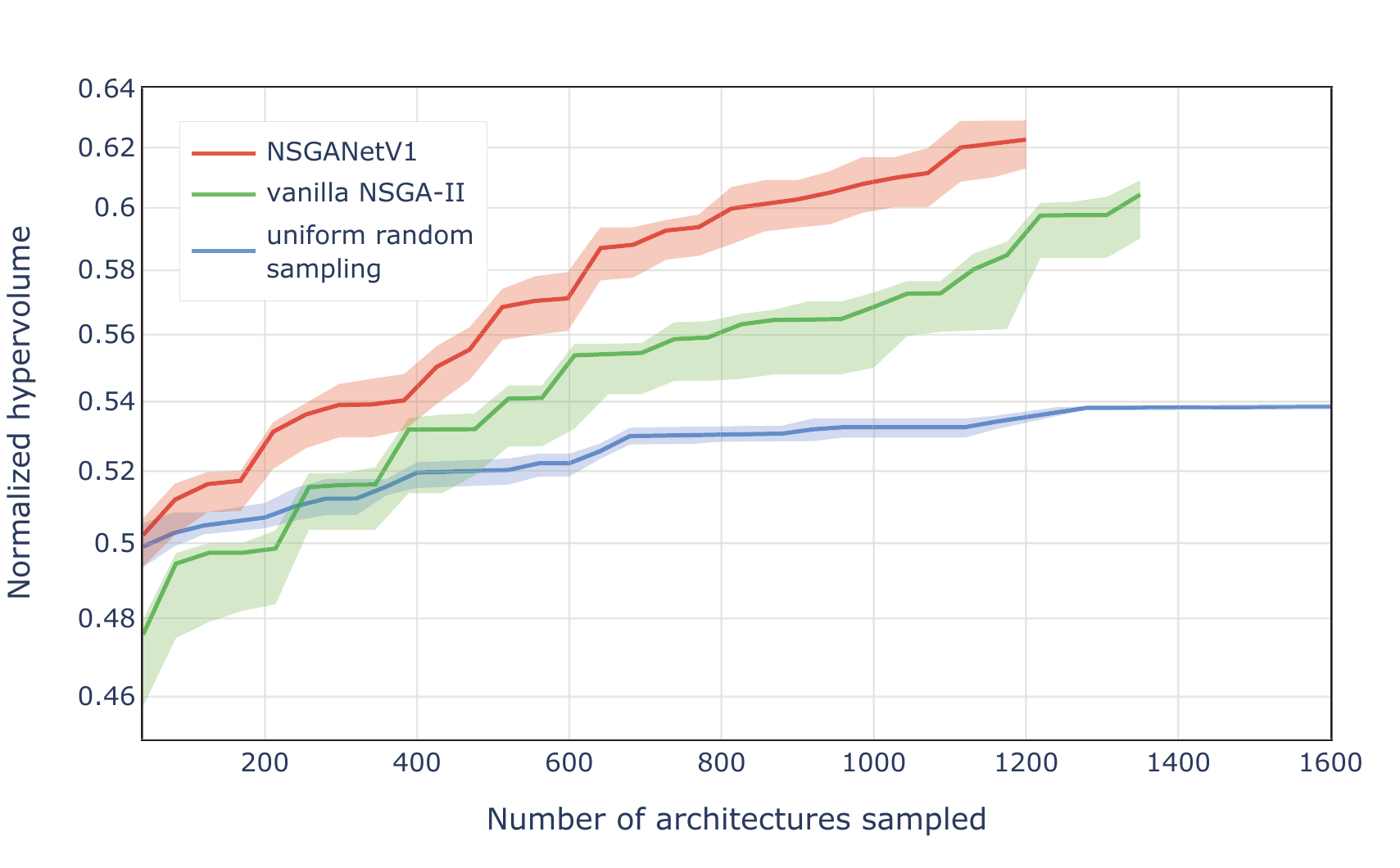}
		\caption{HV \label{fig:nsganet_hv}}
	\end{subfigure} \hfill
	\begin{subfigure}[t]{.48\textwidth}
		\centering
		\includegraphics[width=0.98\textwidth]{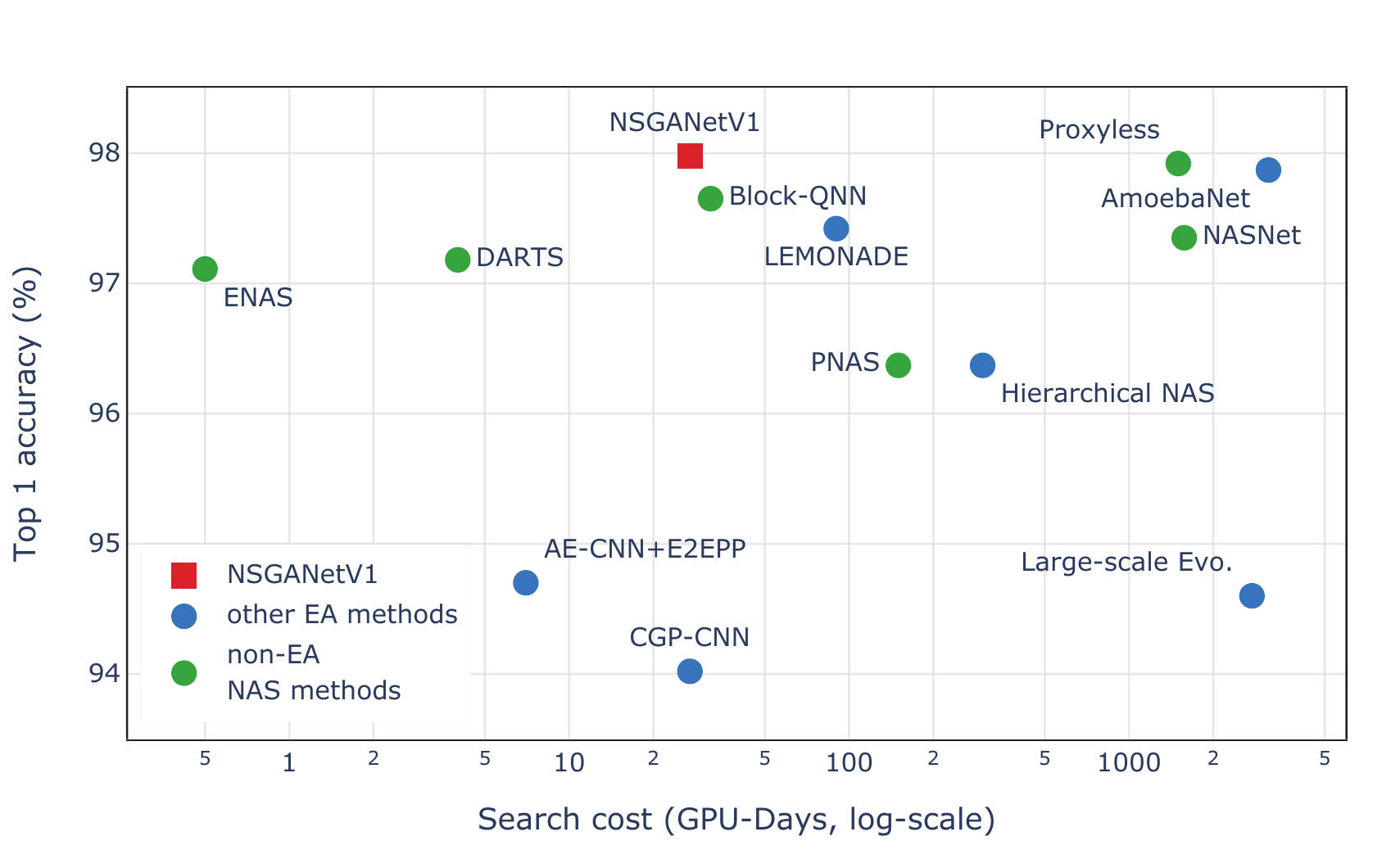}
		\caption{Search cost vs. Accuracy\label{fig:nsganet_search_cost}}
	\end{subfigure}
	\caption{Search efficiency comparison between \ourmethod{} and other baselines in terms of (a) HV, and (b) the required compute time in GPU-Days. The search cost is measured on CIFAR-10 for most methods, except \ourmethod{} and Block-QNN \cite{zhong2017blockqnn}, where the CIFAR-100 dataset is used for.
	\label{fig:search_efficiency}}
	\vspace{-1em}
\end{figure*}

Comparing the search phase contribution to the success of different NAS algorithms can be difficult and ambiguous due to substantial differences in search spaces and training procedures used during the search. Therefore, we use \emph{vanilla NSGA-II} and \emph{uniform random sampling} as comparisons to demonstrate the efficiency of the search phase in \ourmethod{}. All three methods use the same search space and performance estimation strategy as described in Section~\ref{sec:approach}. The vanilla NSGA-II is implemented by discretizing the crossover and mutation operators in the original NSGA-II \cite{deb2002fast} algorithm with all hyper-parameters set to default values; and it does not utilize any additional enhancements---e.g., the Bayesian-Network-model-based exploitation. The uniform random sampling method is implemented by replacing the crossover and mutation operators in the original NSGA-II algorithm with an initialization method that uniformly samples the search space. We run each of the three methods five times and record the 25-percentile, median and 75-percentile of the normalized HV (NHV) that we obtain. The NHV measurements shown in Fig.~\ref{fig:nsganet_hv} suggest that \ourmethod{} is capable of finding more accurate and simpler architectures than vanilla NSGA-II or uniform random sampling (even with an extended search budget), in a more efficient manner.

Apart from the HV metric, another important aspect of demonstrating the efficiency of NAS is the computational complexity of the methods. Since theoretical analysis of the computational complexity of different NAS methods is challenging, we compare the computation time spent on Graphics Processing Units (GPUs), \emph{GPU-Days}, by each approach to arrive at the reported architectures. The number of GPU-Days is calculated by multiplying the number of employed GPU cards by the execution time (in units of days).

One run of \ourmethod{} on the CIFAR-100 dataset takes approximately 27 GPU-Days to finish, averaged over five runs. The search costs of most of the peer methods are measured on the CIFAR-10 dataset, except for Block-QNN \cite{zhong2017blockqnn} which is measured on CIFAR-100. From the search cost comparison in Fig.~\ref{fig:nsganet_search_cost}, we observe that our proposed algorithm is more efficient at identifying a set of architectures than a number of other approaches, and the set of architectures obtained has higher performance. More specifically, \ourmethod{} simultaneously finds multiple architectures while using {\textbf{10x} to \textbf{100x less GPU-days}} in searching than most of the considered peer methods, including Hierarchical NAS \cite{liu2018hierarchical}, AmoebaNet \cite{real2019regularized}, NASNet \cite{nasnet2018}, and Proxyless NAS \cite{cai2018proxylessnas}, all of which find a single architecture at a time. When compared to the peer multi-objective NAS method, LEMONADE \cite{elsken2018efficient}, \ourmethod{} manages to obtain a better (in the Pareto dominance sense) set of architectures than LEMONADE with {\textbf{3x fewer GPU-Days}}. Further experiments and comparisons are provided in the supplementary materials under Section VII-A.

\vspace{-2mm}
\subsection{Observations on Evolved Architectures}
\begin{figure}[!htbp]
	\centering
	\begin{subfigure}[t]{.35\textwidth}
		\centering
		\includegraphics[width=0.98\textwidth]{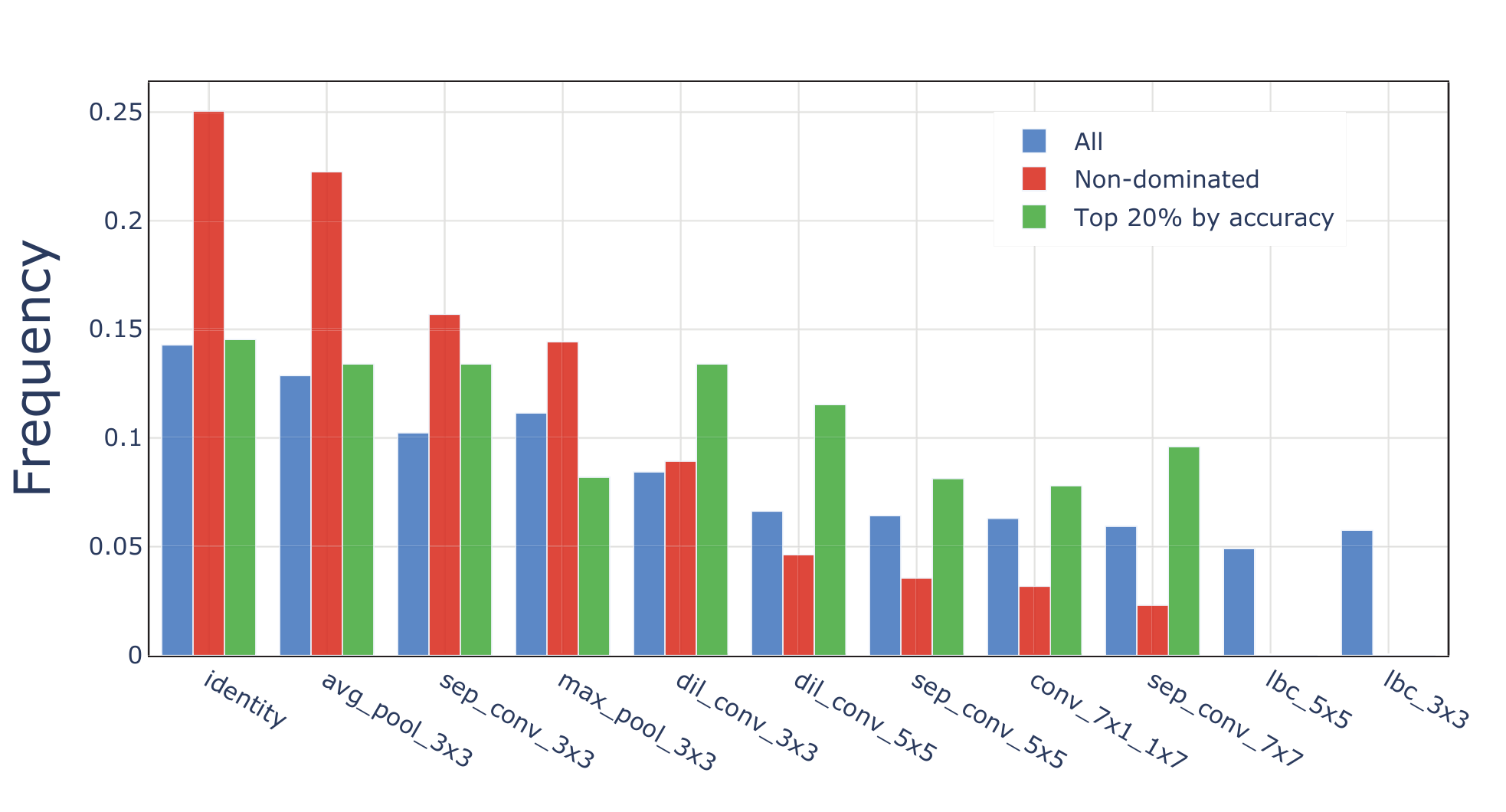}
		\caption{Frequency of operation choices.\label{fig:ops_frequency}}
	\end{subfigure}\\
	\begin{subfigure}[t]{.35\textwidth}
		\centering
		\includegraphics[width=0.98\textwidth]{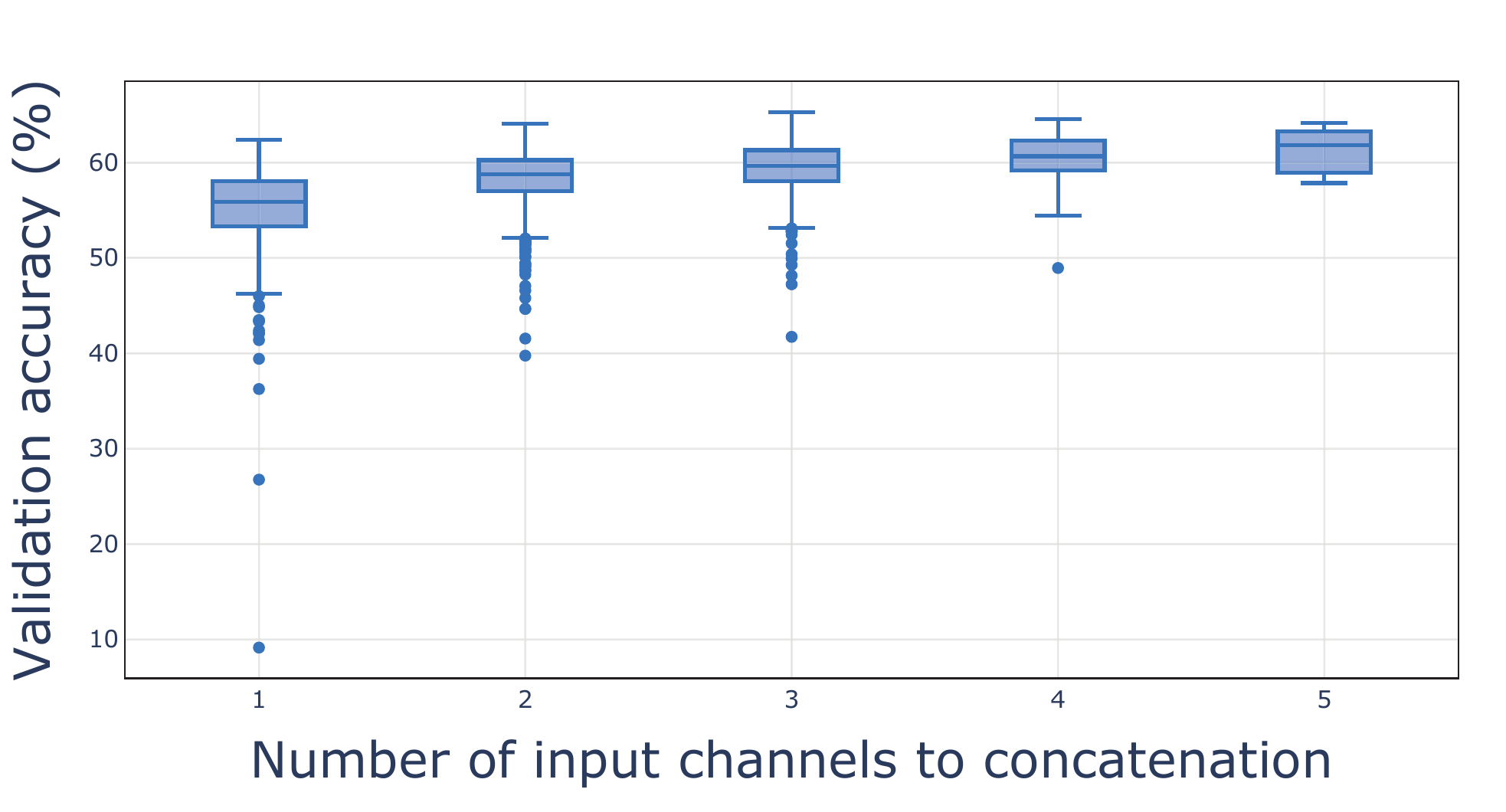}
		\caption{Concatenation dimension vs. Accuracy\label{fig:concat_inputs}}
	\end{subfigure}
	\caption{\textbf{Post-Search Analysis:} (a) Frequency of each operation selected during the search. (b) Effect of number of input channels that are concatenation on the validation accuracy. More channels improve the predictive performance of the architectures, but adversely affect the computational efficiency.
	\label{fig:intepretation}}
	\vspace{-1em}
\end{figure}

Population-based approaches with multiple conflicting objectives often lead to a set of diverse solution candidates, which can be ``mined'' for commonly shared design principles \cite{deb2006innovization}. In order to discover any patterns for more efficient architecture design, we analyzed the entire history of architectures generated by \ourmethod{}. We make the following observations:

\begin{itemize}
    \item Non-parametric operations---e.g., skip connections (\emph{identity}) and average pooling (\emph{avg_pool_3x3})---are effective in trading off performance for complexity. Empirically, we notice that three out of the four most frequently used operations in non-dominated architectures are non-parametric, as shown in Fig.~\ref{fig:ops_frequency} ({see also supplementary materials under Section~VII-B for our follow-up study}).

    \item Larger kernel size and parallel operations improve classification accuracy, as shown in Fig.~\ref{fig:ops_frequency} and Fig.~\ref{fig:concat_inputs} respectively. In particular, the frequencies of convolutions with large kernel sizes (e.g., \emph{dil_conv_5x5}, \emph{conv_7x1_1x7}) are significantly higher in the top-20\% most accurate architectures than in non-dominated architectures in general, which must also balance FLOPs. Similar findings are also reported in previous work \cite{xie2017aggregated,real2019regularized}.
\end{itemize}

The above common properties of multiple final non-dominated solutions stay as important knowledge for future applications. It is noteworthy that such a post-optimal knowledge extraction process is possible only from a multi-objective optimization study, another benefit that we enjoy for posing NAS as a multi-objective optimization problem.


%% file: ablation.tex
\section{Further Analysis\label{sec:analysis}}
The overarching goal of NAS is to find architecture models that generalize to new instances of what the models were trained on. We usually quantify generalization by measuring the performance of a model on a held-out testing set. Since many computer vision benchmark datasets, including the three datasets used in this paper---i.e. CIFAR-10, CIFAR-100, and ImageNet, have been the focus of intense research for almost a decade, does the steady stream of promising empirical results from NAS simply arise from overfitting of these excessively re-used testing sets? Does advancement on these testing sets imply better robustness vis-a-vis commonly observable corruptions in images and adversarial images by which the human vision system is more robust? To answer these questions in a quantitative manner, in this section, we provide systematic studies on newly proposed testing sets from the CNN literature, followed by hyper-parameter analysis.

\vspace{-2mm}
\subsection{Generalization}
By mimicking the documented curation process of the original CIFAR-10 and ImageNet datasets, Recht et al. \cite{recht2019imagenet} propose two new testing sets, CIFAR-10.1 and ImageNet-V2. Refer to supplementary materials under Section~V-A for details and examples of the new testing sets. Representative architectures are selected from each of the main categories (i.e., RL, EA, relaxation-based, and manual). The selected architectures are similar in number of parameters or FLOPs, except DenseNet-BC \cite{densenet} and Inception-V1 \cite{googlenet}. All architectures are trained on the original CIFAR-10 and ImageNet training sets as in Section~\ref{sec:effectiveness}, then evaluated on CIFAR-10.1 and ImageNet-V2, respectively.

\begin{figure}[t]
	\centering
	\begin{subfigure}[t]{.24\textwidth}
		\centering
		\includegraphics[width=0.98\textwidth]{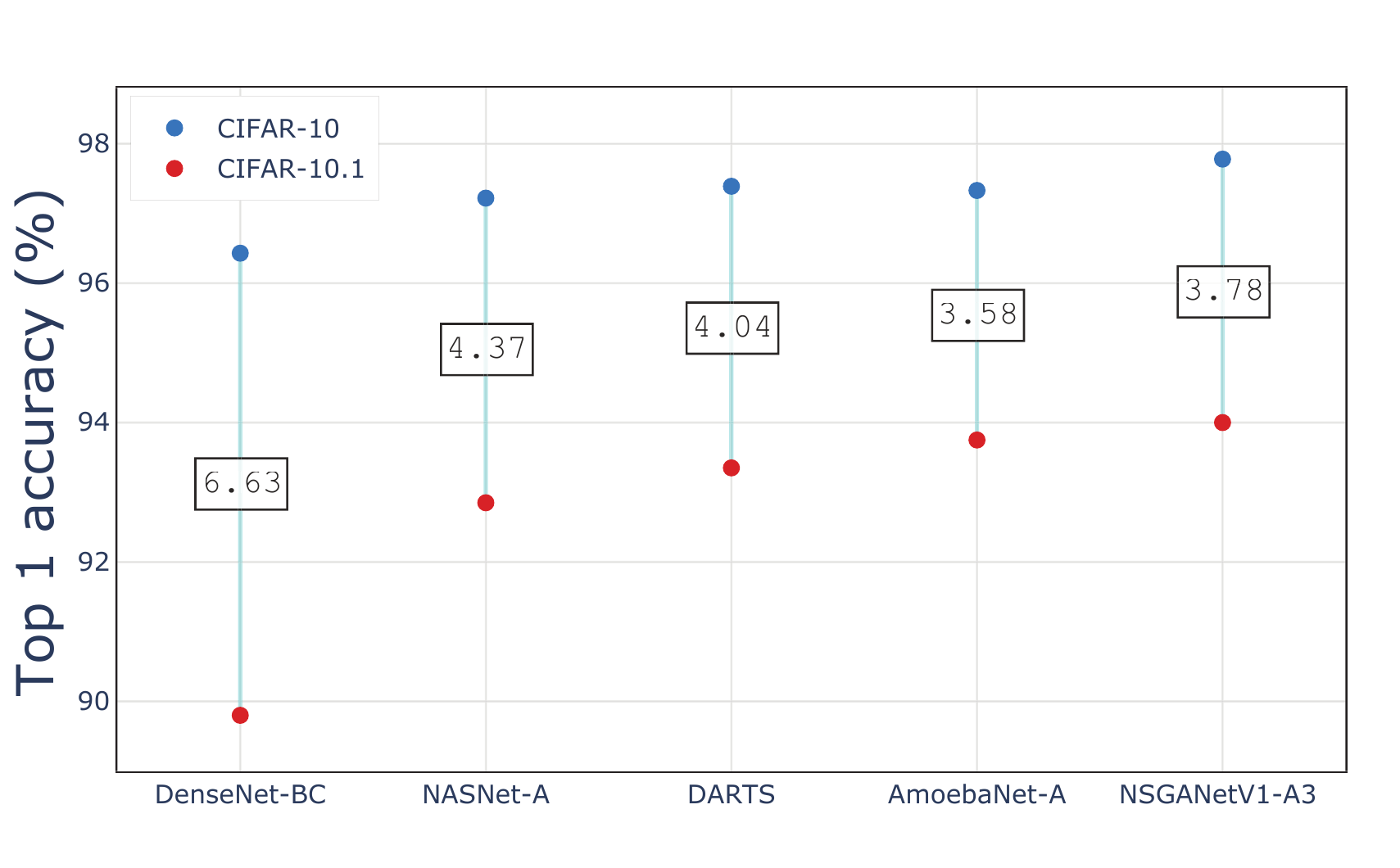}
		\caption{CIFAR-10.1\label{fig:cifar10_generalization}}
	\end{subfigure}
	\begin{subfigure}[t]{.24\textwidth}
		\centering
		\includegraphics[width=0.98\textwidth]{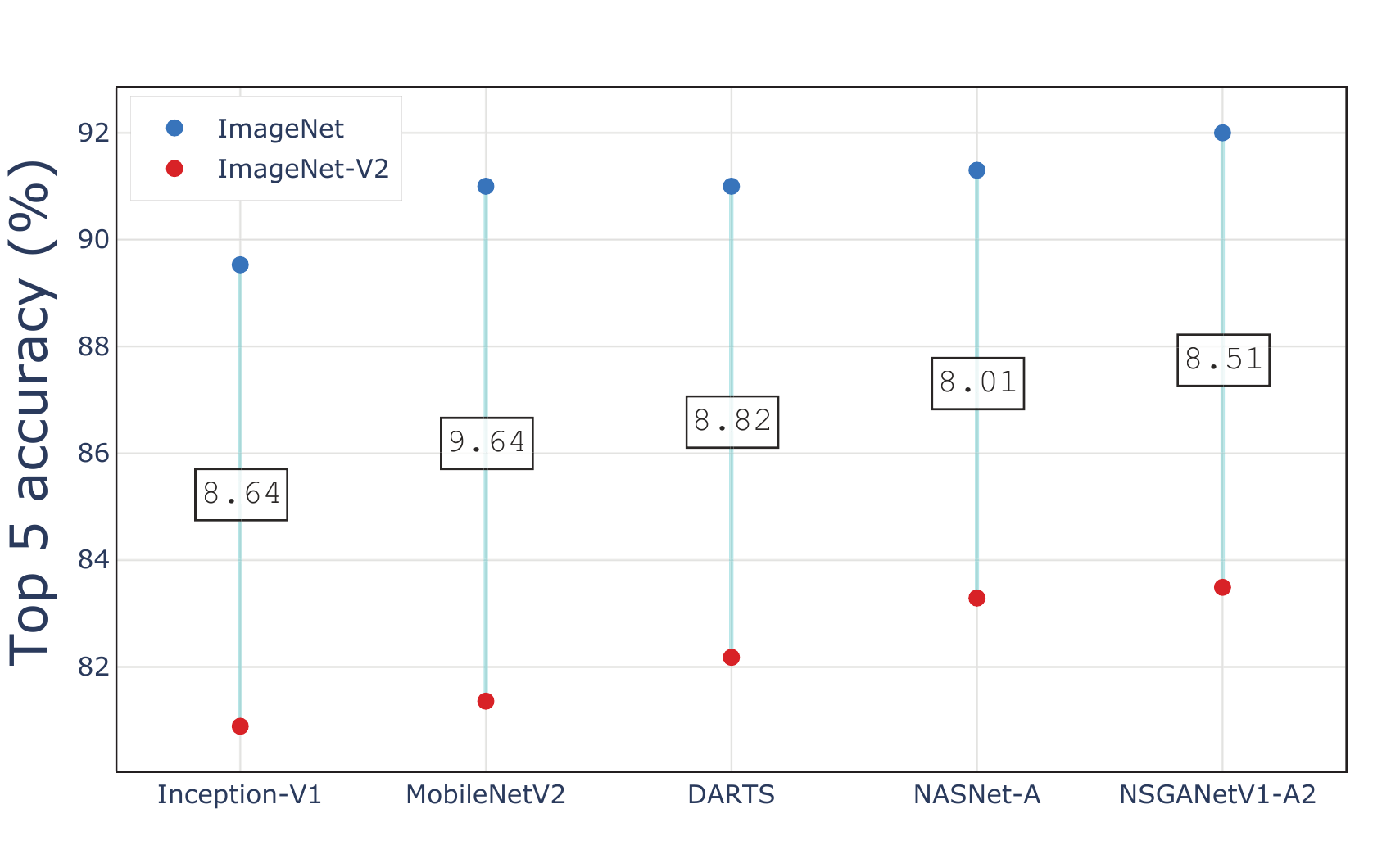}
		\caption{ImageNet-V2\label{fig:imagenet_generalization}}
	\end{subfigure}
	\caption{\textbf{Generalization:} We evaluate the models on new and extended test sets for (a) CIFAR-10, and (b) ImageNet. Numbers in the boxes indicate absolute drop in accuracy (\%).
	\label{fig:generalization}\vspace{-0.3cm}}
\end{figure}

It is evident from the results summarized in Figs.~\ref{fig:cifar10_generalization} and \ref{fig:imagenet_generalization} that there is a significant drop in accuracy of 3\% - 7\% on CIFAR-10.1 and 8\% to 10\% on ImageNet-V2 across architectures. However, the relative rank-order of accuracy on the original testing sets translates well to the new testing sets, i.e., the architecture with the highest accuracy on the original testing set (\ourmethod{} in this case) is also the architecture with the highest accuracy on new testing sets. Additionally, we observe that the accuracy gains on the original testing sets translate to larger gains on the new testing sets, especially in the case of CIFAR-10 (curvatures of red vs. blue markers in Fig.~\ref{fig:generalization}). These results provide evidence that extensive benchmarking on the original testing sets is an effective way to measure the progress of architectures.

\vspace{-2mm}
\subsection{Robustness}
The vulnerability to small changes in query images may very likely prevent the deployment of deep learning vision systems in safety-critical applications at scale. Understanding the architectural advancements under the scope of robustness against various forms of corruption is still in its infancy. Hendrycks and Dietterich \cite{hendrycks2018benchmarking} recently introduced two new testing datasets, CIFAR-10-C and CIFAR-100-C, by applying commonly observable corruptions (e.g., noise, weather, compression, etc.) to the clean images from the original datasets. Each dataset contains images perturbed by 19 different types of corruption at five different levels of severity. More details and visualizations are provided in supplementary materials {under} Section~V-B. In addition, we include adversarial images as examples of worst-case corruption. We use the fast gradient signed method (FGSM) \cite{goodfellow2014explaining} to construct adversarial examples for both the CIFAR-10 and -100 datasets. The severity of the attack is controlled via a hyper-parameter $\epsilon$ as shown below:
\begin{align*}
    \bm{x}' = \bm{x} + \epsilon ~ \text{sign}( \nabla_{\bm{x}}\mathcal{L}(\bm{x}, y_{true})),
\end{align*}
\noindent where $\bm{x}$ is the original image, $\bm{x}'$ is the adversarial image, $y_{true}$ is the true class label, and $\mathcal{L}$ is the cross-entropy loss. Following the previous section, we pick representative architectures of similar complexities from each of the main categories. {Using the weights learned on the clean images from the original CIFAR-10/100 training sets, we evaluate each architecture's classification performance on the corrupted datasets as our measure of robustness}.

\begin{figure}[t]
	\centering
	\begin{subfigure}[t]{.45\textwidth}
		\centering
		\includegraphics[width=0.98\textwidth]{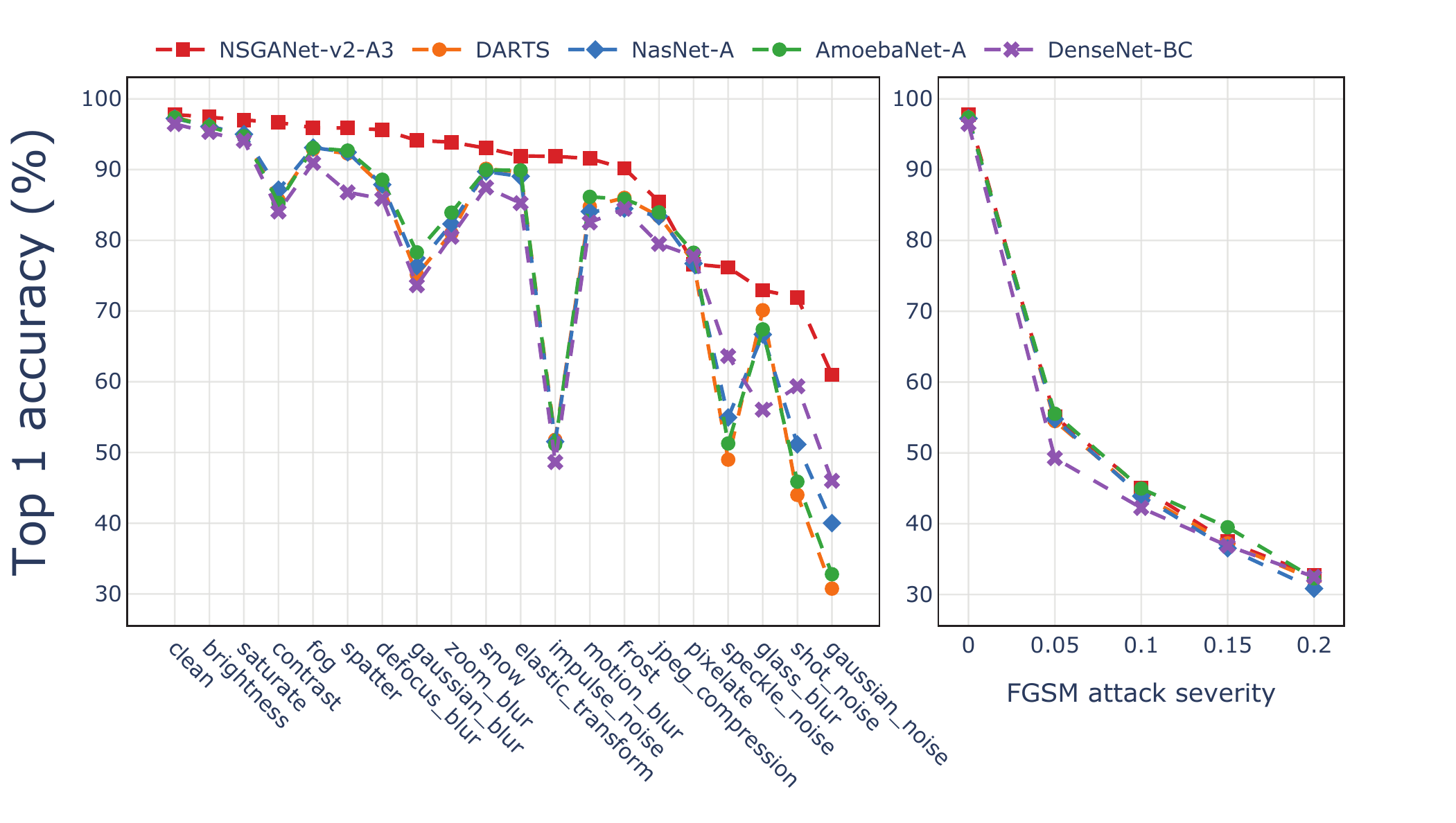}
		\caption{CIFAR-10\label{fig:cifar10_robustness_adv}}
	\end{subfigure}\\
	\begin{subfigure}[t]{.45\textwidth}
		\centering
		\includegraphics[width=0.98\textwidth]{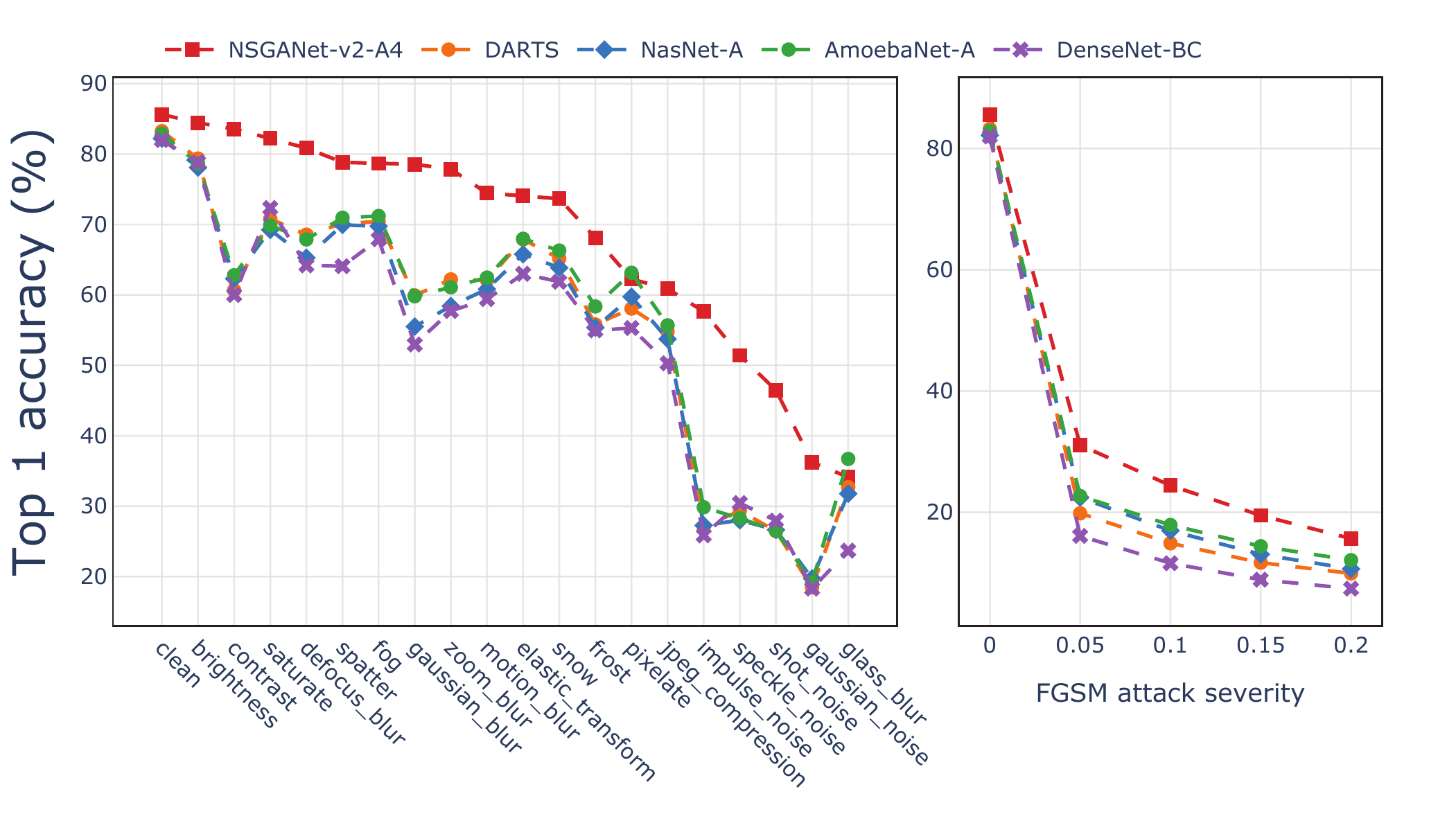}
		\caption{CIFAR-100\label{fig:cifar100_robustness_adv}}
	\end{subfigure}
	\caption{\textbf{Robustness:} Effect of commonly observable corruptions and adversarial attacks on (a) CIFAR-10, and (b) CIFAR-100. Higher values of $\epsilon$ indicate more severe adversarial attacks. {We use prediction accuracy on the corrupted test images (from each dataset) as a measurement of robustness.}
	\label{fig:robustness}\vspace{-0.3cm}}
\end{figure}

Our empirical findings summarized in Figs.~\ref{fig:cifar10_robustness_adv} and \ref{fig:cifar100_robustness_adv} appear to suggest that a positive correlation exists between the generalization performance on clean data and data under commonly observable corruptions -- i.e., we observe that \ourmethod{} architectures perform noticeably better than other peer methods' architectures on corrupted datasets even though the robustness measurement was never a part of the architecture selection process in \ourmethod{}. However, we emphasize that no architectures are considered robust to corruption, especially under adversarial attacks. We observe that the architectural advancements have translated to minuscule improvements in robustness against adversarial examples. The classification accuracy of all selected architectures deteriorates drastically with minor increments in adversarial attack severity $\epsilon$, leading to the question of whether architecture is the ``right'' ingredient to investigate in pursuit of adversarial robustness. A further step towards answering this question is provided in supplementary materials {under} Section~V-C.


\vspace{-2mm}
\subsection{Ablation Studies\label{sec:hyper-param-analysis}}
\begin{figure}[t]
	\centering
	\begin{subfigure}[t]{.24\textwidth}
		\centering
		\includegraphics[width=\textwidth]{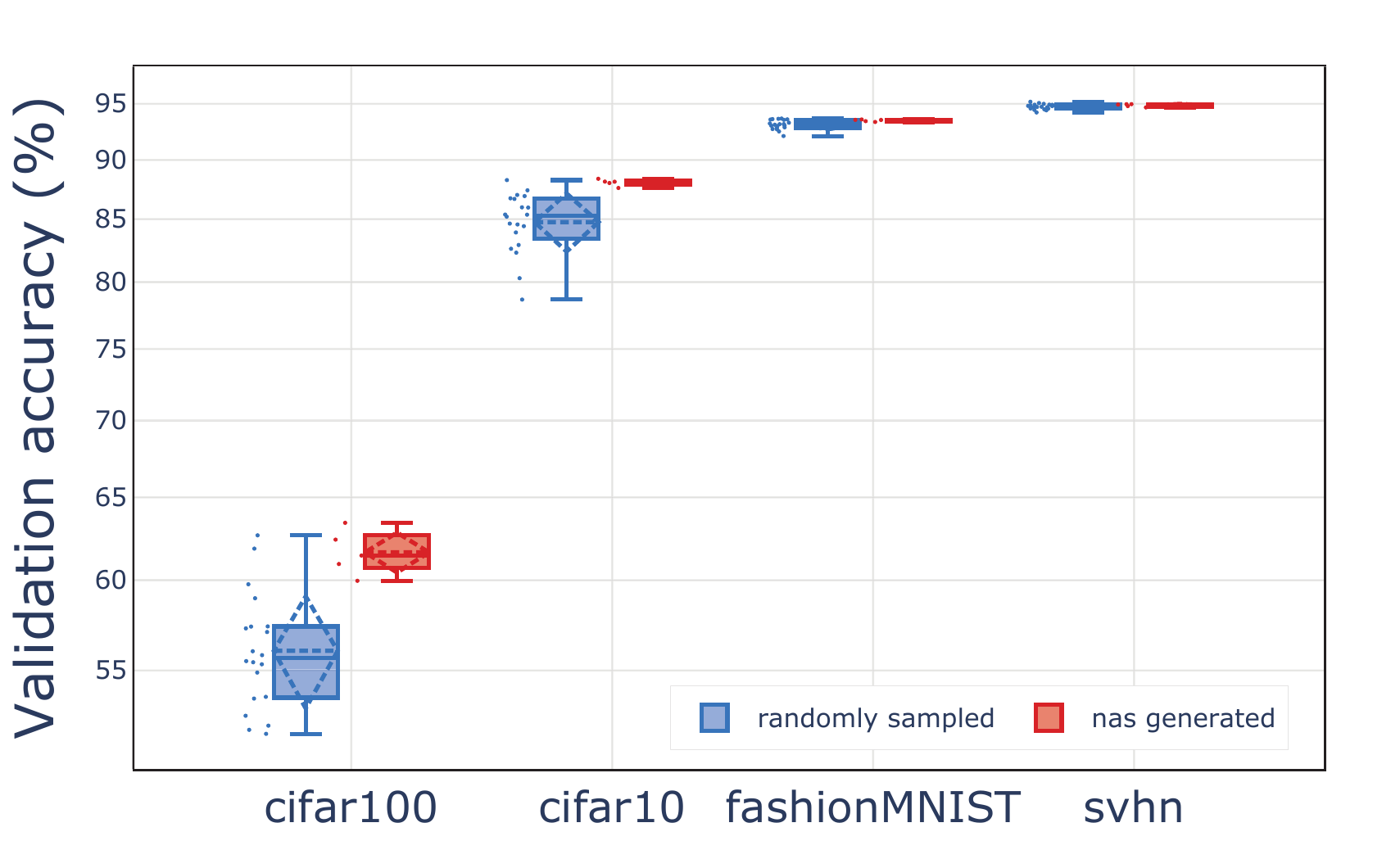}
		\caption{Effect of dataset choice.\label{fig:which_dataset_to_search}}
	\end{subfigure}
	\begin{subfigure}[t]{.24\textwidth}
		\centering
		\includegraphics[width=\textwidth]{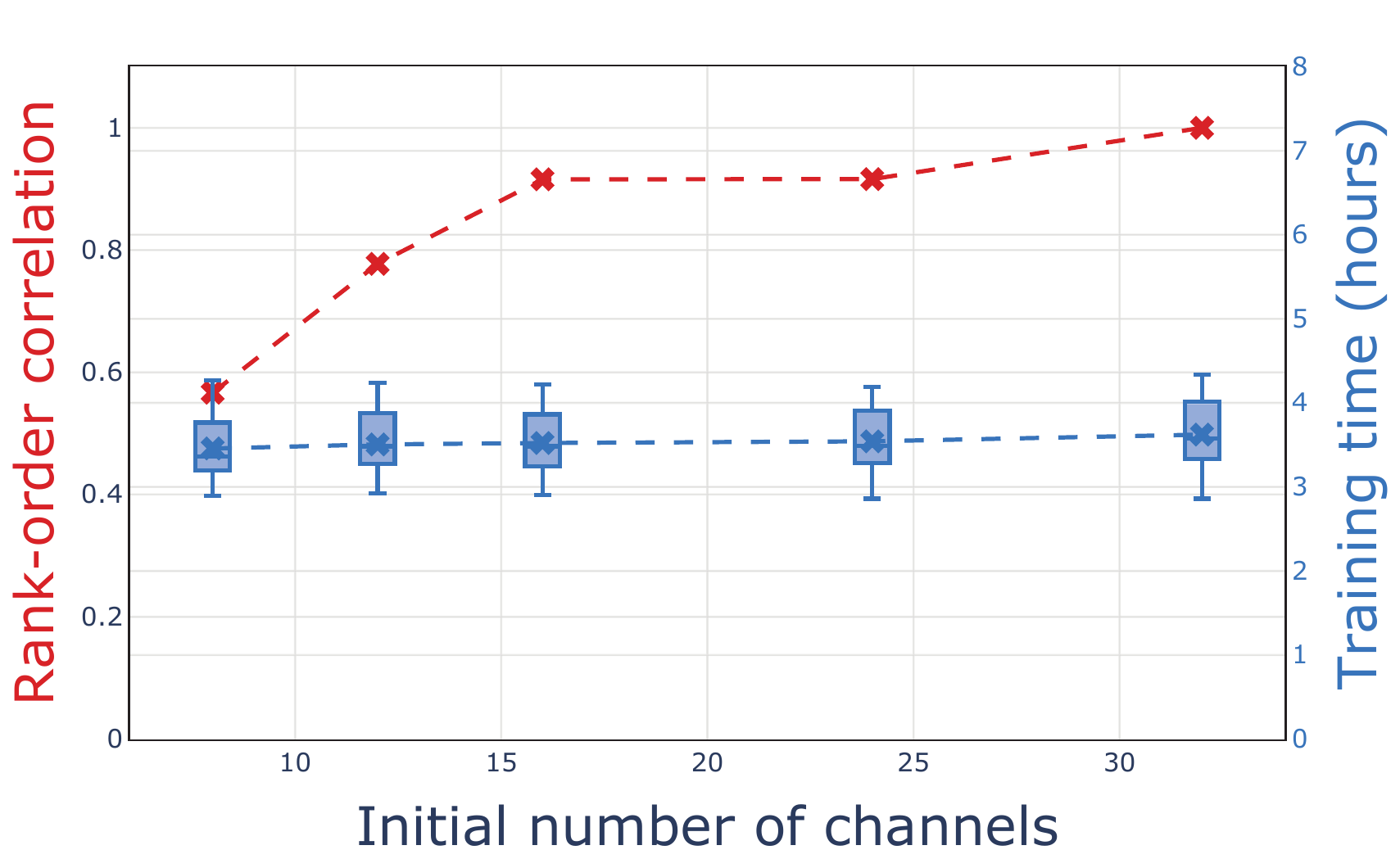}
		\caption{Effect of proxy model's width.\label{fig:proxy_width}}
	\end{subfigure}\\
	\begin{subfigure}[t]{.24\textwidth}
		\centering
		\includegraphics[width=\textwidth]{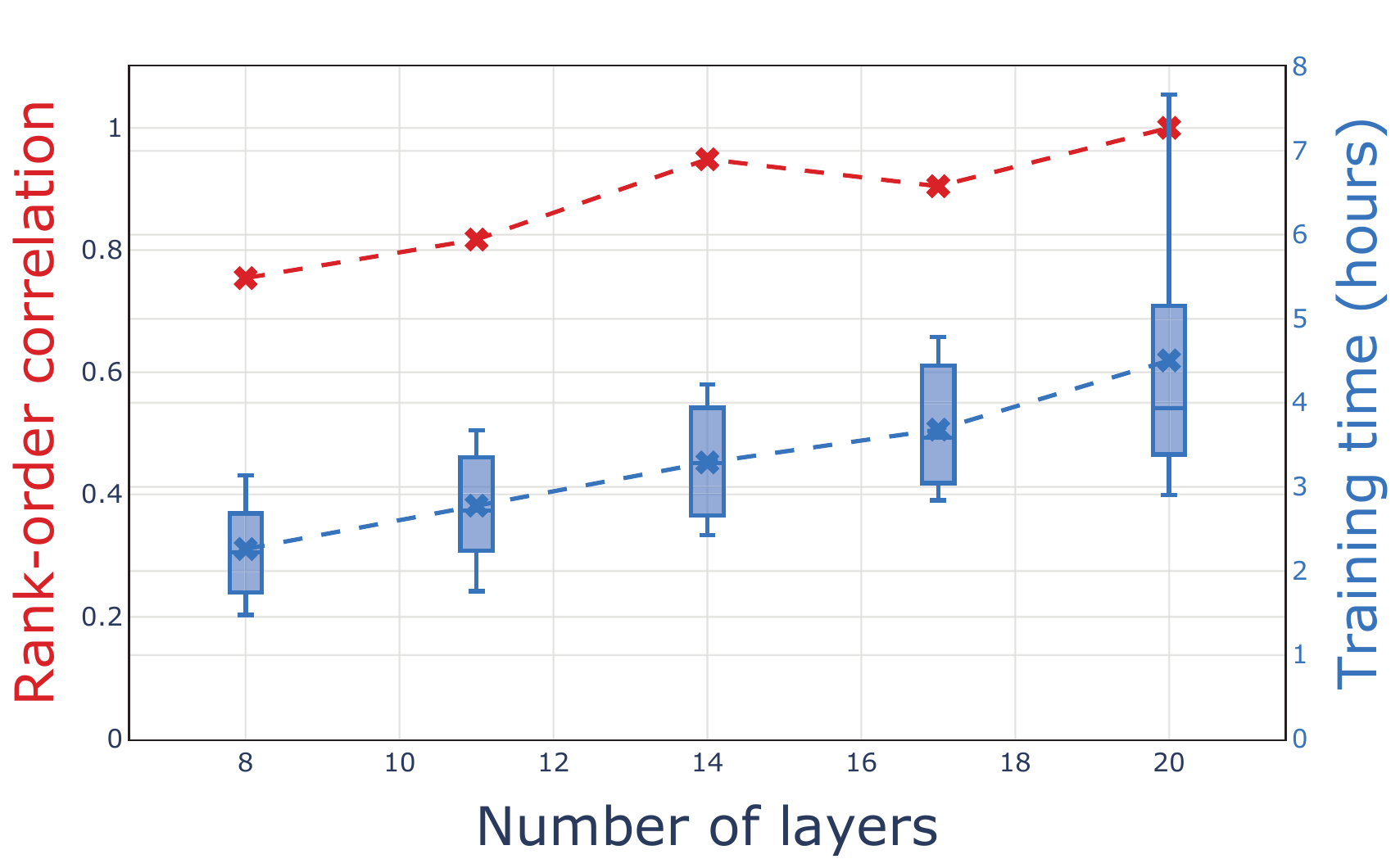}
		\caption{Effect of proxy model's depth.\label{fig:proxy_depth}}
	\end{subfigure}
	\begin{subfigure}[t]{.24\textwidth}
		\centering
		\includegraphics[width=\textwidth]{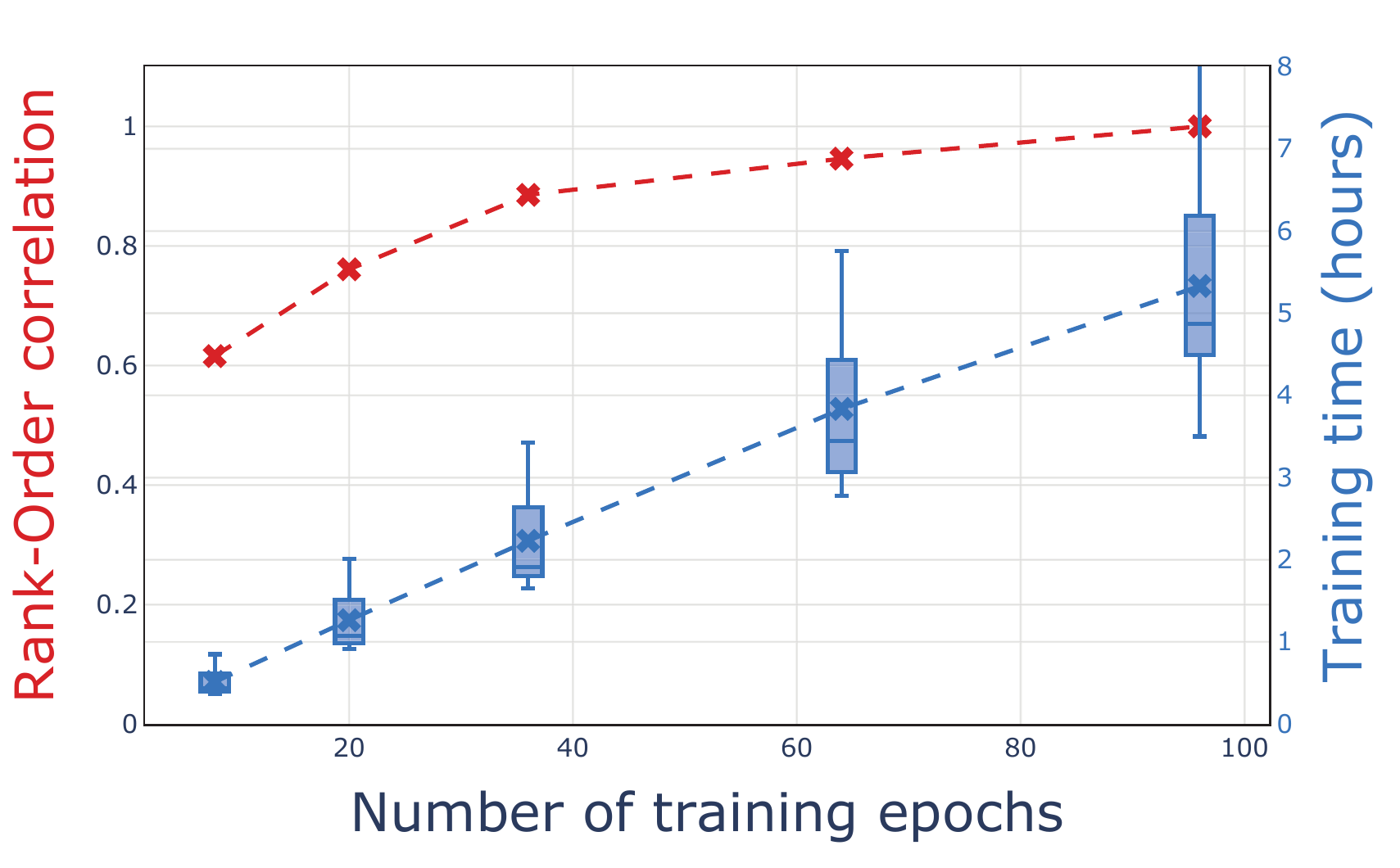}
		\caption{Effect of training epochs.\label{fig:proxy_epoch}}
	\end{subfigure}
	\caption{(a) Mean classification accuracy distribution of randomly generated architectures and architectures from peer NAS methods on four datasets. Correlation in performance (red lines) vs. Savings in gradient descent wall time (blue boxes) by reducing (b) the number of channels in layers, (c) the number of layers, and (d) the number of epochs to train. Note that (b), (c) and (d) have two y-axis labels corresponding to the color of the lines.
	\label{fig:proxy}\vspace{-0.3cm}}
\end{figure}

\vspace{3pt}
\noindent\textbf{Dataset for Search:} As previously mentioned in Section~\ref{sec:performance_estimation}, our proposed method differs from most of the existing peer methods in the choice of datasets on which the search is carried out. Instead of directly following the current practice of using the CIFAR-10 dataset, we investigated the utility of search on multiple benchmark datasets in terms of their ability to provide reliable estimates of classification accuracy and generalization. We carefully selected four datasets, SVHN \cite{netzer2011reading}, fashionMNIST \cite{xiao2017fashion}, CIFAR-10, and CIFAR-100 for comparison. The choice was based on a number of factors including the number of classes, numbers of training examples, resolutions and required training times. We uniformly sampled 40 architectures from the search space (described in Section~\ref{sec:approach}) along with five architectures generated by other peer NAS methods. We trained every architecture three times with different initial random seeds and report the averaged classification accuracy on each of the four datasets in Fig.~\ref{fig:which_dataset_to_search}. Empirically, we observe that the CIFAR-100 dataset is challenging enough for architectural differences to affect predicted performance. This can be observed in Fig.~\ref{fig:which_dataset_to_search} where the variation (blue boxes) in classification accuracy across architectures is noticeably larger on CIFAR-100 than on the other three datasets. In addition, we observe that mean differences in classification accuracy on CIFAR-100 between randomly generated architectures and architectures from principle-based methods have higher deviations, suggesting that it is less likely to find a good architecture on CIFAR-100 by chance.

\vspace{3pt}
\noindent\textbf{Proxy Models:} The main computational bottleneck of NAS approaches resides in evaluating the classification accuracy of the architectures by invoking the lower-level weight optimization. One such evaluation typically takes hours to finish, which limits the practical utility of the search under a constrained search budget. In our proposed algorithm, we adopt the concept of a proxy model \cite{nasnet2018,real2019regularized}. Proxy models are small-scale versions of the intended architectures, where the number of layers ($N$ in Fig.~\ref{fig:search_space_structure}) and the number of channels ($Ch_{init}$ in Fig.~\ref{fig:search_space_structure}) in each layer are reduced. Due to the downscaling, proxy models typically require much less compute time to train\footnote{Small architecture size allows larger batch size to be used and a lower number of epochs to converge under Algorithm~\ref{algo:back-prop}.}. However, there exists a trade-off between gains in computation efficiency and loss of  prediction accuracy. Therefore, it is not necessary that the performance of an architecture measured at proxy-model scale can serve as a reliable indicator of the architecture's performance measured at the desired scale.

To determine the smallest proxy model that can provide a reliable estimate of performance at a larger scale, we conducted parametric studies that gradually reduced the sizes of the proxy models of 100 randomly sampled architectures from our search space. Then, we measured the rank-order correlation and the savings in lower-level optimization compute time between the proxy models and the same architectures at the full scale. Figures \ref{fig:proxy_width}, \ref{fig:proxy_depth} and \ref{fig:proxy_epoch} show the effect of numbers of channels, layers and epochs, respectively, on the training time, and the Spearman rank-order correlation between the proxy and full scale models. We make the following observations, (1) increasing the number of channels does not significantly affect the wall clock time, and (2) reducing the number of layers or training epochs significantly reduces the wall clock time but also reduces the rank-order correlation. Based on these observations and the exact trade-offs from the plots, for our proxy model, we set the number of channels to 36 (maximum desired), number of epochs to 36, and number of layers to 14. Empirically, we found that this choice of parameters offers a good trade-off between practicality of search and reliability of proxy models.

%% file: application.tex
\section{An Application to Chest X-Ray Classification
\label{sec:appl}}
The ChestX-Ray14 benchmark was recently introduced in \cite{wang2017chestx}. The dataset contains 112,120 high resolution frontal-view chest X-ray images from 30,805 patients, and each image is labeled with one or multiple common thorax diseases, or ``Normal'', otherwise. More details are provided in supplementary materials under Section~V-C .
Past approaches \cite{wang2017chestx,yao2017learning,rajpurkar2017chexnet} typically extend from existing architectures, and the current state-of-the-art method \cite{rajpurkar2017chexnet} uses a variant of the DenseNet \cite{densenet} architecture, which is designed manually by human experts.
For reference purpose, we call the obtained architecture \ourmethod{}-X, and we re-train the weights thoroughly from scratch with an extended number of epochs. The learning rate is gradually reduced when the AUROC on the validation set plateaus.

\begin{table}[t]
\centering
\caption{AUROC on ChestX-Ray14 testing set.}
\label{tab:chexray}
\resizebox{0.45\textwidth}{!}{
\begin{threeparttable}
\begin{tabular}{@{ }lccc@{}}
\toprule
Method               & Type   & \#Params & Test AUROC (\%) \\ \midrule
Wang et al. (2017) \cite{wang2017chestx}  & manual & -        & 73.8            \\
Yao et al. (2017)  \cite{yao2017learning}  & manual & -        & 79.8            \\
CheXNet (2017) \cite{rajpurkar2017chexnet}      & manual & 7.0M    & 84.4            \\
Google AutoML (2018) \cite{blog2017automl} & RL     & -        & 79.7            \\
LEAF (2019) \cite{Liang-automl}         & EA     & -        & 84.3            \\ \midrule
\textbf{\ourmethod{}-A3}              & \textbf{EA}     & \textbf{5.0M}    & \textbf{84.7}            \\
\textbf{\ourmethod{}-X}              & \textbf{EA}     & \textbf{2.2M}    & \textbf{84.6}            \\ \bottomrule
\end{tabular}
\begin{tablenotes}
\scriptsize{
\item[\textdagger] Google AutoML result is from \cite{Liang-automl}.
\item[\textdaggerdbl] \ourmethod{}-A3 represents results under the standard transfer learning setup.
}
\end{tablenotes}
\end{threeparttable}
}
\end{table}

Table~\ref{tab:chexray} compares the performance of \ourmethod{}-X with peer methods that are extended from existing manually designed architectures. This includes architectures used by the authors who originally introduced the ChestX-Ray14 dataset \cite{wang2017chestx}, and the CheXNet \cite{rajpurkar2017chexnet}, which is the current \sota{} on this dataset. We also include results from commercial AutoML systems, i.e., Google AutoML \cite{blog2017automl}, and LEAF \cite{Liang-automl}, as comparisons with NAS-based methods. The setup details of these two AutoML systems are available in \cite{Liang-automl}. Noticeably, the performance of \ourmethod{}-X exceeds Google AutoML's by a large margin of nearly {\textbf{4 AUROC points}}. In addition, \ourmethod{}-X matches the \sota{} results from human engineered CheXNet, while using {\textbf{3.2x fewer parameters}}. For completeness, we also include the result from \ourmethod{}-A3, which is evolved on CIFAR-100, to demonstrate the transfer learning capabilities of \ourmethod{}.

\begin{figure}[t]
	\centering
	\begin{subfigure}[t]{.24\textwidth}
		\centering
		\includegraphics[width=0.98\textwidth]{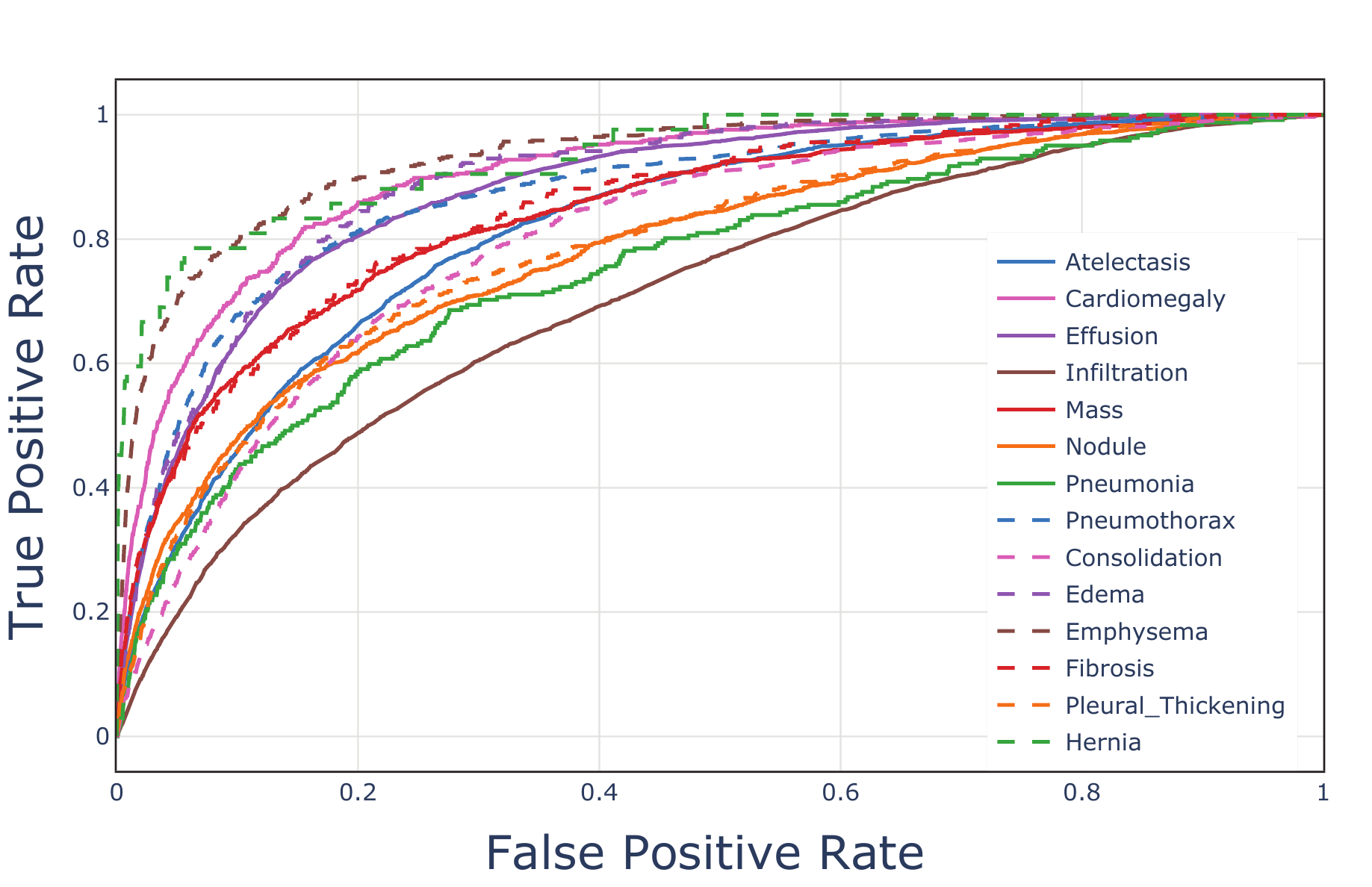}
		\caption{\label{fig:nsganet_x_ray}}
	\end{subfigure} \hfill
	\begin{subfigure}[t]{.24\textwidth}
		\centering
		\includegraphics[width=0.98\textwidth]{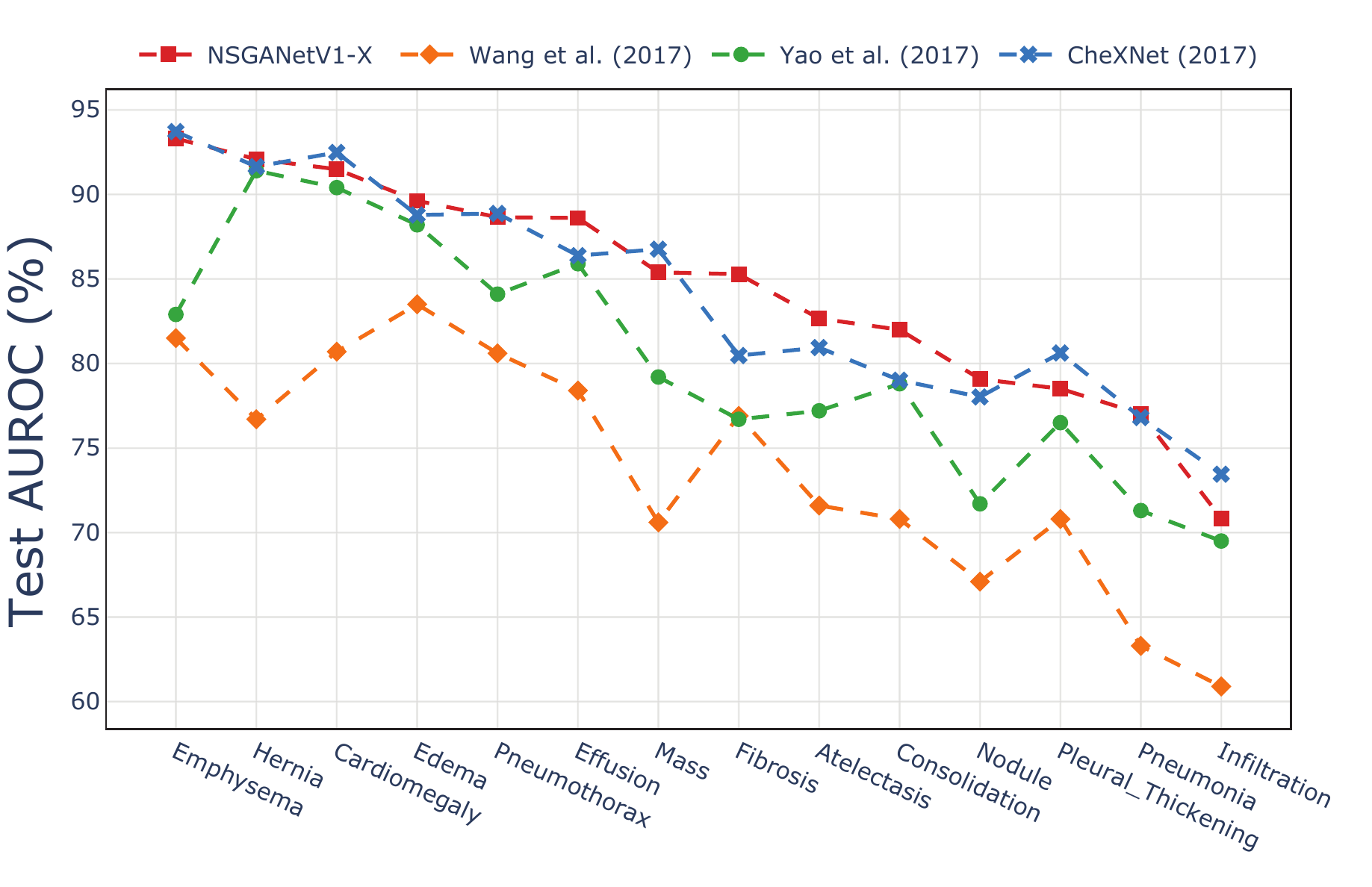}
		\caption{\label{fig:chexay14}}
	\end{subfigure}
	\caption{(a) \ourmethod{}-X multi-label classification performance on ChestX-Ray14 and (b) the class-wise mean test AUROC comparison with peer methods.
	\label{fig:nsganet_x}\vspace{-0.3cm}}
\end{figure}

\begin{figure}[t]
	\centering
	\begin{subfigure}[t]{.15\textwidth}
		\centering
		\includegraphics[width=0.95\textwidth]{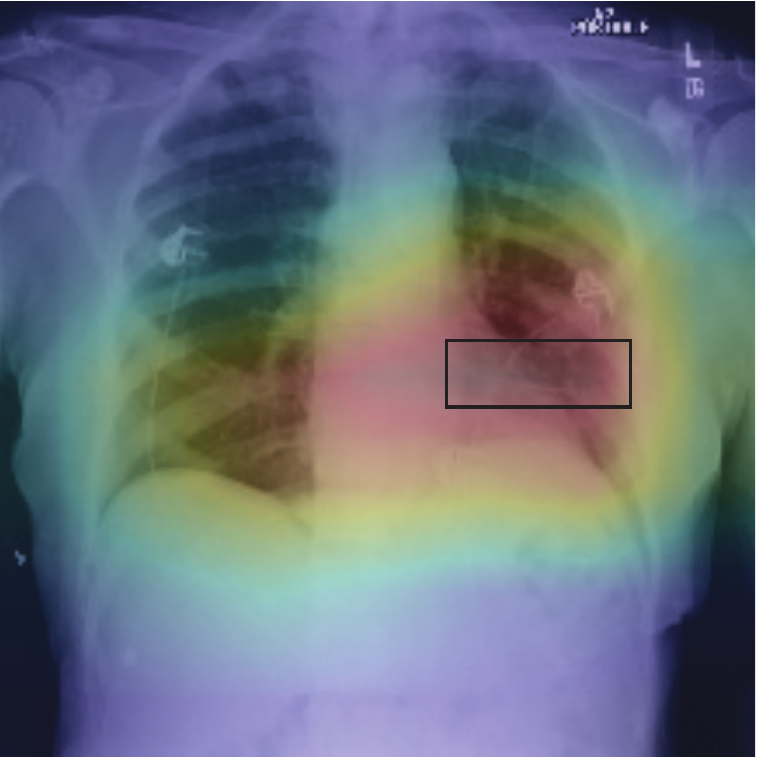}
		\caption{Atelectasis\label{fig:Atelectasis}}
	\end{subfigure}
	\begin{subfigure}[t]{.15\textwidth}
		\centering
		\includegraphics[width=0.95\textwidth]{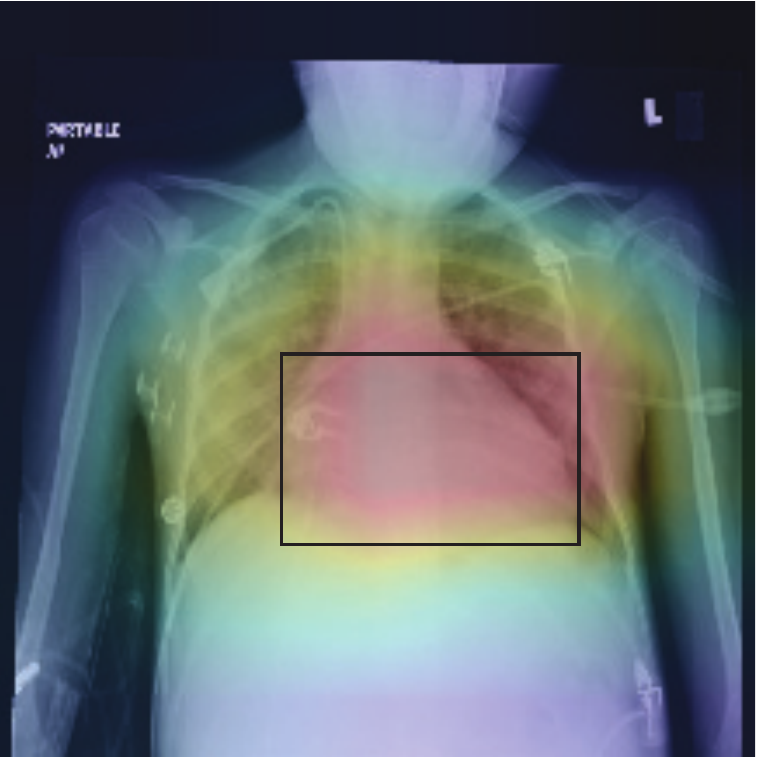}
		\caption{Cardiomegaly\label{fig:Cardiomegaly}}
	\end{subfigure}
	\begin{subfigure}[t]{.15\textwidth}
		\centering
		\includegraphics[width=0.95\textwidth]{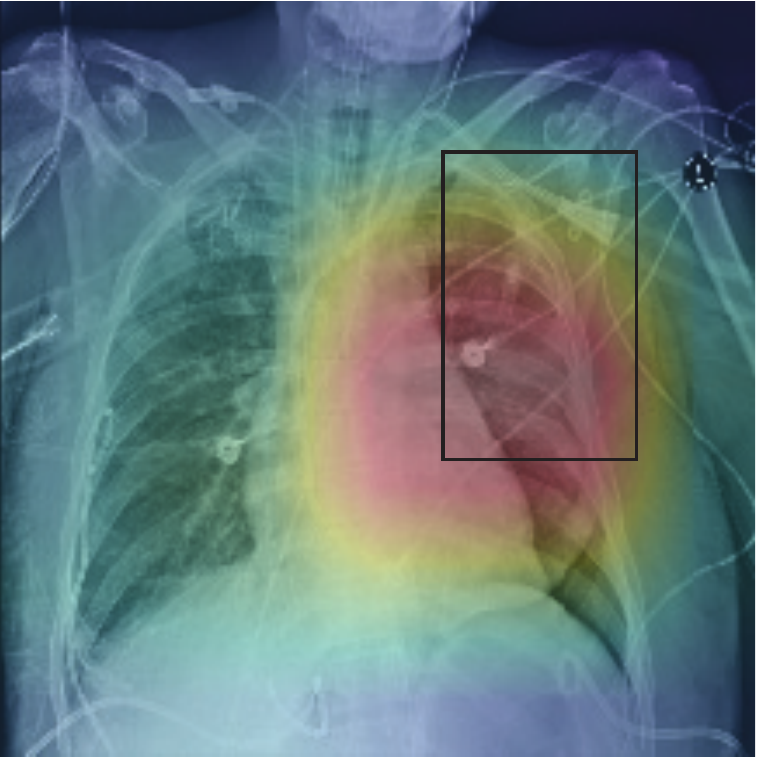}
		\caption{Effusion\label{fig:Effusion}}
	\end{subfigure}\\
	\begin{subfigure}[t]{.15\textwidth}
		\centering
		\includegraphics[width=0.95\textwidth]{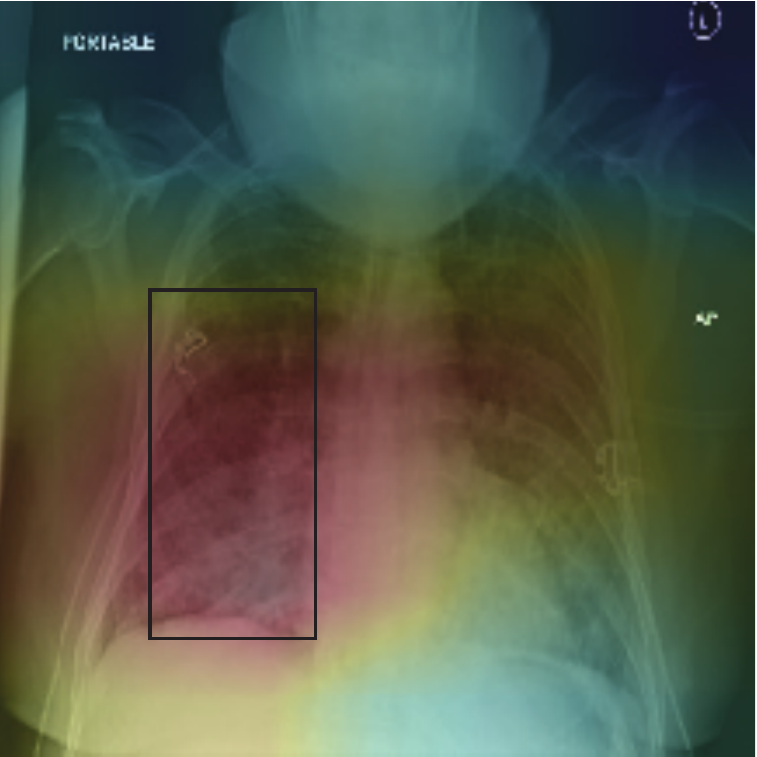}
		\caption{Infiltrate\label{fig:Infiltrate}}
	\end{subfigure}
	\begin{subfigure}[t]{.15\textwidth}
		\centering
		\includegraphics[width=0.95\textwidth]{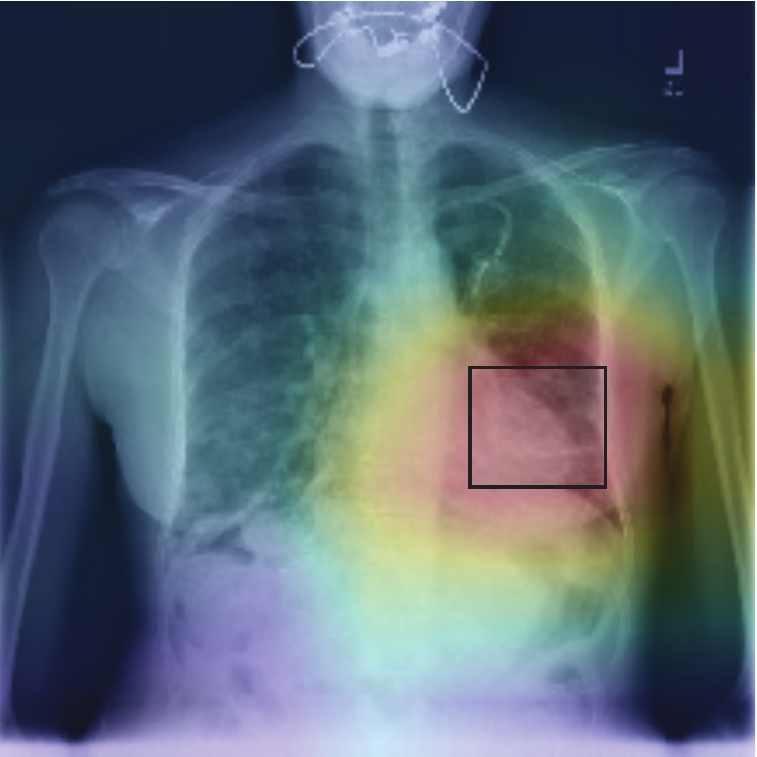}
		\caption{Pneumonia\label{fig:Pneumonia}}
	\end{subfigure}
	\begin{subfigure}[t]{.15\textwidth}
		\centering
		\includegraphics[width=0.95\textwidth]{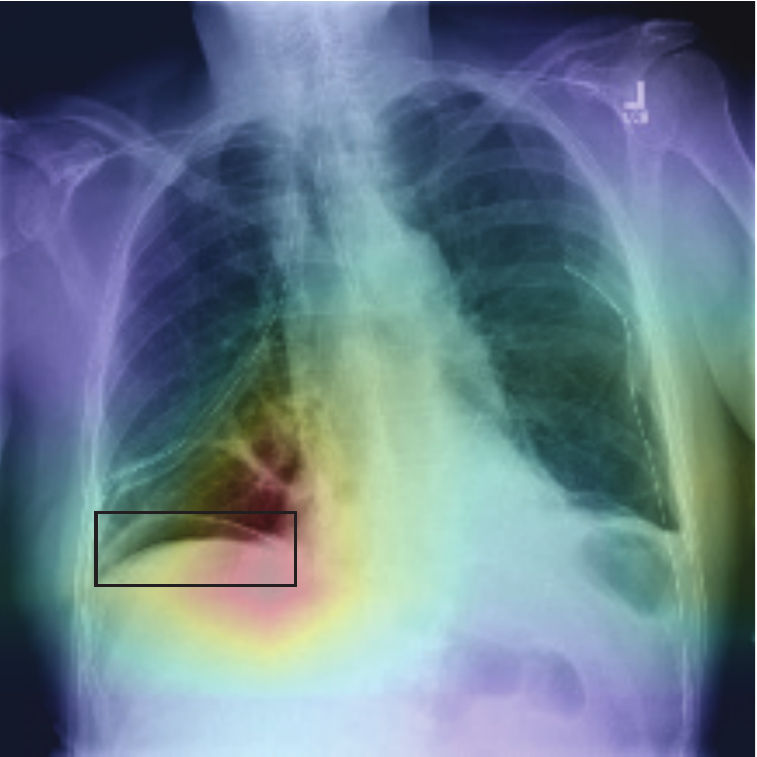}
		\caption{Pneumothorax\label{fig:Pneumothorax}}
	\end{subfigure}
	\caption{Examples of class activation map \cite{zhou2016learning} of \ourmethod{}-X, highlighting the class-specific discriminative regions. The ground truth bounding boxes are plotted over the heatmaps.
	\label{fig:x_ray_heatmaps}\vspace{-0.3cm}}
\end{figure}

More detailed results showing the disease-wise ROC curve of \ourmethod{}-X and disease-wise AUROC comparison with other peer methods are provided in Figs.~\ref{fig:nsganet_x_ray} and \ref{fig:chexay14}, respectively. To understand the pattern behind the disease classification decisions of \ourmethod{}-X, we visualize the class activation map (CAM) \cite{zhou2016learning}, which is commonly adopted for localizing the discriminative regions for image classification. In the examples shown in Fig.~\ref{fig:Atelectasis} - \ref{fig:Pneumothorax}, stronger CAM areas are covered with warmer colors. We also outline the bounding boxes provided by the ChestX-Ray14 dataset \cite{wang2017chestx} as references.

These results further validate the ability of our proposed algorithm to generate task-dependent architectures automatically. Conventional approaches, e.g., transfer learning from existing architectures, can be effective in yielding similar performance, however, as demonstrated by \ourmethod{}, simultaneously considering complexity along with performance in an algorithmic fashion allows architectures to be practically deployed in resource-constrained environments. {We observe this phenomenon in another application of \ourmethod{} to keypoint prediction on cars (see the supplementary materials under Section~V-D).}

%% file: conclusion.tex
\section{Conclusions\label{sec:conclusion}}
In this paper, we have presented \ourmethod{}, an evolutionary multi-objective algorithm for neural architecture search. \ourmethod{} explores the design space of architectures through recombining and mutating architectural components. \ourmethod{} further improves the search efficiency by exploiting the patterns among the past successful architectures via {distribution estimation through a Bayesian Network model.} Experiments on CIFAR-10, CIFAR-100, and ImageNet datasets have demonstrated the effectiveness of \ourmethod{}. Further analysis towards validating the generalization and robustness aspects of the obtained architectures is also provided along with an application to common thorax disease classification on human chest X-rays. We believe these results are encouraging {and demonstrate
the importance of customized and efficient evolutionary algorithms for neural architecture search in achieving superior performance compared to other contemporary machine learning methods.}

%% file: appendix.tex
In this supplementary document, we provide additional details on (1) related works in Section~\ref{sec:related_work_continued}; (2) multi-objective related issues in NAS in Section~\ref{sec:supple_mo}; (3) bilevel optimization in Section~\ref{sec:supple_bilevel}; (4) layer operations in Section~\ref{sec:operation_detail}; (5) datasets in Section~\ref{sec:dataset_detail}; (6) implementation in Section~\ref{sec:supple_train}; (7) other potential utilities of our proposed algorithm in Section~\ref{sec:utilities}; (8) hypervolume in Section~\ref{sec:supple_hypervolume}.

\subsection{Related Work Continued\label{sec:related_work_continued}}
{
Existing NAS approaches can be broadly classified into evolutionary algorithm (EA), reinforcement learning (RL), and relaxation-based approaches -- with a few additional methods falling outside these categories.

\noindent\textbf{Reinforcement Learning (RL):} $Q$-learning \cite{watkins1989qlearning} is a widely-used value iteration method used for RL. The MetaQNN method \cite{baker2017metaqnn} employs an $\epsilon$-greedy $Q$-learning strategy with experience replay to search connections between convolution, pooling, and fully connected layers, and the operations carried out inside the layers. Zhong \etal{}\cite{zhong2017blockqnn} extended this idea with the BlockQNN method. BlockQNN searches the design of a computational block with the same $Q$-learning approach. The block is then repeated to construct a network, resulting in a much more general network that achieves better results than its predecessor on CIFAR-10 \cite{cifar10}. A policy gradient method seeks to approximate non-differentiable reward functions to train a model that requires parameter gradients, like a neural network architecture. Zoph and Le \cite{zoph2016} first apply this method in NAS to train a recurrent neural network controller that constructs networks. The original method in \cite{zoph2016} uses the controller to generate the entire network at once. This contrasts with its successor, NASNet \cite{nasnet2018}, which designs a convolutional and pooling block that is repeated to construct a network. NASNet outperforms its predecessor and produces a network achieving state-of-the-art performance on CIFAR-10 and ImageNet. 

\vspace{3pt}
\noindent\textbf{Relaxation-based Approaches and Others:}
Approximating the connectivity between different layers in CNN architectures by real-valued variables weighting the importance of each layer is the common principle of relaxation-based NAS methods. Liu et al. first implement this idea in the DARTS algorithm \cite{liu2018darts}. DARTS seeks to improve search efficiency by fixing the weights while updating the architectures, showing convergence on both CIFAR-10 and Penn Treebank \cite{ptb} within one day in wall clock time on a single GPU card. Subsequent approaches in this line of research include \cite{Dong_2019_CVPR,xie2018snas,wu2019fbnet,liu2019auto}. The search efficiency of these approaches stems from weight sharing during the search process. This idea is complementary to our approach and can be incorporated into \ourmethod{} as well. However, it is beyond the scope of this paper and is a topic of future study.

Methods not covered by the EA-, RL- or relaxation-based paradigms have also shown success in architecture search. Liu et al. \cite{liu2018progressive} proposed a method that progressively expands networks from simple cells and only trains the best $K$ networks that are predicted to be promising by a RNN meta-model of the encoding space. PPP-Net \cite{dong2018ppp-net} extended this idea to use a multi-objective approach, selecting the $K$ networks based on their Pareto-optimality when compared to other networks. Li and Talwalkar \cite{li2019random} show that an augmented random search approach is an effective alternative to NAS. Kandasamy et al. \cite{kandasamy2018neural} present a Gaussian-process-based approach to optimize network architectures, viewing the process through a Bayesian optimization lens. 

\vspace{3pt}
\noindent\textbf{Multi-obj NAS through Scalarization:} A portfolio of works that aims to design hardware-specific network architectures emerges. This include, ProxylessNAS \cite{cai2018proxylessnas}, MnasNet \cite{mnasnet}, FBNet \cite{wu2019fbnet}, and MobileNetV3 \cite{mobilenetv3} which use a scalarized objective that encourages high accuracy and penalizes compute inefficiency at the same time, e.g., maximize $Acc * (Latency / Target)^{-0.07}$. These methods require a pre-defined preference weighting of the importance of different objectives before the search, which in itself requires a number of trials.

\vspace{3pt}
\noindent\textbf{Weight Sharing:} Another recently proposed approach for improving the search efficiency of NAS is through \emph{weight sharing}. Approaches in this category involve training a \emph{supernet} that contains all searchable architectures as its subnets. They can be broadly classified into two categories depending on whether the supernet training is coupled with architecture search or decoupled into a two-stage process. Approaches of the former kind \cite{pmlr-v80-pham18a,liu2018darts,cai2018proxylessnas} are computationally efficient but return sub-optimal models. Numerous studies \cite{li2019random,xie2019exploring,Yu2020Evaluating} allude to weak correlation between performance at the search and final evaluation stages. Methods of the latter kind \cite{brock2018smash,one-shot} use performance of subnets (obtained by sampling the trained supernet) as a metric to select architectures during search. However, training a supernet beforehand for each new task is computationally prohibitive.}

{
\subsection{Multi-objective Optimization in NAS\label{sec:supple_mo}}
In addition to high predictive accuracy, real-world applications demand NAS algorithms to simultaneously balance a few other network complexity related objectives that are specific to the deployment scenarios. For instance, mobile or embedded devices often have restrictions in terms of model size, multiply-adds, latency, power consumption, and memory footprint. 

It has been a common observation in the Deep Learning literature that classification performance is positively correlated with the complexity of the neural network. Since we want to maximize one (performance) while minimizing the other (FLOPS), they constitute a conflicting scenario. Optimization of a single composite objective obtained by weighting two objectives into one will produce a neural architecture and weight combination which may be too complex (requiring more FLOPS) or too inaccurate (having less accuracy). A generative approach of simply applying a single-obj optimization does not solve the issue: (i) many common scalarization methods do not work if the interesting optimal solutions lie on the non-convex part of the efficient frontier, and (ii) Generative methods are more computationally expensive (due to the lack of any parallel search efforts) than simultaneous methods, such as the method used in this paper. 
    
In particular, ResNet \cite{resnet} showed the classification accuracy on ImageNet continuously to improve as the number of layers increases from 18 (2G FLOPs) to 152 (11G FLOPs). Similar trends are also observed in DenseNet \cite{densenet}, NASNet \cite{nasnet2018}, EfficientNet \cite{tan2019efficientnet}, etc. The aforementioned observation implies the competing nature of these objectives of simultaneously maximizing classification performance and minimizing complexity in terms of FLOPs. Additionally, posing NAS as a multi-objective problem is beneficial from the decision-making perspective, as it allows designers to choose a suitable network architecture \emph{a posteriori} as opposed to requiring a pre-defined preference weighting of each objective prior to the search. Empirically, we also observe that the type of diversity provided by multi-objective optimization contributes to its outperforming on the classification accuracy objective achieved relative to single-objective optimization. (This can be seen, for example, in comparing \ourmethod{}-Ax from the main paper and \ourmethod{}-Bx from Table~\ref{tab:nsganet-early} in this supplementary materials.
}

\subsection{Bilevel Optimization in NAS\label{sec:supple_bilevel}}
Recall that we formulate the problem of designing custom architectures for different deployment scenarios as a bilevel multi-objective NAS problem, mathematically as below:
\begin{equation}
\begin{aligned}
\Minimize & \hspace{3mm} \bm{F}(\bm{x}) = \big(f_1(\bm{x}; \bm{w}^*(\bm{x})), f_2(\bm{x})\big)^T, \\
\st  & \hspace{3mm} \bm{w}^*(\bm{x}) \in \argmin~\mathcal{L}(\bm{w};\bm{x}), \\
     & \hspace{3mm} \bm{x} \in \mathbf{\Omega}_{x}, \hspace{3mm} \bm{w} \in \mathbf{\Omega}_{w},
\end{aligned}
\label{def:bi_obj_nas}
\end{equation}

{
The bilevel formulation used above arises from the problem nature of NAS, where one objective function $f_1$ evaluation at the upper-level requires both an architecture $\bm{x}$ and its weights $\bm{w}$. $f_1$ is \emph{not} meaningfully defined at any arbitrary $\bm{w}$; rather, it requires $\bm{w}$ to be a member of the set that minimize the cross-entropy loss $\mathcal{L}$ (in the case of image classification) on training data given $\bm{x}$, mathematically as $\bm{w}^*(\bm{x}) \in \mbox{argmin}~\mathcal{L}(\bm{w}; \bm{x})$. For simplicity, we use $\bm{w}^*(\bm{x})$ to denote weights that satisfy the previously specified condition. However, $\bm{w}^*(\bm{x})$ is typically not analytically computable due to non-linearities in layers and from activation functions in encoded by $\bm{x}$, requiring another (lower) level of optimization. 

The principle of a bilevel optimization is that the upper-level objectives and constraints must be computed using the optimal lower level variables ($\bm{w}^*(\bm{x})$) for the corresponding upper-level variables ($\bm{x}$). Thus, in an implicit manner, $f_1$ becomes a function of $\bm{x}$ alone, making $f_1(\bm{x}; \bm{w}^*(\bm{x}))$. Using a $(1, 1)$ evolution strategy search as an illustration, when $\bm{x}_{t+1} \leftarrow \bm{x}_{t}+\Delta\bm{x}$ is altered at iteration $t$ in the upper-level, we first optimize the corresponding lower-level problem of $\mathcal{L}(\bm{w}; \bm{x}_{t+1})$ to find $\bm{w}^*(\bm{x}_{t+1})$ (through SGD). We then compute $f_1(\bm{x}_{t+1}, \bm{w}^*(\bm{x}_{t+1}))$, and update the current $\bm{x}_t \leftarrow \bm{x}_{t+1}$ if $f_1\big(\bm{x}_{t+1}, \bm{w}^*(\bm{x}_{t+1})\big)$ is better than $f_1\big(\bm{x}_{t}, \bm{w}^*(\bm{x}_{t})\big)$, i.e., $f_1\big(\bm{x}_{t+1}, \bm{w}^*(\bm{x}_{t+1})\big)$ $<$ $f_1\big(\bm{x}_{t}, \bm{w}^*(\bm{x}_{t})\big)$. A bilevel problem allows a more convenient way to optimize a hierarchical problem having two distinct hierarchical variable sets (such as, a weight vector only makes sense when an architecture is provided) than a single level in which both $\bm{x}$ and $\bm{w}$ are considered in the same level. First, the search space becomes huge and second, a good $\bm{x}$ may be deleted simply because the respective $\bm{w}$ is not good.}

Our \ourmethod{} algorithm has two types of optimization problems put together:
\begin{enumerate}
    \item A bilevel optimization having a upper-level optimization problems in which architectures ($\bm{x}$) are decision variables and a lower level problem in which weight vector ($\bm{w}$) for the given architecture is the decision variable. 
    \item upper-level problem uses two conflicting objectives providing a Pareto-optimal front and the lower level problem uses a single objective of minimizing the cross-entropy loss on the validation data. 
\end{enumerate}

Thus, the final outcome of our approach is a set of trade-off architectures and their associated weight vectors, thereby completely specifying neural networks. For upper-level, we employ a customized and advanced NSGA-II-like evol. multi-objective algorithm, so that a set of non-dominated trade-off solutions is obtained at each iteration. The lower level uses the stochastic gradient based back-propagation algorithm for weight learning. {Despite the fact that our approach also hybridizes a global search (EMO algorithm at the upper-level) and a local search (SGD) into a unified paradigm, which is also done in memetic algorithms, our bilevel approach is conceptually very different from memetic computing. To be more specific, the local search used in memetic algorithms is mainly to improve an individual's fitness within its local neighborhood, while the local search (SGD) used in our approach is to find the remaining set of variables (lower-level variables; weights), which jointly with the upper-level variables (architectures) compute the objective function (classification error).}

Our bilevel approach is nested in nature, meaning that for each $\bm{x}$ at the upper-level, a respective optimized $\bm{w}^{\ast}$ is found by using the back-propagation method. However, the NAS at the upper-level is expedited by using a Bayesian learning method of already found good solutions and by using customized coding and genetic operators. Although more sophisticated surrogate-assisted bilevel algorithms, such as BLEAQ or BLEAQ2  \cite{bilevel-cma,bleaq} can be used, in this work we keep the methods relatively simple and use learning-assisted EMO and use only 1,200 architecture evaluations to achieve the results.

\subsection{Details of the Considered Layer Operations\label{sec:operation_detail}}
As described in Section III-A, we form a operation pool consists of 12 different choices of convolution, pooling and etc., based on their prevalence in the CNN literature. Most of these operations can be directly called from standard Deep Learning libraries, like Pytorch, TensorFLow, Caffe, etc. Here we provide demo Pytorch codes for less commonly \emph{used}\footnote{refer to both less frequently used operations and operations under less commonly followed setups.} operations, including \emph{depth-wise separable convolutions}, \emph{local binary convolutions} and \emph{1{\rm x}7 then 7{\rm x}1 convolution}.

\begin{figure*}[!htbp]
  \centering{
  \parbox{\textwidth}{
    \parbox[b]{.2\textwidth}{%
      \subcaptionbox{\ourmethod{}-A1\label{fig:arch2}}{\includegraphics[width=\hsize, height=6.2cm]{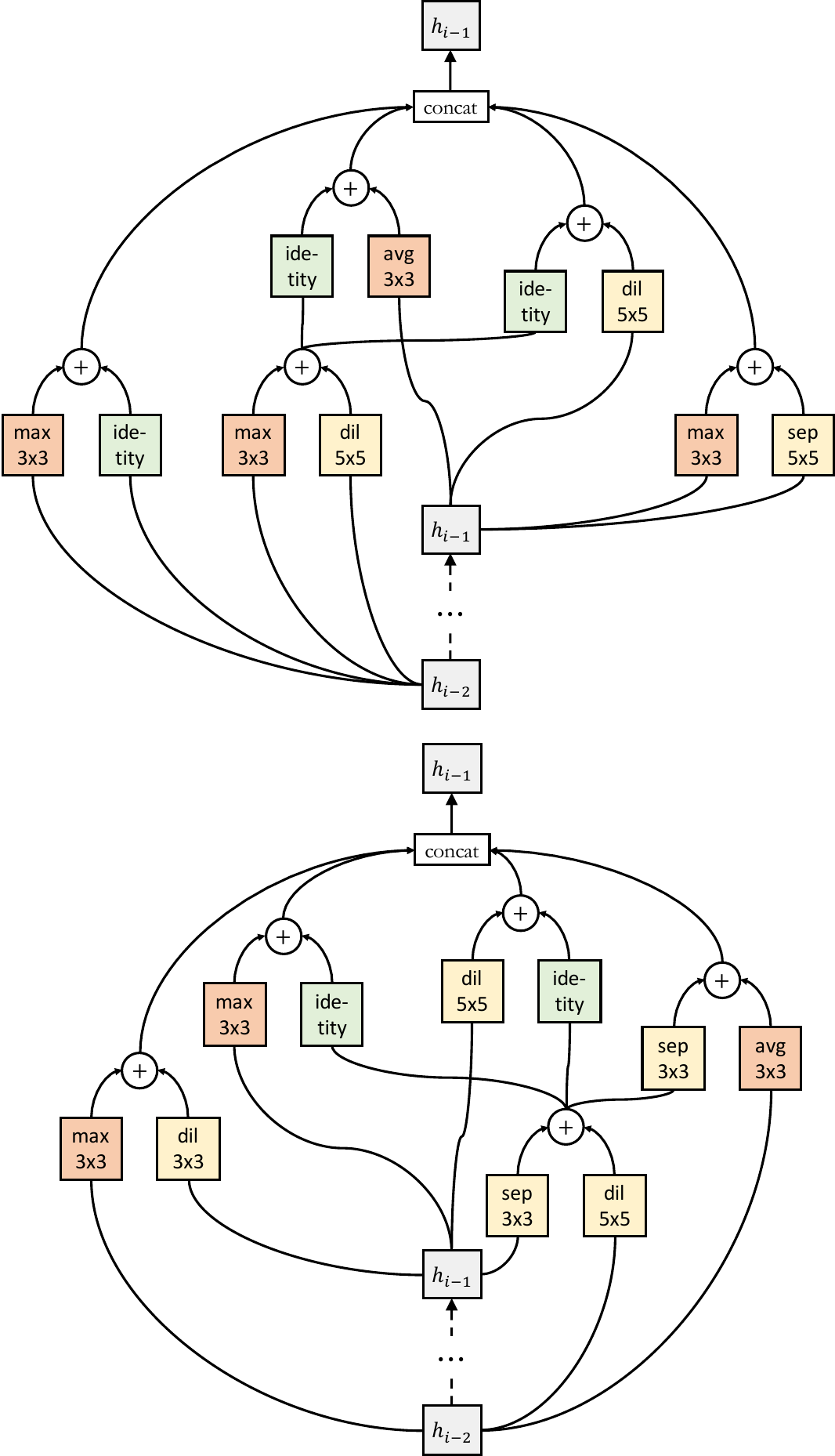}}
    }
    \hskip1em
    \parbox[b]{.2\textwidth}{%
      \subcaptionbox{\ourmethod{}-A2\label{fig:arch1}}{\includegraphics[width=\hsize, height=6.2cm]{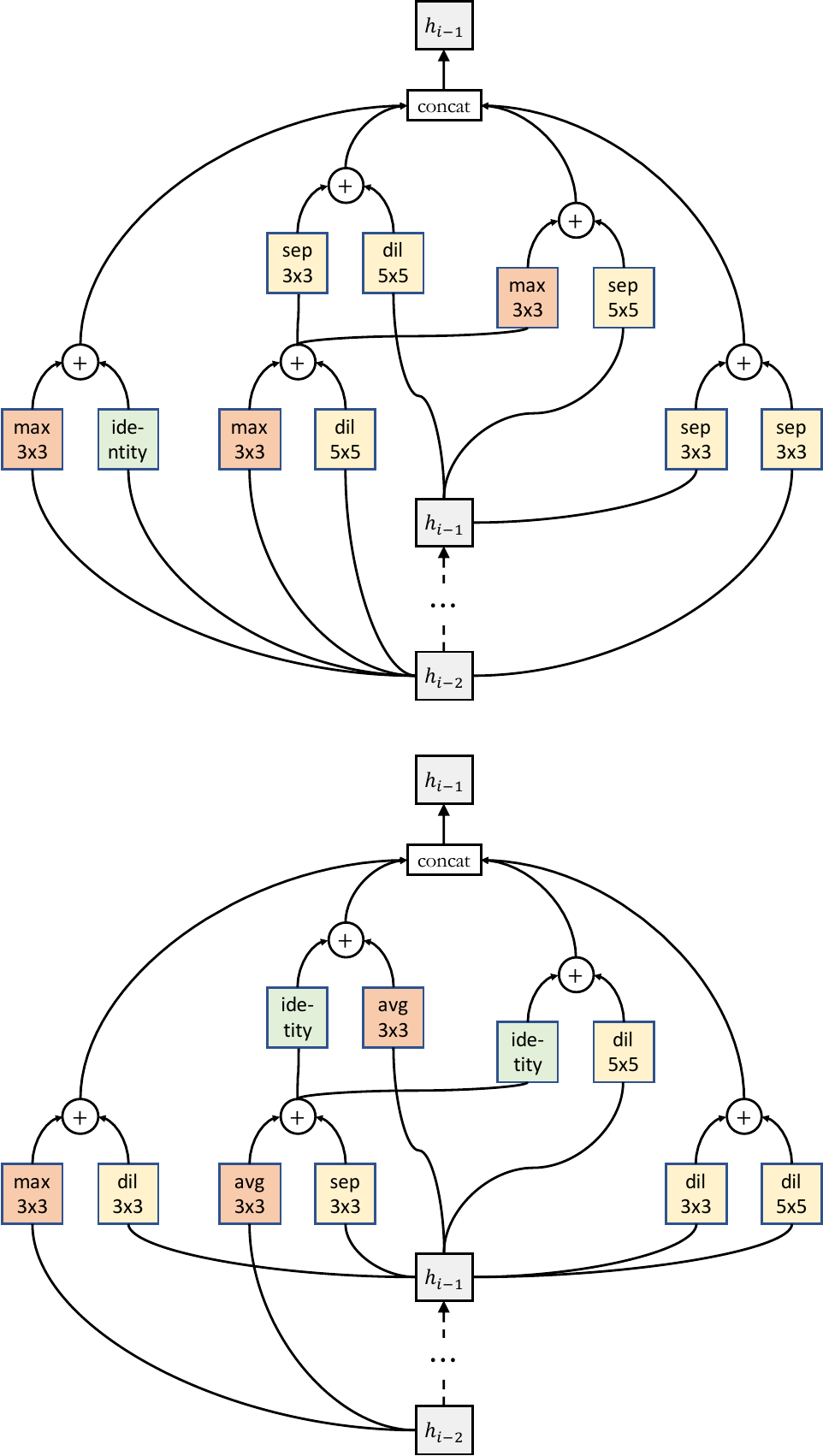}}
    }
    \hskip1em
    \parbox[b]{.2\textwidth}{%
      \subcaptionbox{\ourmethod{}-A4\label{fig:arch3}}{\includegraphics[width=\hsize, height=6.2cm]{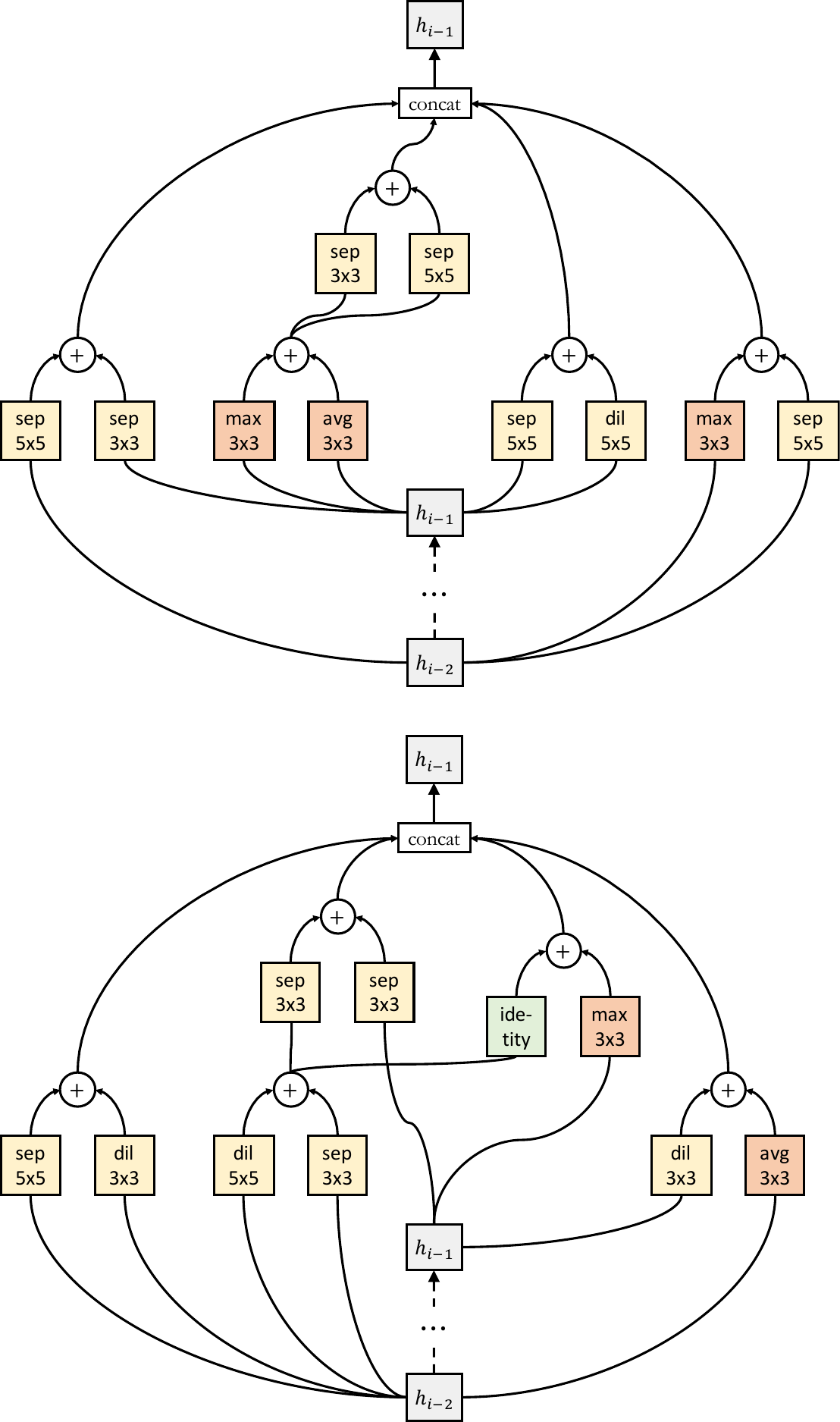}}
    }
    \hskip1em
    \parbox[b]{.32\textwidth}{%
      \begin{minipage}[b]{0.12\textwidth}
      \subcaptionbox{ResNet \cite{resnet}\label{fig:ResNet}}{\includegraphics[width=\hsize]{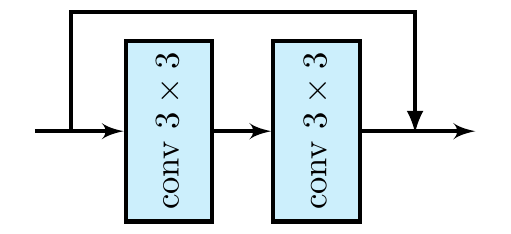}}
      \end{minipage}\hfill
      \begin{minipage}[b]{0.15\textwidth}
      \subcaptionbox{MobileNetV2 \cite{sandler2018mobilenetv2} \label{fig:mobilenetV2}}{\includegraphics[width=\hsize]{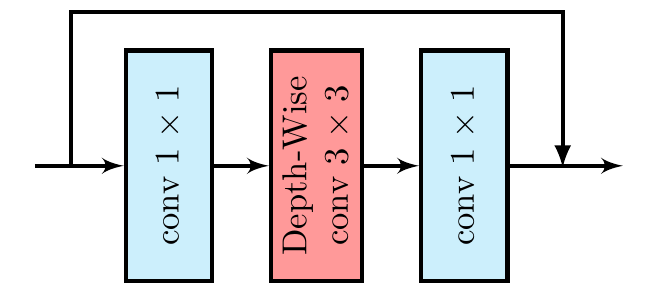}}
      \end{minipage}
      \vskip1em
      \subcaptionbox{NASNet-A \cite{nasnet2018}\label{fig:NasNet-A}}{\includegraphics[width=\hsize, height=3.2cm]{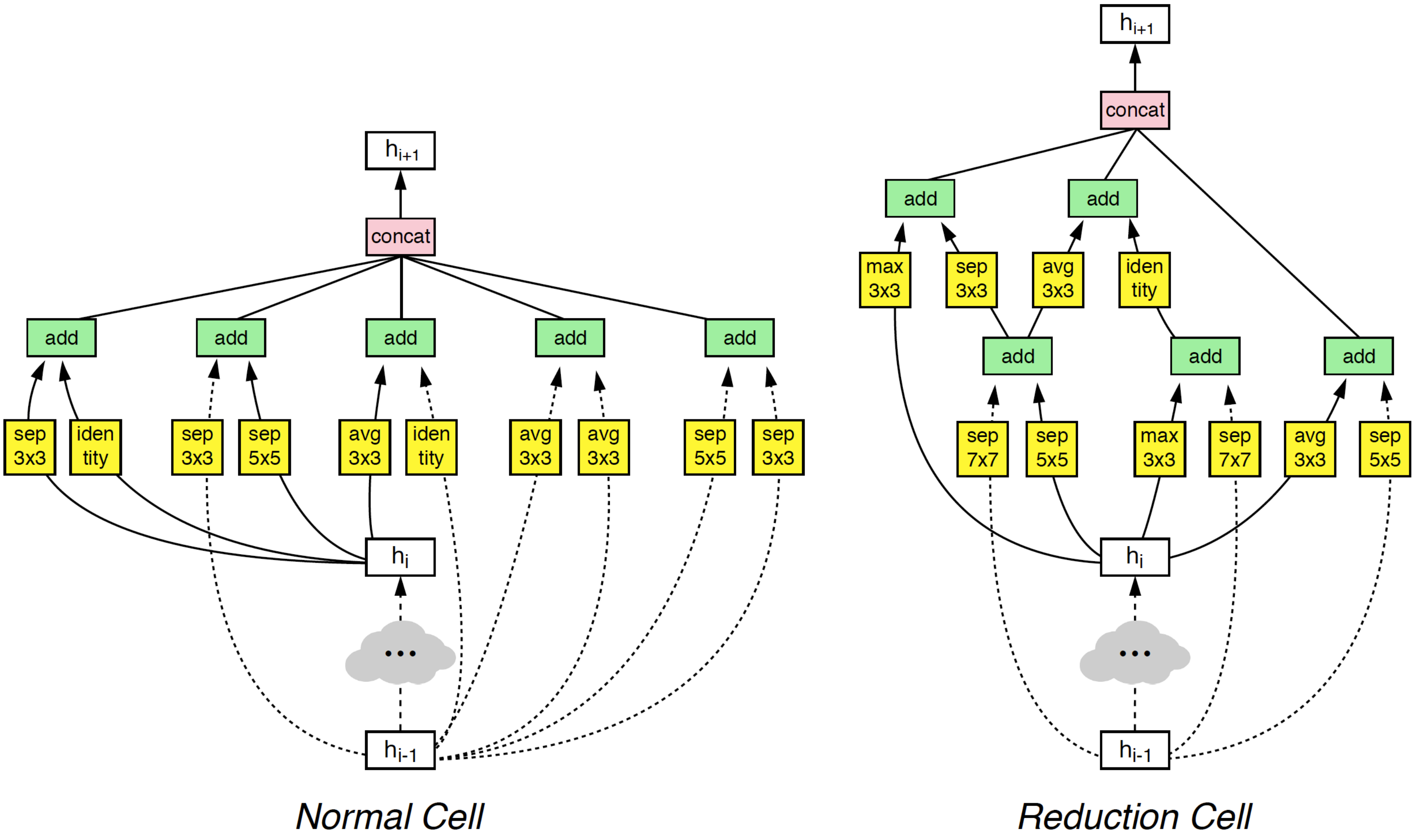}}
    }
  }}
  \caption{Visualization of block-level structures for different architectures. The Normal and Reduction blocks are shown in the first and second rows, respectively for \ourmethod{} architectures. Examples of blocks that are designed manually by experts \cite{resnet,sandler2018mobilenetv2} and from other peer methods \cite{nasnet2018} are also included in (d) - (f) for comparison. \label{fig:archs_visualization}\vspace{-0.3cm}}
\end{figure*}

\begin{table}[!htbp]
\centering
\caption{Demo Pytorch implementation of separable convolution (a), local binary convolution (b) and 1x7 then 7x1 convolution (c) used in \ourmethod{}.\label{tab:pytorch_implementation}}
\begin{subtable}[h]{0.48\textwidth}
    \centering
    \resizebox{0.98\textwidth}{!}{%
    \begin{tabular}{@{ }ll@{ }}
    \toprule
    class SepConv(nn.Module) &  \\
    {\color{gray} \# depth-wise separable convolution in \ourmethod{}} & \\
    {\color{gray} \# consists of two regular depth-wise separable convolutions in series.} & \\
    def \_\_init\_\_(self, C\_in, C\_out, kernel\_size, stride, padding, affine=True) &  \\
    \hspace{5mm} super(SepConv, self).\_\_init\_\_() &  \\
    \hspace{5mm} self.op = nn.Sequential( &  \\
    \hspace{10mm} nn.ReLU(inplace=False), &  \\
    \hspace{10mm} nn.Conv2d(C\_in, C\_in, kernel\_size=kernel\_size, stride=stride, & \\
    \hspace{23mm} padding=padding, groups=C\_in, bias=False), &  \\
    \hspace{10mm} nn.Conv2d(C\_in, C\_in, kernel\_size=1, padding=0, bias=False), &  \\
    \hspace{10mm} nn.BatchNorm2d(C\_in, affine=affine), &  \\
    \hspace{10mm} nn.ReLU(inplace=False), &  \\
    \hspace{10mm} nn.Conv2d(C\_in, C\_in, kernel\_size=kernel\_size, stride=1, & \\
    \hspace{23mm} padding=padding, groups=C\_in, bias=False), &  \\
    \hspace{10mm} nn.Conv2d(C\_in, C\_out, kernel\_size=1, padding=0, bias=False), &  \\
    \hspace{10mm} nn.BatchNorm2d(C\_out, affine=affine), &  \\
    \hspace{5mm}) &  \\
    \hspace{5mm} def forward(self, x) &  \\
    \hspace{10mm} return self.op(x) &  \\ \bottomrule
    \end{tabular}%
    }
    \vspace{2mm} \caption{Separable Convolution\label{tab:sep_conv}}
\end{subtable}
\begin{subtable}[h]{0.48\textwidth}
    \centering
    \resizebox{0.98\textwidth}{!}{%
    \begin{tabular}{@{ }ll@{ }}
    \toprule
    {\color{gray} \# The weight values of local binary convolution filters} & \\
    {\color{gray} \# either -1, 1, or 0, and kept fixed during back-propagation.} & \\
    {\color{gray} \# Number of 0-valued weights are controlled by \emph{sparsity} argument.} & \\
    def LBConv(in\_planes, out\_planes, kernel\_size=3, stride=1, &  \\
    \hspace{5mm} padding=1, dilation=1, groups=1, bias=False, sparsity=0.5) &  \\
    \hspace{5mm} conv2d = nn.Conv2d( &  \\
    \hspace{10mm} in\_planes, out\_planes, kernel\_size=kernel\_size, &  \\
    \hspace{10mm} stride=stride, padding=padding, dilation=dilation, & \\
    \hspace{10mm} groups=groups, bias=bias, &  \\
    \hspace{5mm}) &  \\
    \hspace{5mm} conv2d.weight.requires\_grad = False &  \\
    \hspace{5mm} conv2d.weight.fill\_(0.0) &  \\
    \hspace{5mm} num = conv2d.weight.numel() &  \\
    \hspace{5mm} shape = conv2d.weight.size() &  \\
    \hspace{5mm} index = torch.Tensor(math.floor(sparsity * num)).random\_(num).int() &  \\
    \hspace{5mm} conv2d.weight.resize\_(num) &  \\
    \hspace{5mm} for i in range(index.numel()) &  \\
    \hspace{10mm} conv2d.weight{[}index{[}i{]}{]} = torch.bernoulli(torch.Tensor({[}0.5{]})) * 2 - 1 &  \\
    \hspace{10mm} conv2d.weight.resize\_(shape) &  \\
    \hspace{5mm} return conv2d &  \\ \bottomrule
    \end{tabular}%
    }
    \vspace{2mm} \caption{Local Binary Convolution\label{tab:lbcnn}}
\end{subtable}
\begin{subtable}[h]{0.48\textwidth}
    \centering
    \resizebox{0.98\textwidth}{!}{%
    \begin{tabular}{@{ }ll@{ }}
    \toprule
    class Conv1x7Then7x1 &  \\
    def \_\_init\_\_(self, C, stride, affine=True) &  \\
    \hspace{5mm} super(Conv1x7Then7x1, self).\_\_init\_\_() &  \\
    \hspace{5mm} self.op = nn.Sequential( &  \\
    \hspace{10mm} nn.ReLU(inplace=False), &  \\
    \hspace{10mm} nn.Conv2d(C, C, (1, 7), stride=(1, stride), padding=(0, 3), bias=False), &  \\
    \hspace{10mm} nn.Conv2d(C, C, (7, 1), stride=(stride, 1), padding=(3, 0), bias=False), &  \\
    \hspace{10mm} nn.BatchNorm2d(C, affine=affine) &  \\
    \hspace{5mm} ) &  \\
    \hspace{5mm} def forward(self, x) &  \\
    \hspace{10mm} return self.op(x) &  \\ \bottomrule
    \end{tabular}%
    }
    \vspace{2mm}\caption{1x7 convolution then 7x1 convolution \label{tab:1x7_then_7x1}}
\end{subtable}
\end{table}

\subsection{Datasets Details\label{sec:dataset_detail}}
\begin{figure}[t]
    \vspace{-2mm}
	\centering
	\begin{subfigure}[t]{.24\textwidth}
		\centering
		\includegraphics[width=0.95\textwidth]{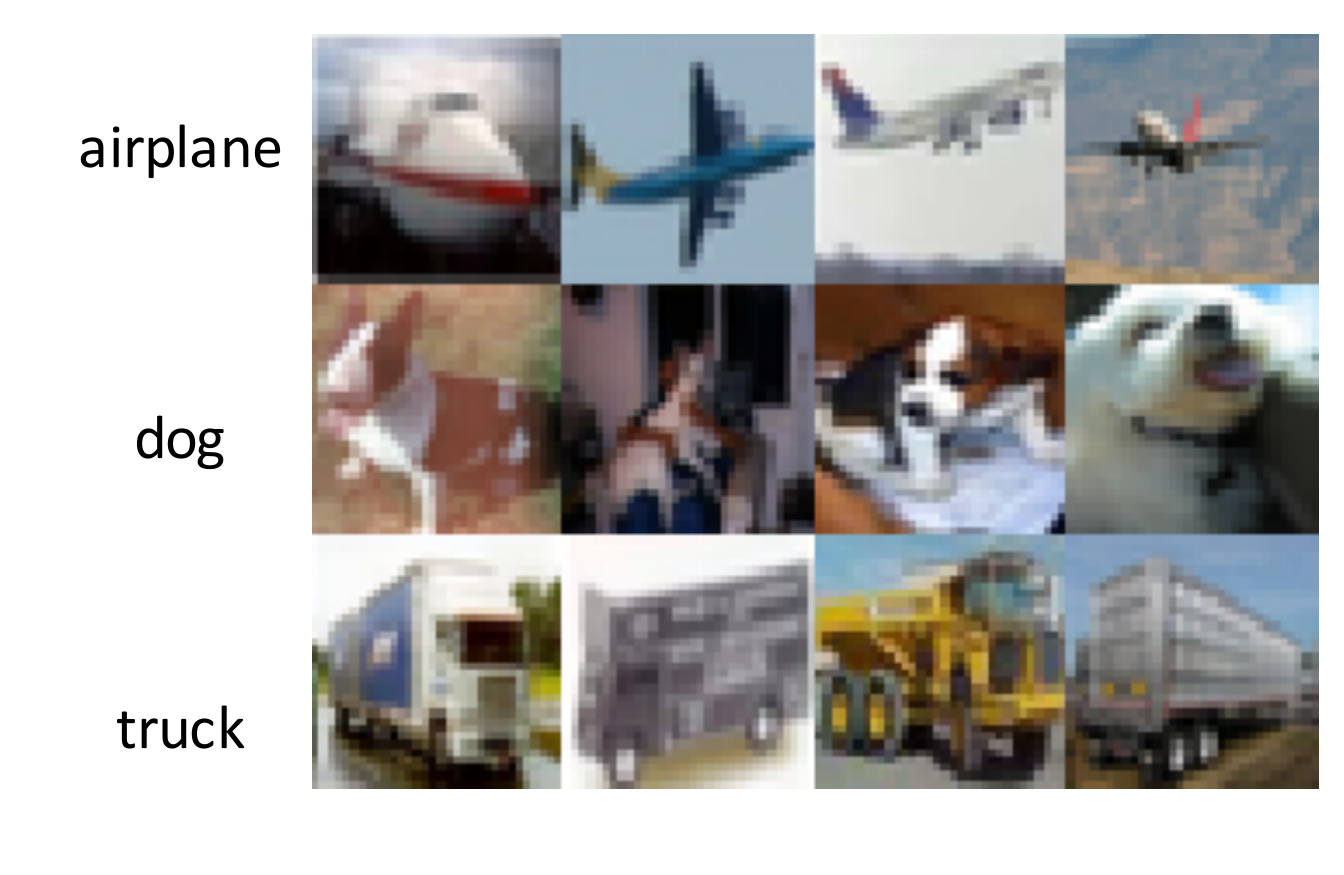}
		\caption{CIFAR-10\label{fig:cifar10_data}}
	\end{subfigure}
	\begin{subfigure}[t]{.24\textwidth}
		\centering
		\includegraphics[width=0.95\textwidth]{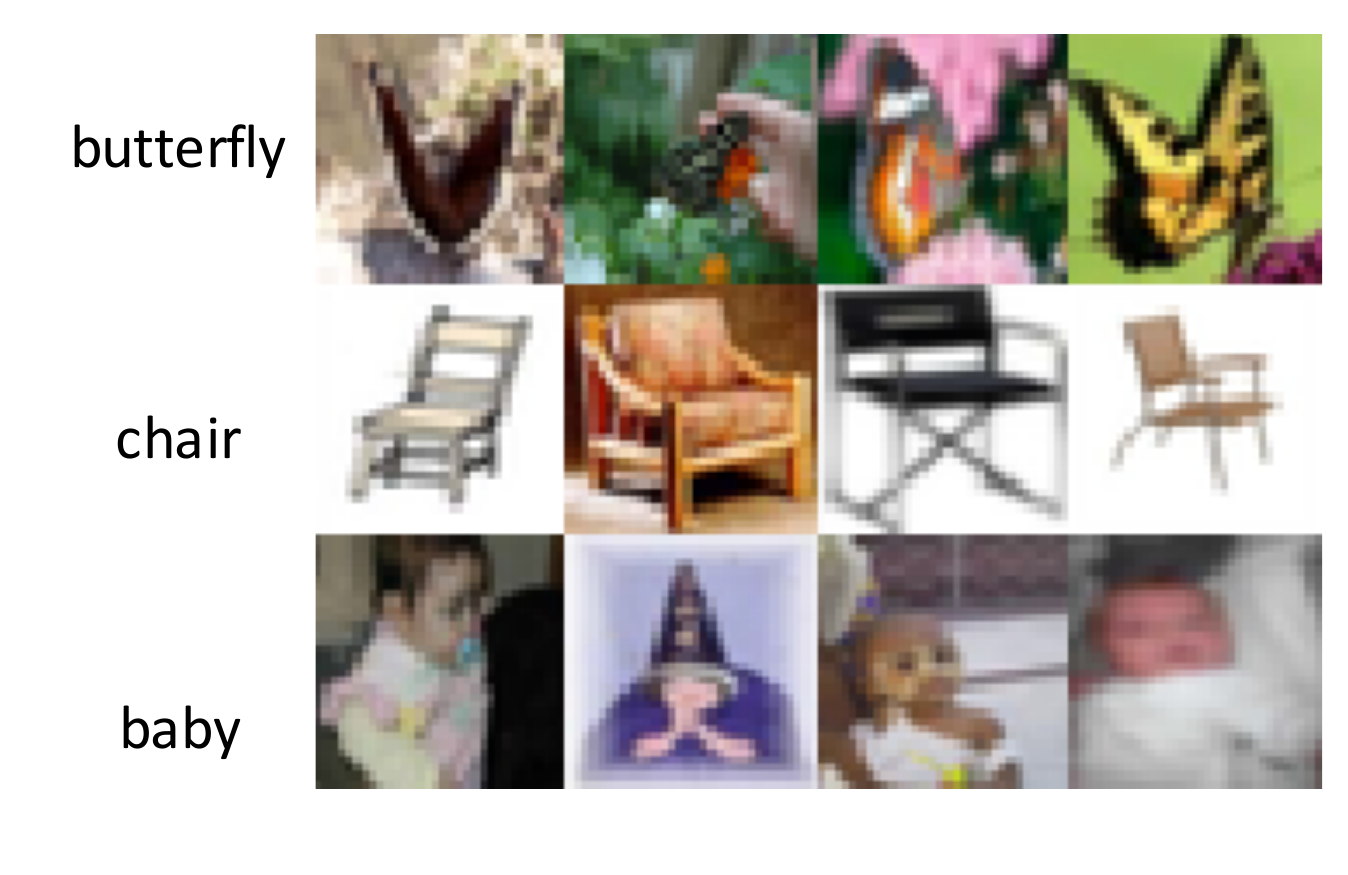}
		\caption{CIFAR-100\label{fig:cifar100_data}}
	\end{subfigure} \\
	\begin{subfigure}[t]{.48\textwidth}
		\centering
		\vspace{+2mm}
		\includegraphics[width=0.92\textwidth]{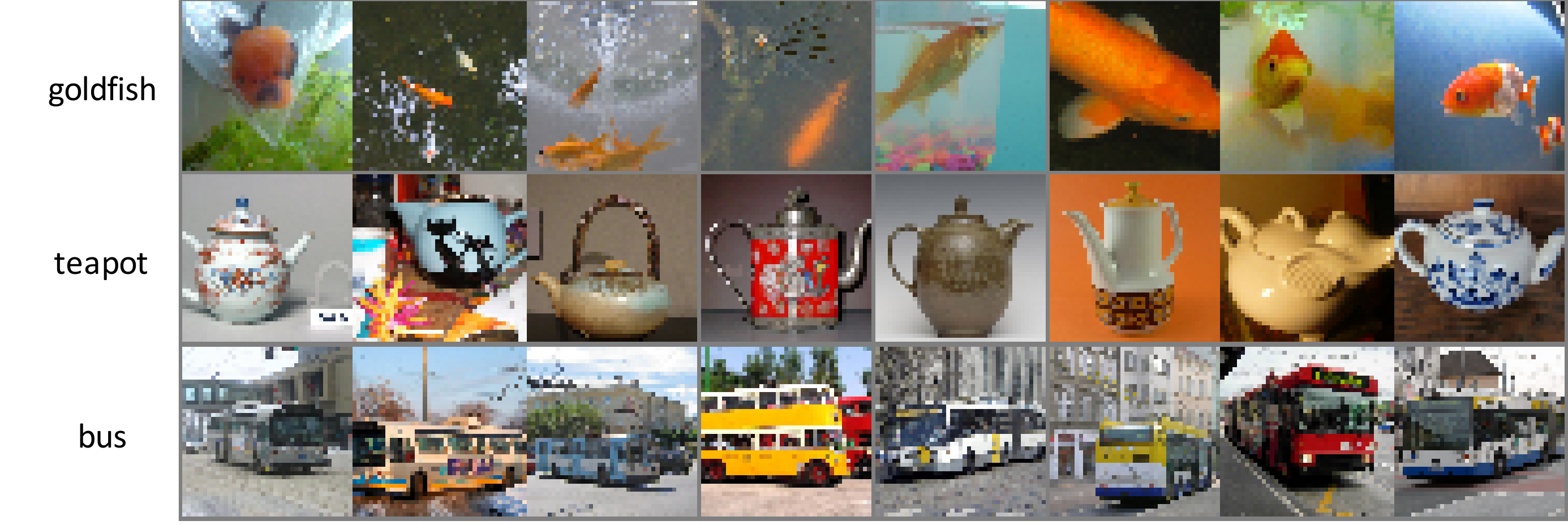}
		\caption{ImageNet\label{fig:imagenet_data}}
	\end{subfigure}
	\caption{Examples from CIFAR-10, CIFAR-100 and ImageNet datasets. Images in each row belong to the same class with label names shown to the left.
	\label{fig:data_gallery}\vspace{-0.3cm}}
\end{figure}

Examples from CIFAR-10, CIFAR-100, and ImageNet are provided in Fig.~\ref{fig:data_gallery}. 

\subsubsection{CIFAR-10.1 and ImageNet-V2}
\begin{figure}[!htbp]
	\centering
	\begin{subfigure}[t]{.48\textwidth}
		\centering
		\includegraphics[width=0.98\textwidth]{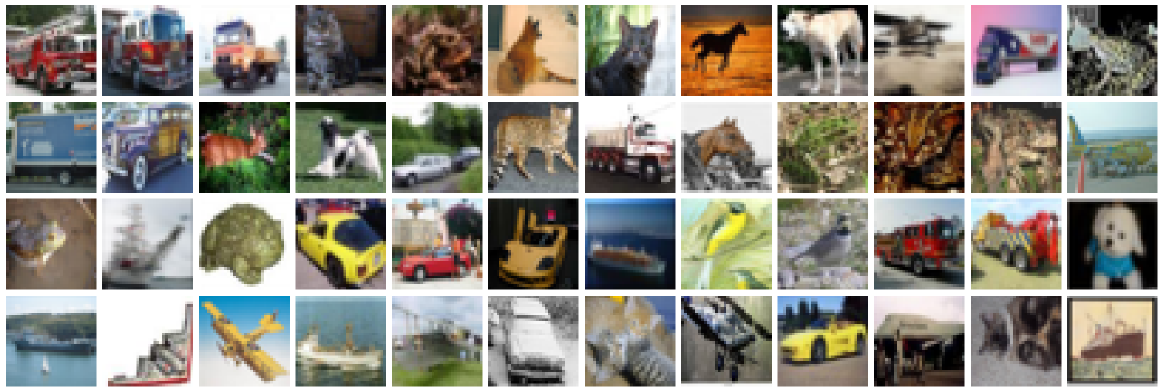}
		\caption{CIFAR-10.1\label{fig:cifar10_1}}
	\end{subfigure}\\ \vspace{3mm}
	\begin{subfigure}[t]{.48\textwidth}
		\centering
		\includegraphics[width=0.98\textwidth]{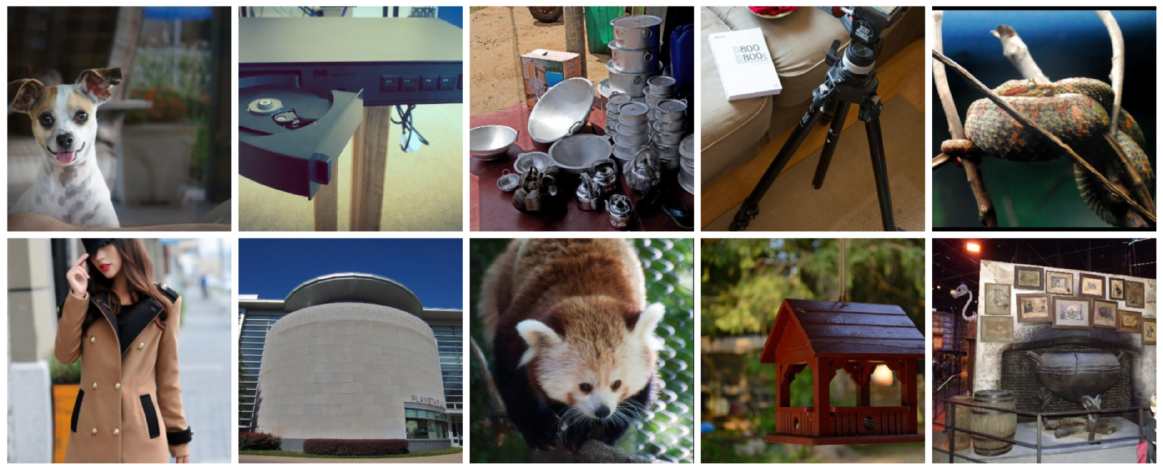}
		\caption{ImageNet-V2\label{fig:imagenetv2}}
	\end{subfigure}
	\caption{Visualization of CIFAR-10.1 (a) and ImageNet-V2 (b). Examples are randomly sampled from the datsets.
	\label{fig:new_test_set}}
\end{figure}

In this work, we use the \emph{MatchedFrequency} version of the ImageNet-V2 dataset. The curation details along with the discussion of the difference among the three versions are available in \cite{recht2019imagenet}. Examples randomly sampled from these two new testing sets are provided in Figs.~\ref{fig:cifar10_1} and \ref{fig:imagenetv2}, repectively. The CIFAR-10.1 is available for download at {\color{red} \url{https://github.com/modestyachts/CIFAR-10.1}}. And the ImageNet-V2 is available at {\color{red} \url{https://github.com/modestyachts/ImageNetV2}}.

\subsubsection{CIFAR-10-C and CIFAR-100-C}
\begin{figure}[!htbp]
	\centering
	\begin{subfigure}[t]{.48\textwidth}
		\centering
		\includegraphics[width=0.98\textwidth]{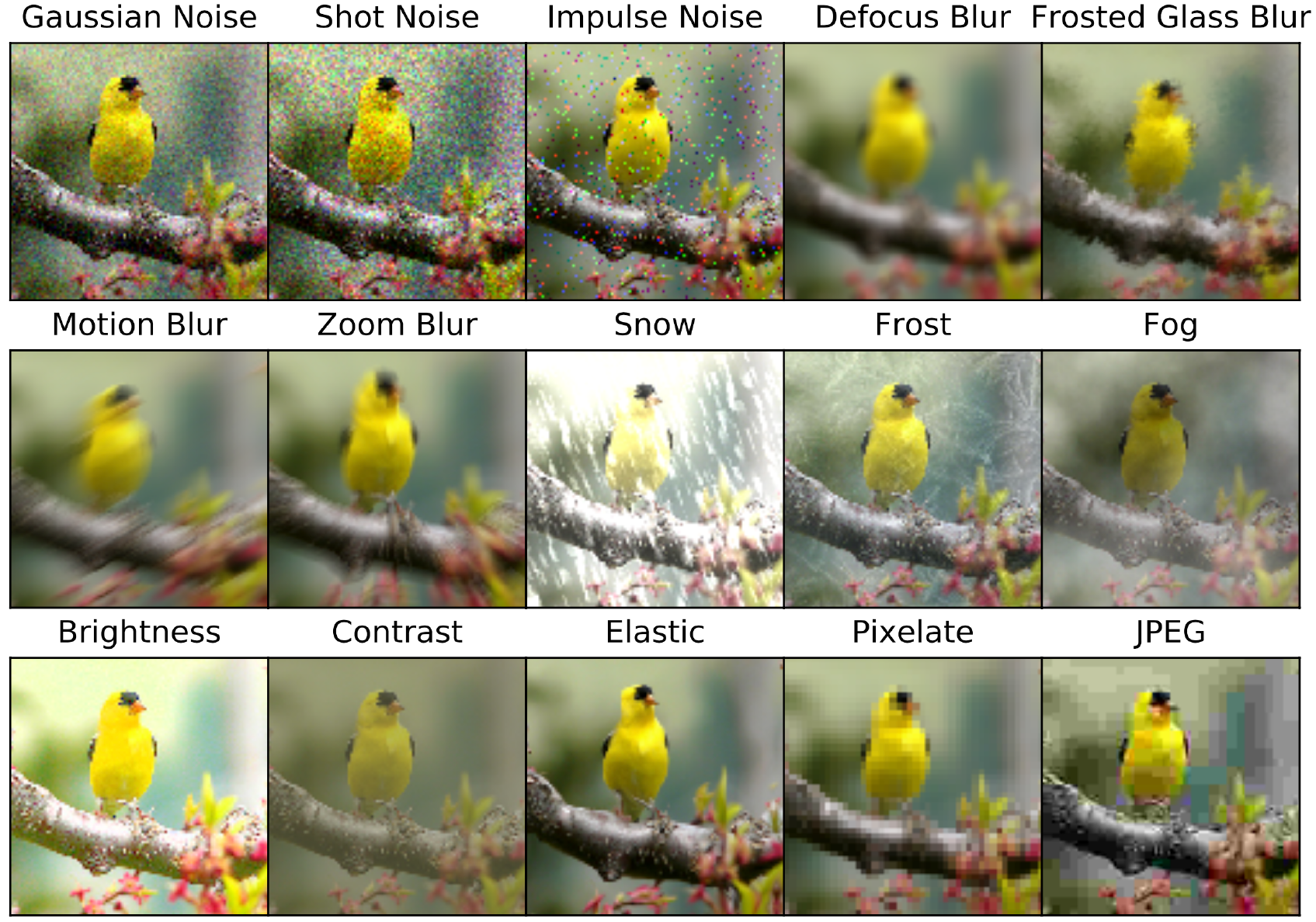}
		\caption{Types of corruptions\label{fig:corruptions}}
	\end{subfigure}\\ \vspace{3mm}
	\begin{subfigure}[t]{.48\textwidth}
		\centering
		\includegraphics[width=0.98\textwidth]{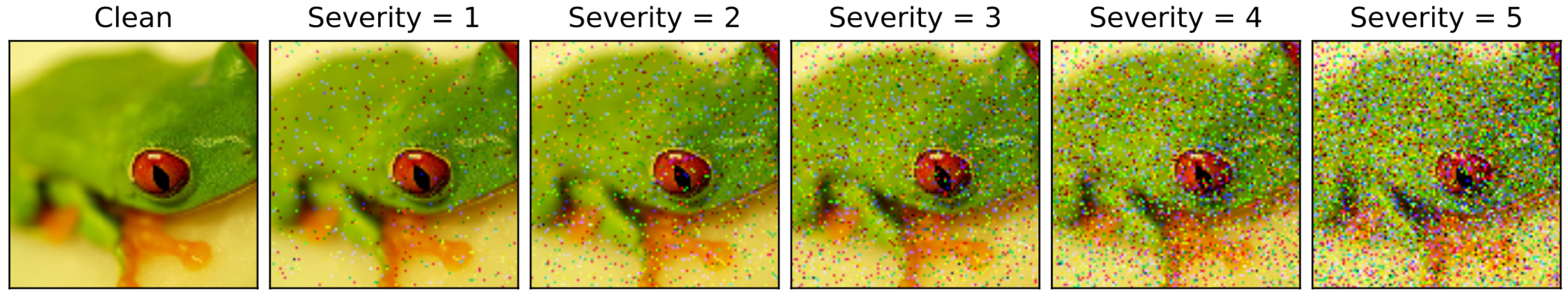}
		\caption{Severity of corruptions\label{fig:severity}}
	\end{subfigure}
	\begin{subfigure}[t]{.48\textwidth}
		\centering
		\includegraphics[width=0.98\textwidth]{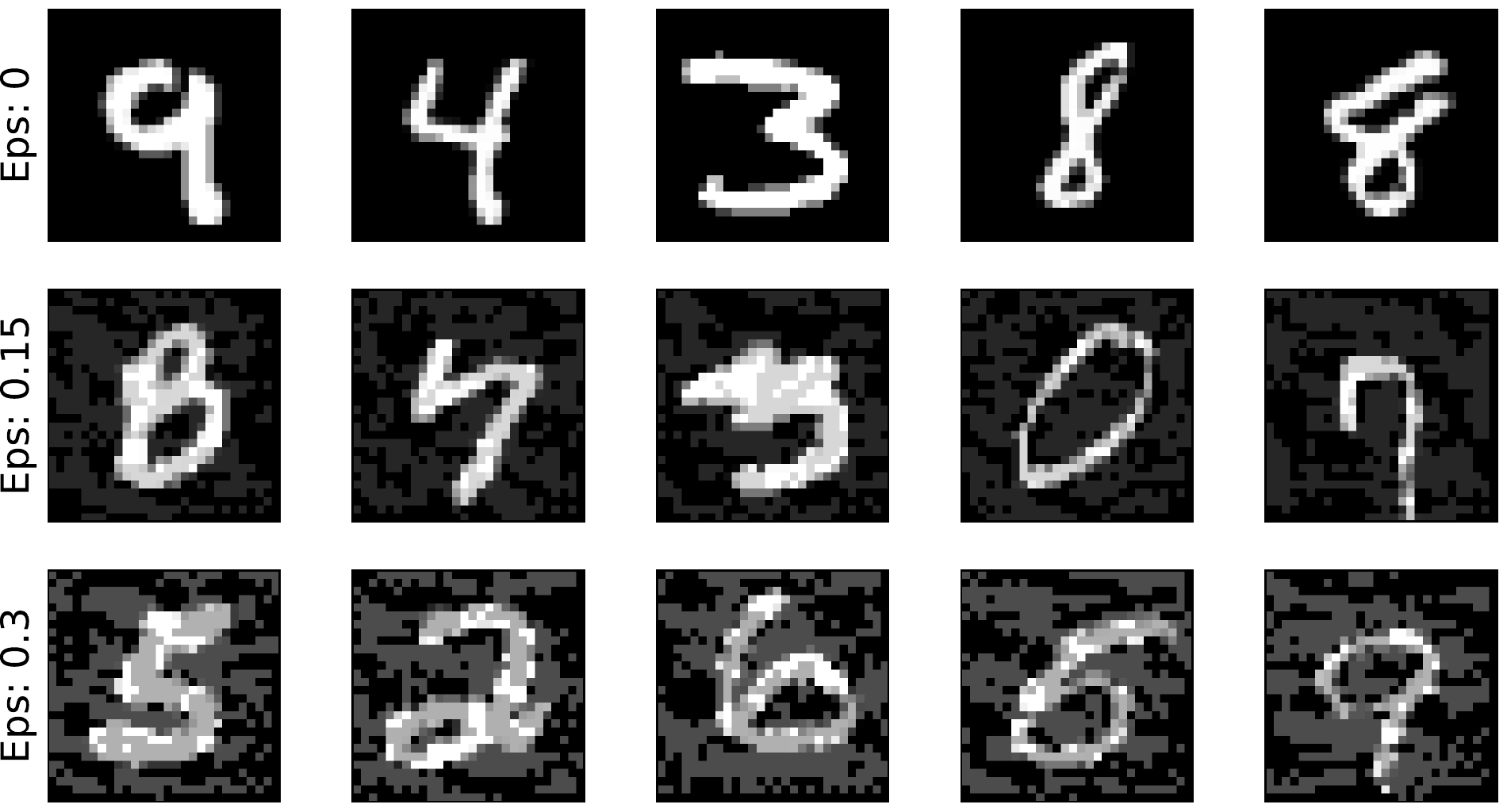}
		\caption{Adversarial examples from FGSM \cite{goodfellow2014explaining} on MNIST.\label{fig:adversarial_mnist}}
	\end{subfigure}
	\caption{Visualization of different types of corruptions and different levels of severity. Examples are from \cite{hendrycks2018benchmarking}. Both CIFAR-10-C and CIFAR-100-C are constructed by applying corruptions to the original testing sets. A demo visualization of adversarial examples from FGSM on MNIST is shown in (c).
	\label{fig:cifar-c}}
\end{figure}

There are in total 19 different commonly observable corruption types considered in both CIFAR-10-C and CIFAR-100-C, including Gaussian noise, shot noise, impulse noise, de-focus blur, frosted glass blur, motion blur, zoom blur, snow, frost, fog, brightness, contrast, elastic, pixelate, jpeg, speckle noise, Gaussian blur, spatter and saturate. Fig.~\ref{fig:corruptions} provides examples for visualization. For every corruption type, there are five different levels of severity, see Fig.~\ref{fig:severity} for visualization. Both datasets are available from the original authors' GitHub page at {\color{red} \url{https://github.com/hendrycks/robustness}}. A demo visualization of adversarial examples created by applying FGSM \cite{goodfellow2014explaining} on MNIST dataset is provided in Fig.~\ref{fig:adversarial_mnist}.

\subsubsection{ChestX-Ray14}
\begin{figure}[!htbp]
	\centering
	\begin{subfigure}[t]{.48\textwidth}
		\centering
		\includegraphics[width=0.98\textwidth]{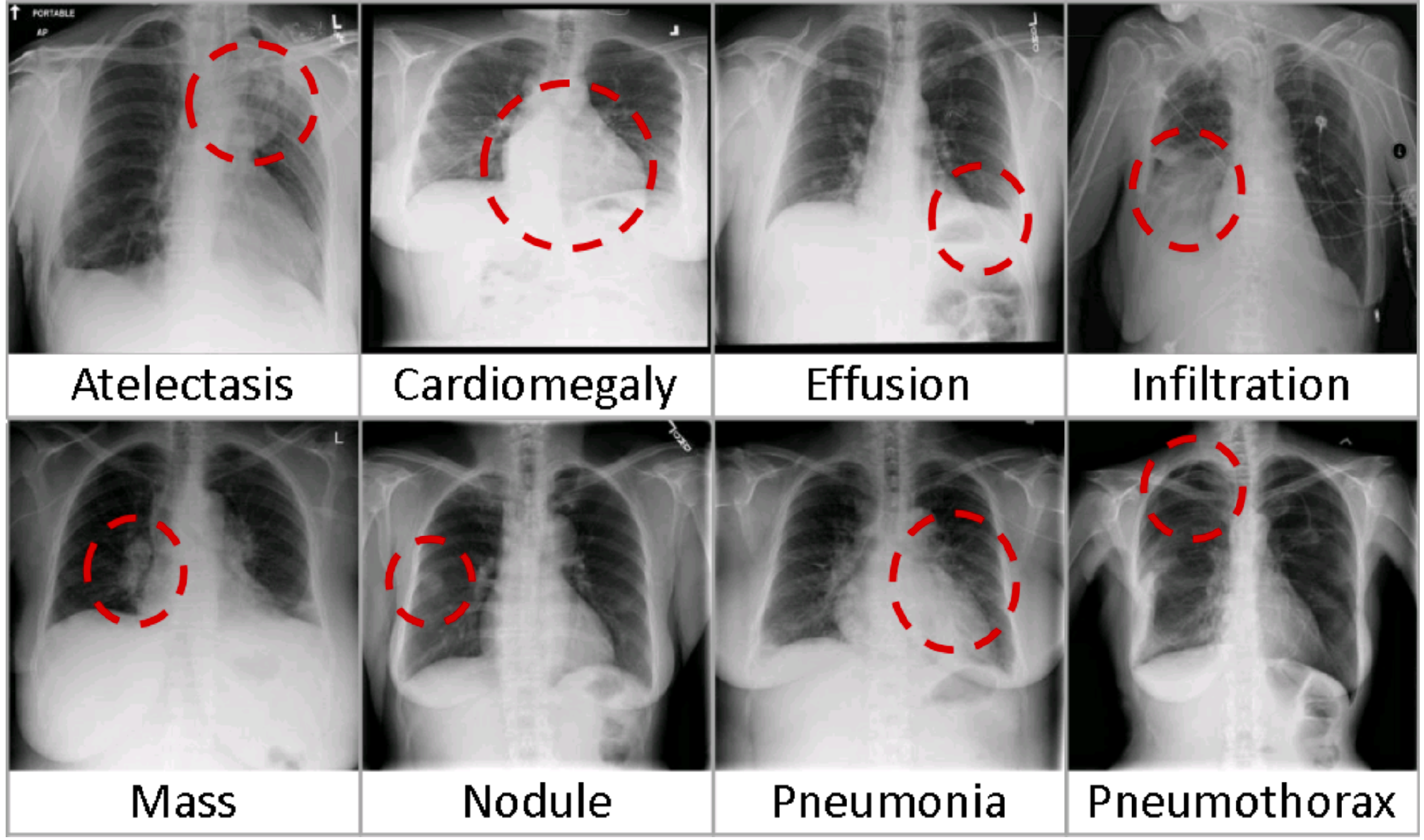}
	\end{subfigure}
	\caption{Visualization of ChestXray14 datasets. Examples showing eight common thoracic diseases are from \cite{wang2017chestx}.
	\label{fig:chexray}}
\end{figure}

ChestX-Ray 14 are hospital-scale Chest X-ray database containing 112,120 frontal-view X-ray images of size 1,024 x 1,024 pixels from 30,805 unique patients. The database is labeled using natural language processing techniques from the associated radiological reports stored in hospitals' Picture Archiving and Communication Systems (PACS). Each image can have one or multiple common thoracic diseases, or "Normal" otherwise. Visualization of example X-ray images from the database is provided in Fig.~\ref{fig:chexray}. The dataset is publicly available from NIH at {\color{red} \url{https://nihcc.app.box.com/v/ChestXray-NIHCC}}. We follow the \emph{train_val_list.txt} and \emph{test_list.txt} provided along with the X-ray images to split the database for training, validation and testing.

{
\subsection{Implementation Details Continued\label{sec:supple_train}}
Evaluating a neural network architecture's performance is computationally expensive---e.g., one evaluation on the CIFAR-10 dataset takes more than 30 minutes. In general, our (GA-related) hyper-parameter choices represent the minimal number of function evaluations required to reproduce the claimed performance. To be more specific, in our proposed algorithm, each architecture is encoded with a 40-position, integer-valued string. Our choice of population size at 40 corresponds to one individual per variable dimension, which follows one of the common suggestions in the GA literature on minimal required population size. Empirically, we observed that the hypervolume stabilized by generation 30 (see Fig.9a in the revised main paper), hence, we chose to terminate the proposed algorithm at generation 30. Other hyper-parameter choices are discussed in Section V.C of the revised main paper.}

\subsection{Follow-up Studies\label{sec:utilities}}
\subsubsection{Single-Objective \ourmethod{}}
Despite the superior effectiveness and efficiency of the proposed algorithm, the computation overheads of 27 GPU-days of \ourmethod{} can be infeasible for users with few GPU cards. Towards understanding of the overall search wall time limit of \ourmethod{}, as well as comparison to the peer methods that use less GPU-days to execute the search, the following experiment has been performed. We minimized the search setup differences by dropping the second objective of minimizing FLOPs and changing the search dataset to CIFAR-10. We also reduce the population size by half and perform early-termination at one and four GPU-days. The obtained architectures are named as \ourmethod{}-B0 and \ourmethod{}-B1, respectively.

\begin{table}[!htbp]
    \caption{\ourmethod{} with single objective of maximizing classification accuracy on CIFAR-10 and early terminations.}
    \label{tab:nsganet-early}
    \centering
    \resizebox{0.45\textwidth}{!}{%
    \begin{tabular}{@{ }l|c|ccc@{ }}
    \toprule
    Method & Type & \#Params & Top-1 Acc. & GPU-Days \\ \midrule
    Genetic CNN \cite{genetic-cnn} & EA & - & 92.90\% & 17 \\
    AE-CNN+E2EPP \cite{ae-cnn-e2epp} & EA & 4.3M & 94.70\% & 7 \\
    ENAS \cite{pmlr-v80-pham18a} & RL & 4.6M & 97.11\% & 0.5 \\
    DARTS \cite{liu2018darts} & differential & 3.3M & 97.24\% & 1 \\\midrule
    \textbf{\ourmethod{}-B0} & \textbf{EA} & \textbf{3.3M} & \textbf{96.15\%} & \textbf{1} \\
    \textbf{\ourmethod{}-B1} & \textbf{EA} & \textbf{3.3M} & \textbf{97.25\%} & \textbf{4} \\ \bottomrule
    \end{tabular}%
    }
\end{table}

Results in Table~\ref{tab:nsganet-early} confirm that our proposed algorithm can be more efficient in GPU-days than the other two EA-based peer methods, Genetic CNN \cite{genetic-cnn} and AE-CNN-E2EPP \cite{ae-cnn-e2epp}. Specifically, \ourmethod{} obtains the architecture B1 in 3 less GPU-days than AE-CNN-E2EPP, in addition to the B1 architecture being more accurate in CIFAR-10 classification and less complex in number of parameters. Due to the use of weight sharing that partially eliminates the back-propagation weight learning process, ENAS \cite{pmlr-v80-pham18a} and DARTS \cite{liu2018darts} are still more efficient in GPU-days than our proposed method. The weight sharing method could in principle be applied to \ourmethod{} as well, however this is beyond the scope of this paper.

\subsubsection{Effectiveness of Non-learnable Operations}
\begin{figure}[!htbp]
	\centering
	\begin{subfigure}[t]{.45\textwidth}
		\centering
        \includegraphics[width=0.98\textwidth{}]{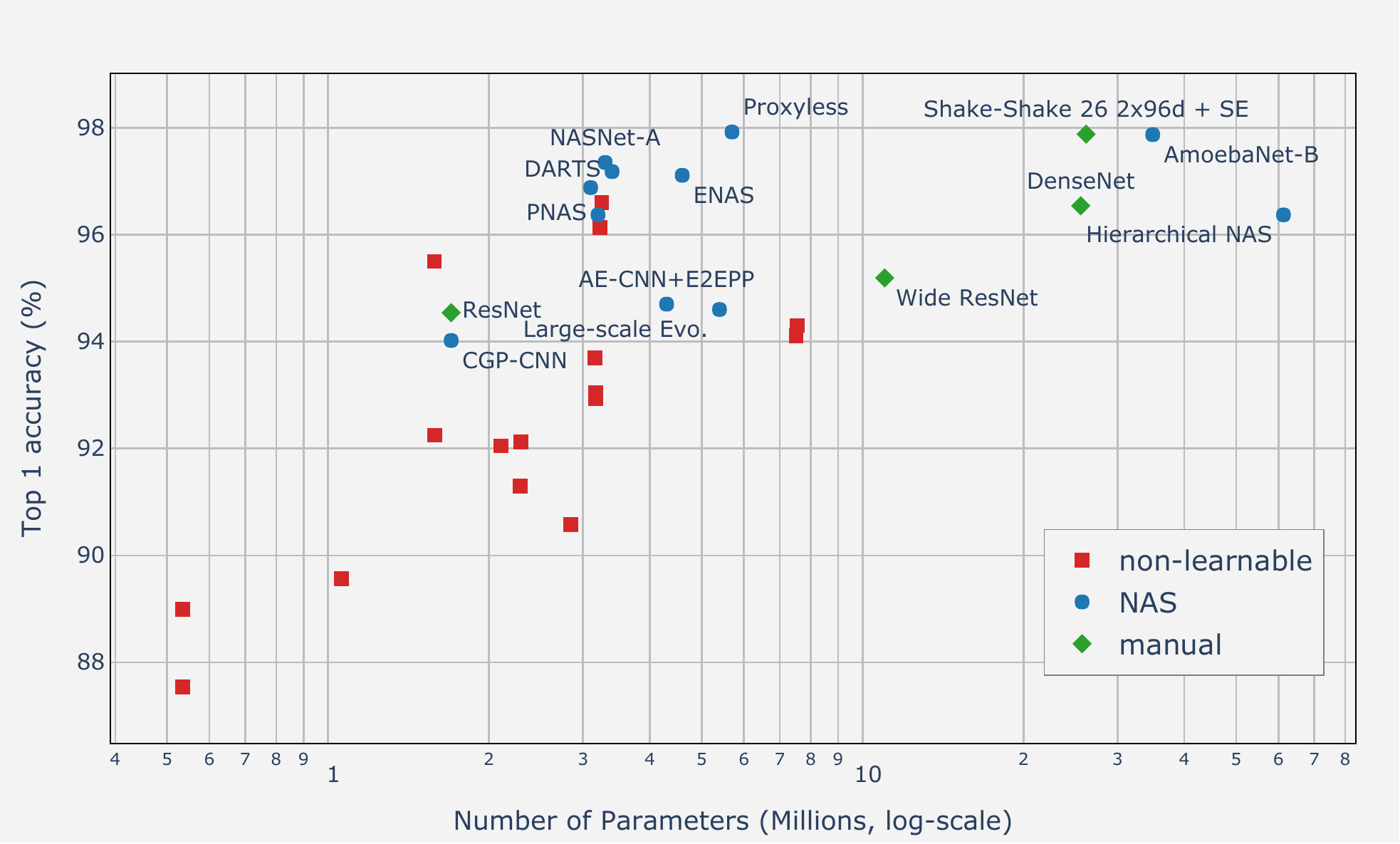}
        \caption{}
        \label{fig:manual_cifar10}
	\end{subfigure}\\
	\begin{subfigure}[t]{.40\textwidth}
		\centering
        \includegraphics[width=0.98\textwidth{}]{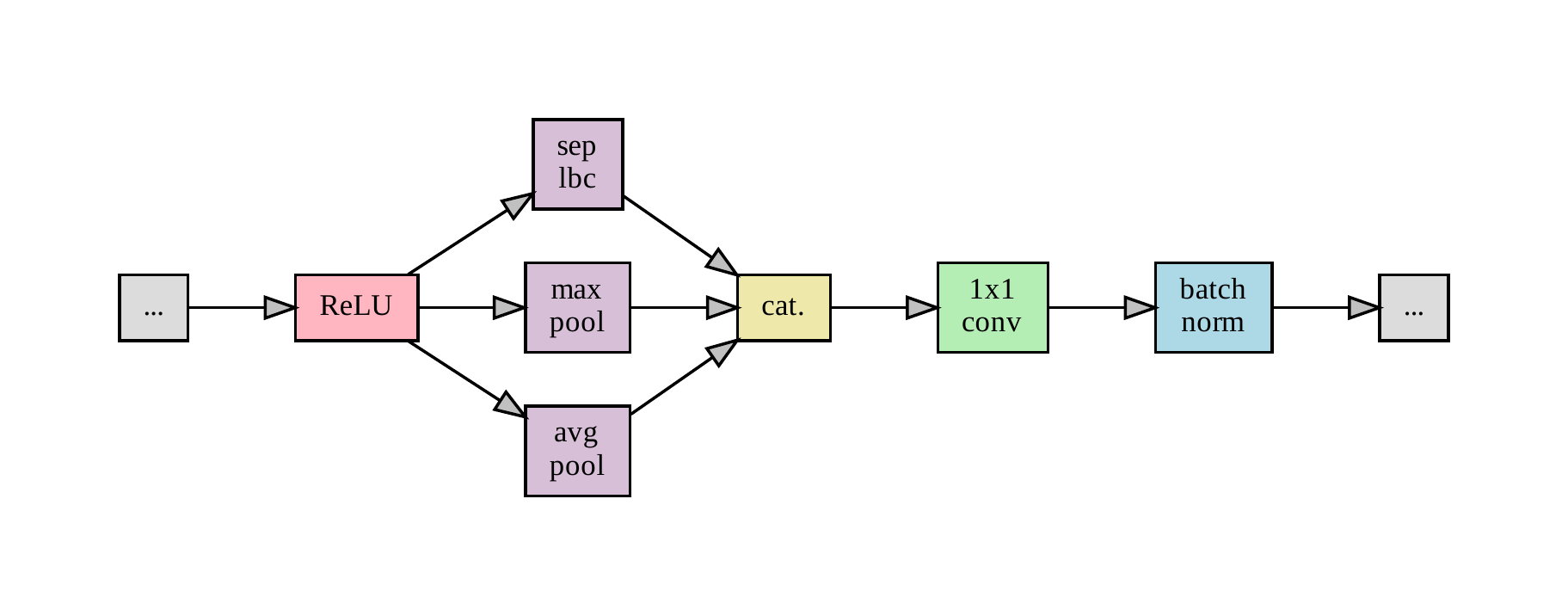}
        \caption{}
        \label{fig:auxiliary_utd}
	\end{subfigure}\\
	\begin{subfigure}[t]{.48\textwidth}
		\centering
        \includegraphics[width=0.98\textwidth{}]{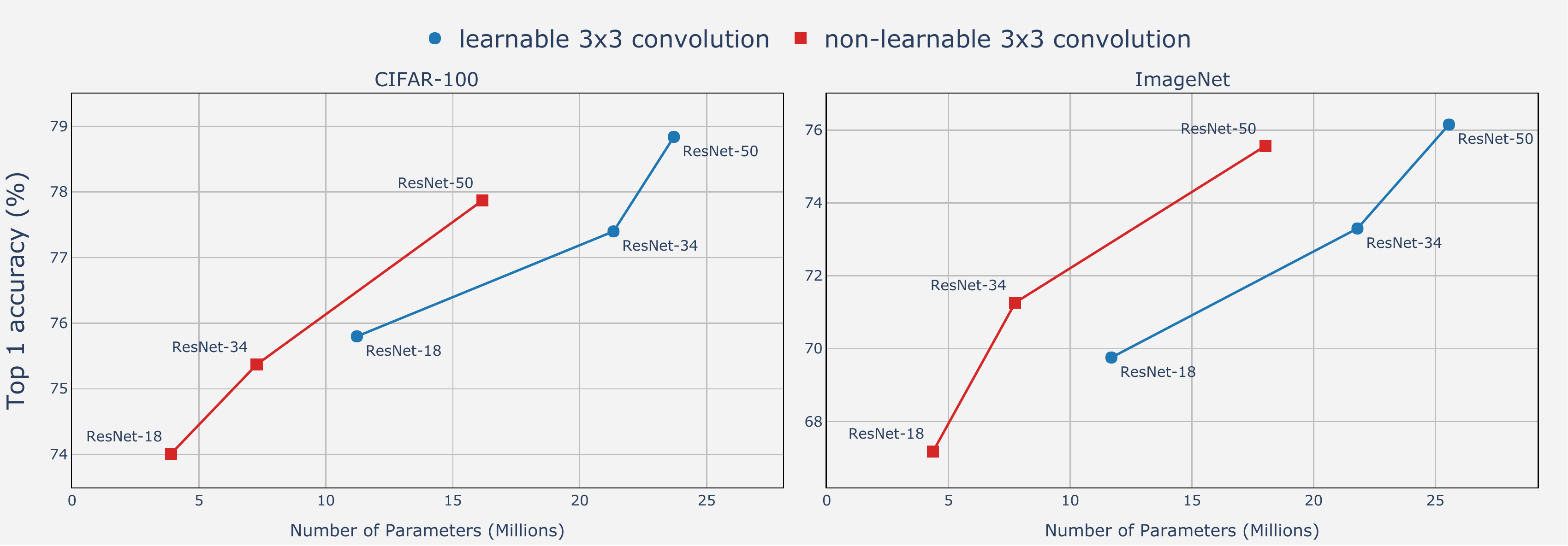}
        \caption{}
        \label{fig:manual_cifar100_imagenet}
	\end{subfigure}
	\caption{Preliminary experiment on constructing DNN architectures using layers with non-learnable weights. Each layer is composed of several non-learnable operations in parallel. We manually constructed a handful of such layers and evaluate them on CIFAR-10 (a). An example configuration, based on the trade-off between accuracy and the number of the parameters, is shown in (b). We validate the effectiveness of non-learnable layers by replacing the original 3$\times$3 convolution in different ResNet models with the chosen configuration on both CIFAR-100 and ImageNet (c). Evidently, layer with non-learnable weights is capable of yielding competitive classification performance while being computational efficient as opposed to conventional learnable convolutions.
	\label{fig:preliminary_non_learnable}\vspace{-0.3cm}}
\end{figure}

Our post-optimization analysis on the evolved architectures, shown in Section IV-E, has revealed some interesting findings, one of which being the effectiveness of non-parametric operations, e.g. identity mapping, average/max pooling, etc., in trading off classification performance for architectural complexity. To further validate this observation, we consider a expanded range of operations including both non-parametric and weight-fixed operations, which we name as \emph{non-learnable operations} in this paper. We manually construct such layers by concatenate multiple non-learnable operations in parallel. The obtained results are shown in Figs.~\ref{fig:manual_cifar10} - \ref{fig:manual_cifar100_imagenet}.

Our preliminary results on manual construction of non-learnable layers are very encouraging. In additional to the comparative performance to regular fully learned layers, non-learnable layers offer unique advantages in terms of re-usable weights for multi-tasking network architectures, as the weights are agnostic (not specifically learned on a particular task). We believe designing dedicated search algorithm to shape the construction of these non-learnable layers is a promising direction for NAS towards automatic design for multi-tasking architectures.

\subsubsection{Robustness Against Adversarial Attacks}
Based on our analysis in Section V-B, years of architectural advancements have translated to minuscule improvements in robustness against adversarial examples. Simple one-iteration attack strategy like FGSM \cite{goodfellow2014explaining} is enough to constructing examples that turn many modern DNN classifiers to random-guess (see Fig. 12 for examples). In this section, we make an effort to improve adversarial robustness from the architectural perspective. The search space used in the main paper searches over both layer operations and layer connections (see Section III-A). To isolate the effect of these two aspects to the adversarial robustness, we fix the layer operation to basic residual module \cite{resnet} and search over the connections among these modules to improve both classification accuracy on clean images and robustness against adversarial examples.

\begin{figure}[!htbp]
    \centering
    \includegraphics[width=0.3\textwidth]{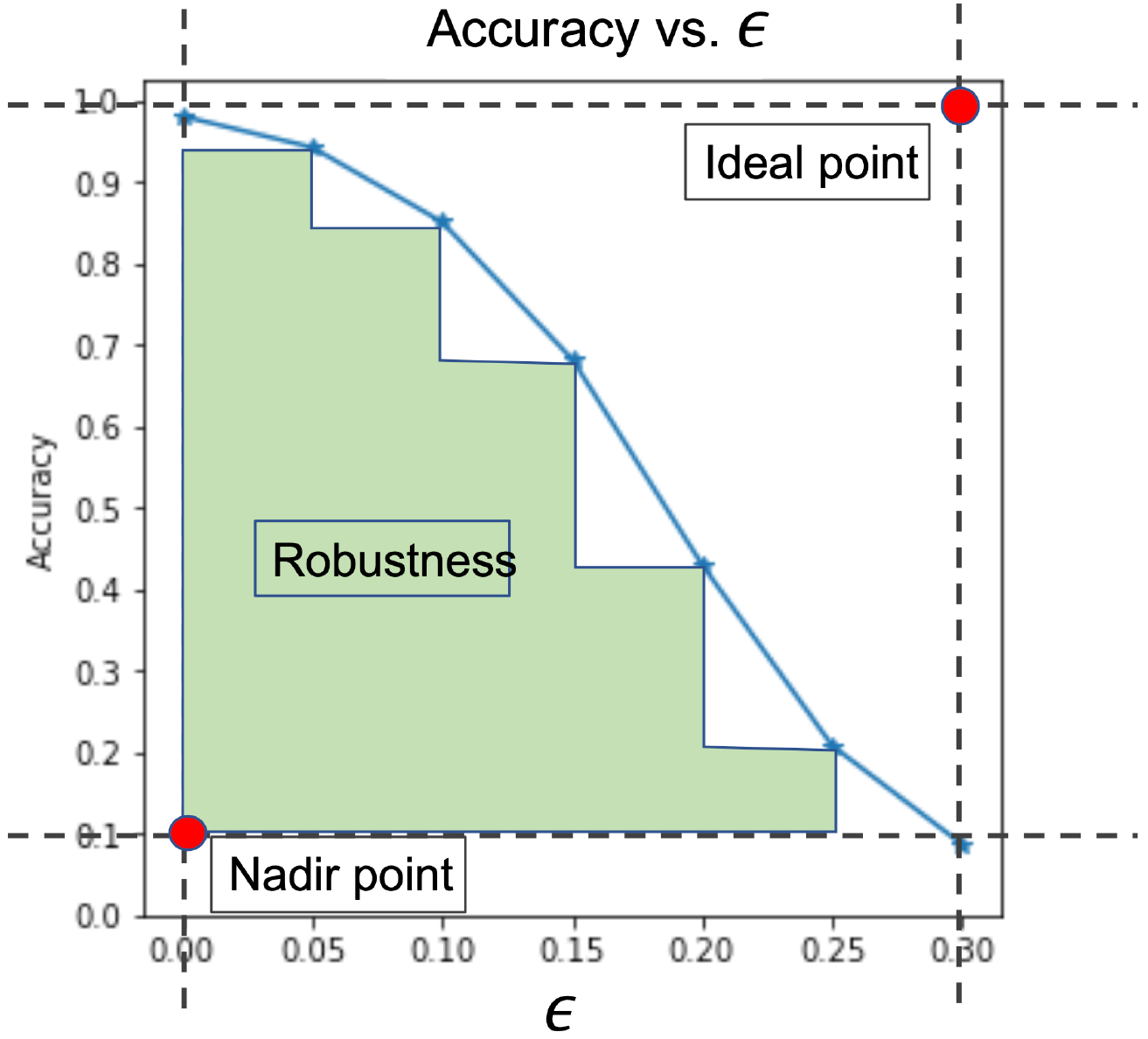}
    \caption{\textbf{Robustness Objective:} We define a robustness objective under the FGSM \cite{goodfellow2014explaining} attack as follows: 1) obtain classification accuracy on adversarial images generated by FGSM as we vary $\epsilon$, 2) compute the area under the curve (blue line), approximated by the area of the green region; 2) normalize the robustness value to the rectangular area formed by the \textit{Ideal point} and \textit{Nadir point}; 3) Ideal point is defined at 100\% accuracy at pre-defined maximum $\epsilon$ value, and the nadir point is defined as the accuracy of random guessing at $\epsilon=0$ (clean images).}
    \label{fig:robustness_calculation}
\end{figure}

Designing a measure/objective for robustness against adversarial robustness is an area of active research (e.g., \cite{carlini2017towards}). For our purposes, we present a possible measure here, illustrated in Fig.~\ref{fig:robustness_calculation}. Using the FSGM presented by \cite{goodfellow2014explaining}, this robustness objective  progressively increases noise produced by FSGM. The $\epsilon$ axis in Fig.~\ref{fig:robustness_calculation} refers to the hyper-parameter in the FSGM equation,
\begin{align*}
    \bm{x}' = \bm{x} + \epsilon ~ \text{sign}( \nabla_{\bm{x}}\mathcal{L}(\bm{x}, y_{true})),
\end{align*}
\noindent where $\bm{x}$ is the original image, $\bm{x}'$ is adversarial image, $y_{true}$ is true class label, and $\mathcal{L}$ cross-entropy loss. Therefore, for this experiment, we seek to maximize two objectives, namely, classification accuracy and the robustness objective defined above.

The setup for the robustness experiment is as follows. For training we use 40,000 CIFAR-10 images from the official CIFAR-10 training data, 10,000 of which are reserved for validation. Each network is encoded with three blocks using the macro space encoding from our previous work \cite{lu2019nsga}. In each phase a maximum of size nodes may be active---where the computation at each node is 3x3 convolution followed by ReLU and batch normalization. Each network is trained for 20 epochs with SGD on a cosine annealed learning rate schedule. The epsilon values used in the FSGM robustness calculation are [0.0, 0.01, 0.03, 0.05, 0.07, 0.1, 0.15]. As before, \ourmethod{} initiates the search with 40 randomly created network architecture, and 40 new network architectures are created at each generation (iteration) via genetic operations (see main paper for details). The search is terminated at 30 generations.

\begin{figure}[hbt]
    \centering
    \includegraphics[width=0.45\textwidth]{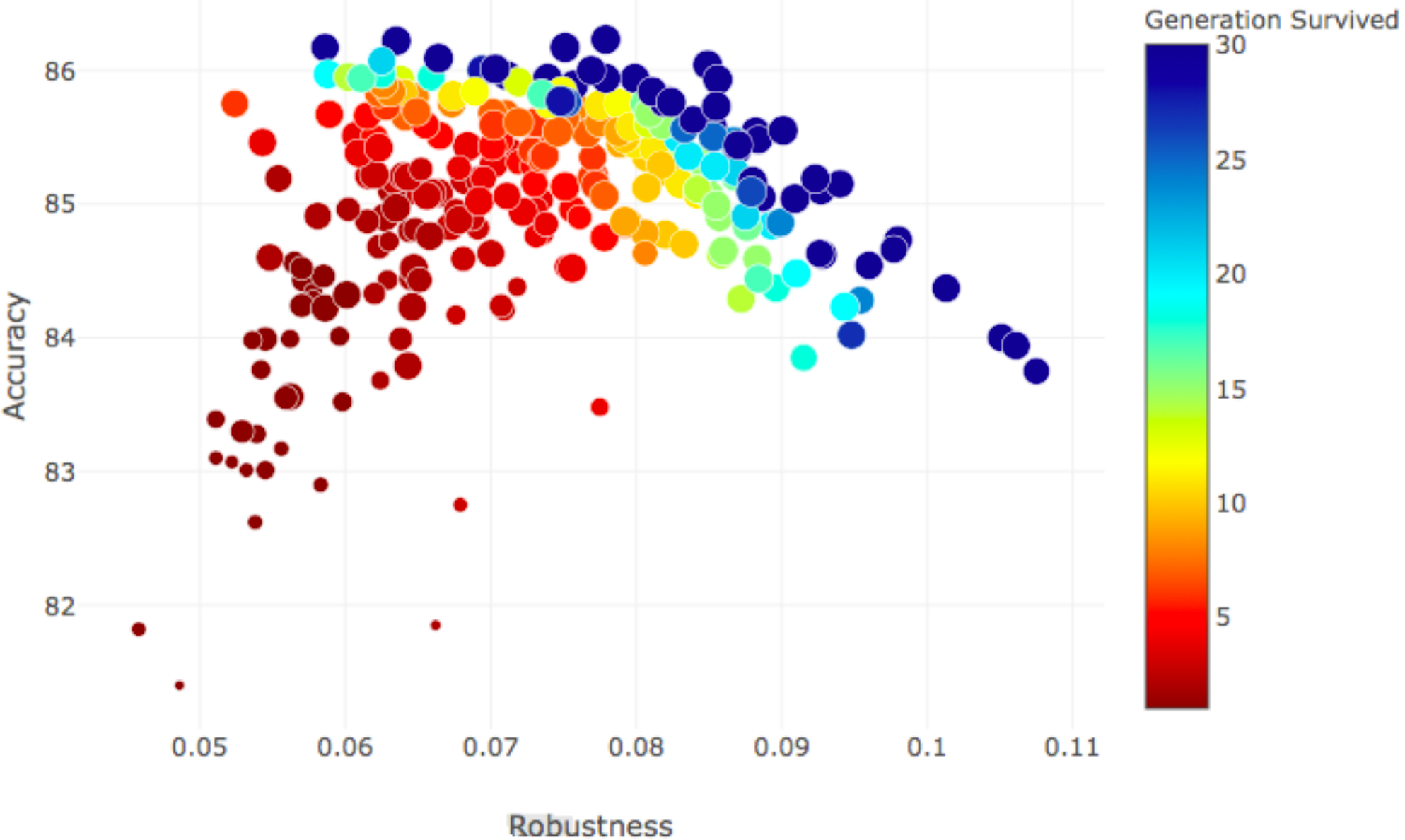}
    \caption{Trade-off frontier of the robustness experiments. Color indicates the generation (iteration) at which a network architecture is eliminated from the surviving parent population. The size of each point is proportional to the network architecture's number of trainable parameters. We note that networks for latter generations form the Pareto front (dark blue points).}
    \label{fig:robustness_pareto_progress}
\end{figure}

\begin{figure}
    \centering
    \begin{subfigure}{0.3\textwidth}
        \centering
        \includegraphics[height=0.1\textheight]{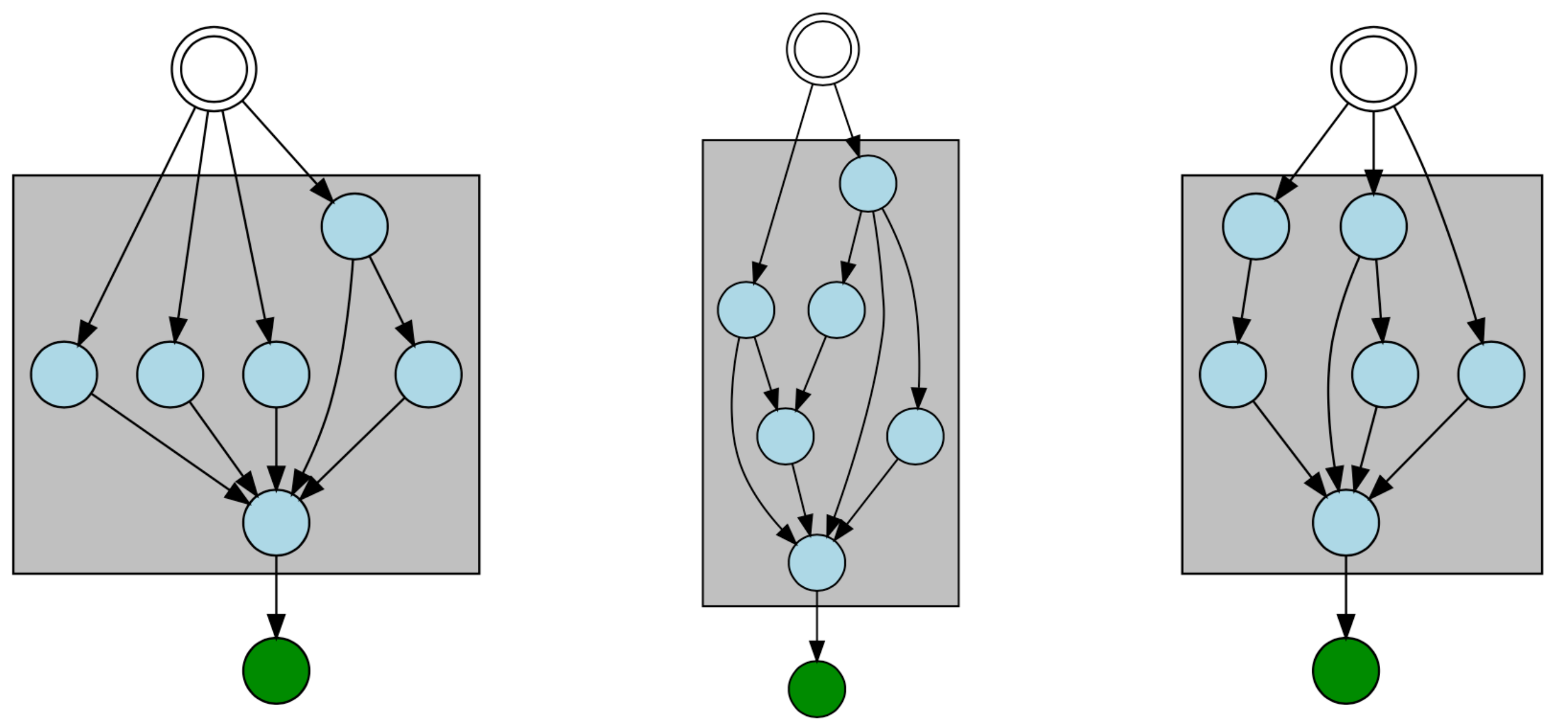}
        \caption{\label{fig:accurate_phases}}
    \end{subfigure}%
    \begin{subfigure}{0.175\textwidth}
        \centering
        \includegraphics[height=0.1\textheight]{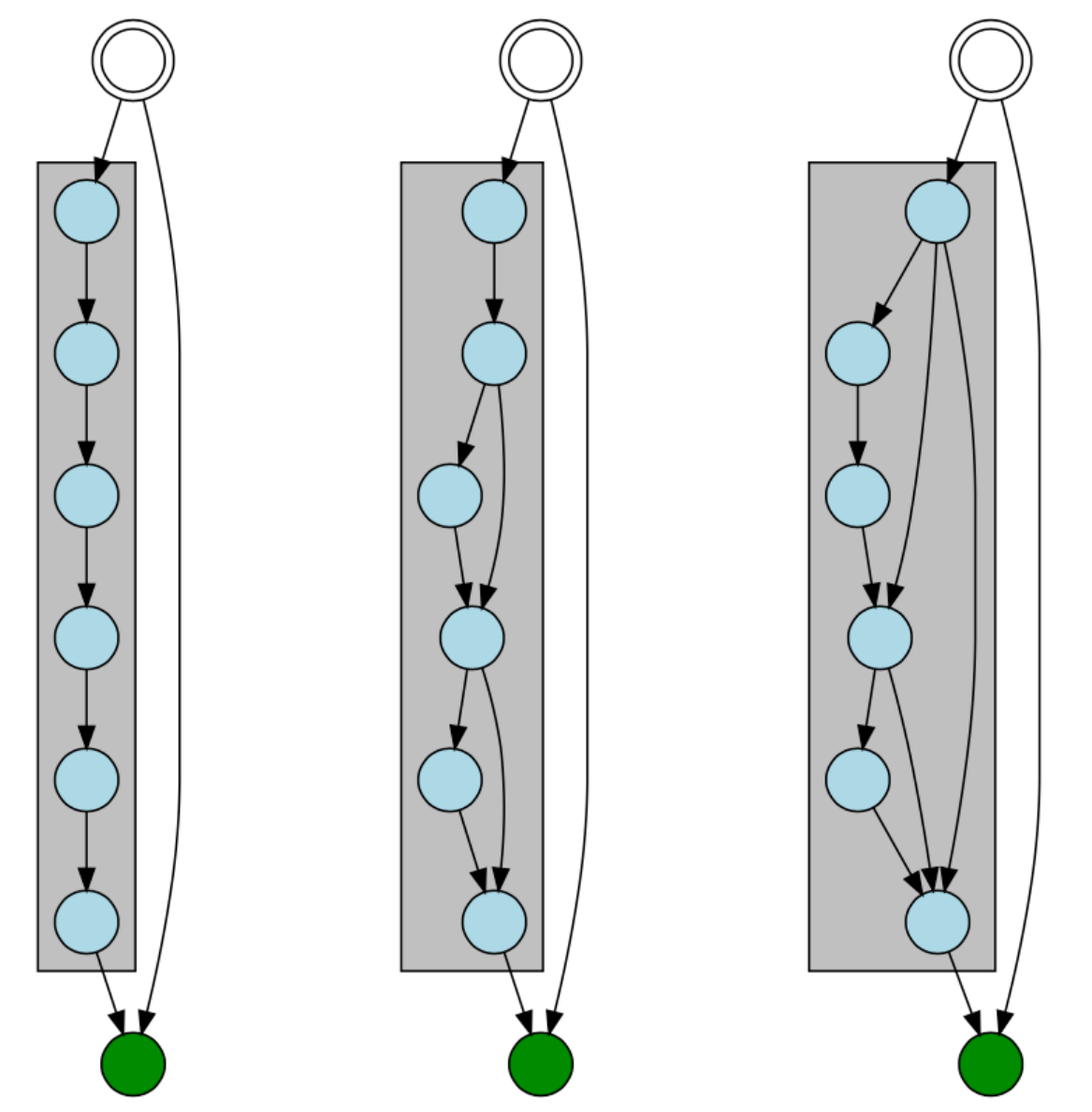}
        \caption{\label{fig:robust_phases}}
    \end{subfigure}
    \begin{subfigure}{0.48\textwidth}
        \centering
        \includegraphics[height=0.12\textheight]{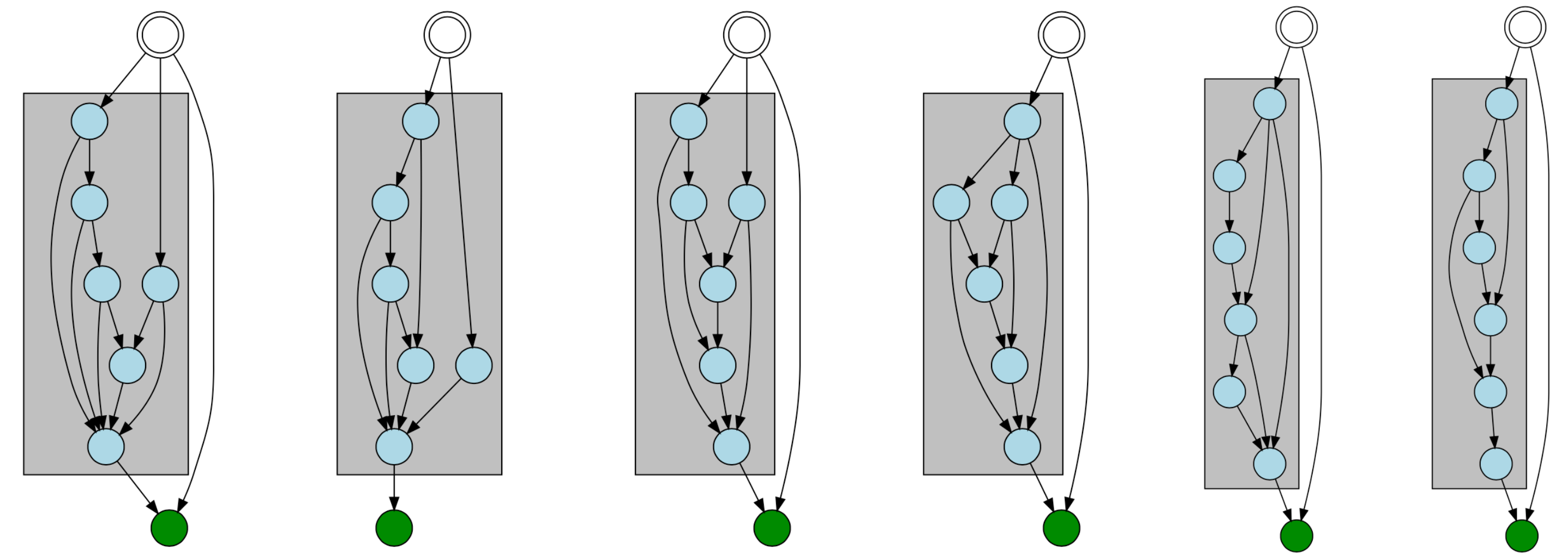}
        \caption{\label{fig:pareto_phases}}
    \end{subfigure}
    \caption{(a) Examples of the computational blocks discovered with high classification accuracy. For these networks, the mean accuracy and robustness objectives are 0.8543 and 0.0535, respectively; (b) Examples of the computational blocks discovered with high robustness against FGSM attack, the mean accuracy and robustness objectives are 0.8415 and 0.1036, respectively; (c) Examples of the computational blocks discovered along the pareto-front that provides an efficient trade-off between classification accuracy and adversarial robustness. They are arranged in the order of descending accuracy and ascending robustness.}
\end{figure}

\begin{figure}
    \centering
    \begin{subfigure}{0.48\textwidth}
        \includegraphics[width=0.98\textwidth]{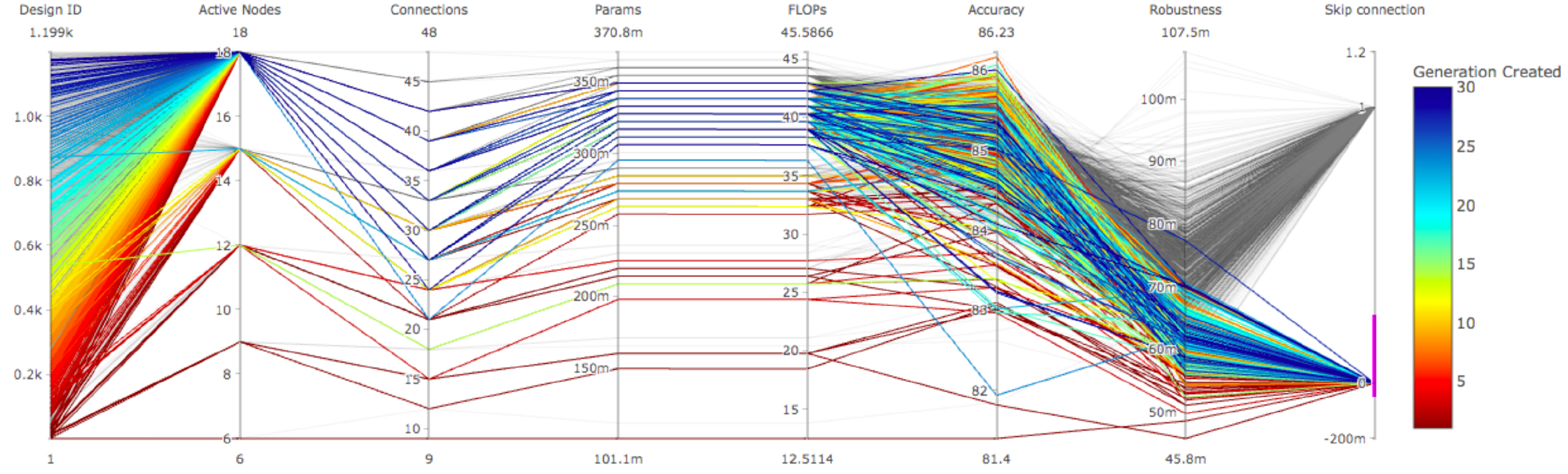}
        \caption{\label{fig:robustness_pcp}}
    \end{subfigure}%
    \\
    \begin{subfigure}{0.48\textwidth}
        \includegraphics[width=0.98\textwidth]{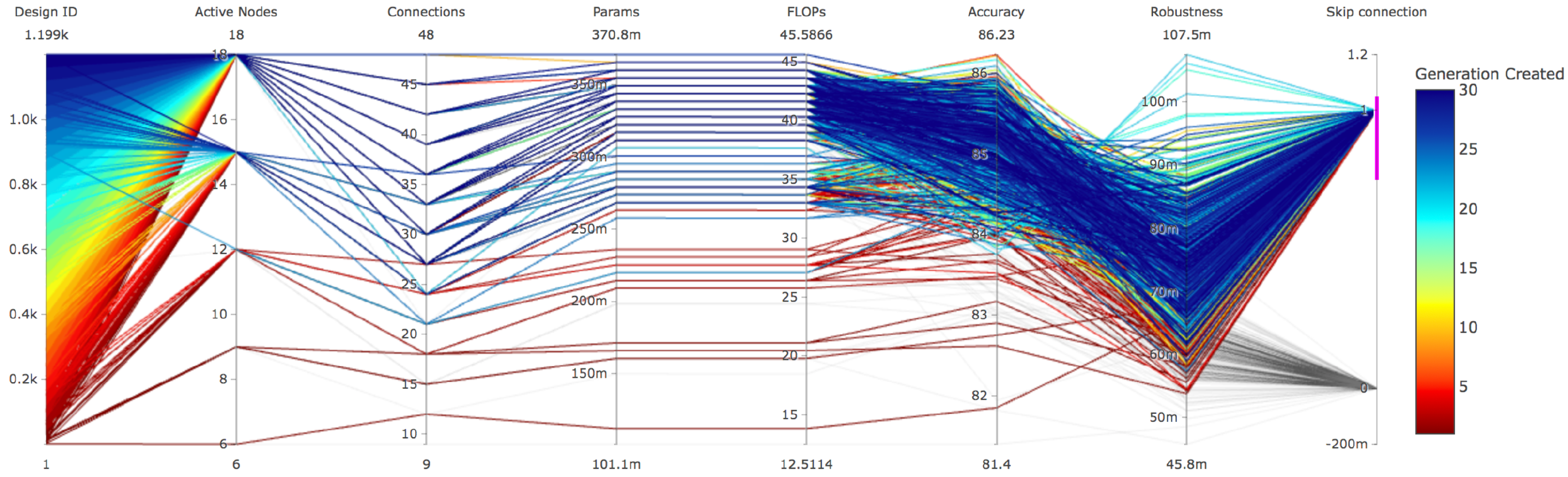}
        \caption{\label{fig:robustness_pcp2}}
    \end{subfigure}%
    \caption{Parallel coordinate plots of the 1,200 network architectures sampled by \ourmethod{}. Each line represents a network architecture, each vertical line is an attribute associated with the network. (a) Networks that have the skip connection bit inactive, we can see that none of them have good measurement on robustness against adversarial attacks. (b) Networks that have the skip connection bit active. This skip connection bit refers to the connection that goes past all computation within a phase, as a normal residual connection would. When the skip connection is active, the networks cover the full range of adversarial robustness.}
\end{figure}

Empirically, we observe a clear trade-off between accuracy and robustness, as shown in Fig.~\ref{fig:robustness_pareto_progress}. Visualization of the non-dominated architectures are provided in Fig.~\ref{fig:pareto_phases}. In our opinion, \ourmethod{} is useful in capturing patterns that differentiate architectures that are good for competing objectives. We find that the ``wide'' networks (like ResNeXt \cite{xie2017aggregated} or Inception blocks \cite{googlenet}) appear to provide good accuracy on standard benchmark images, but are fragile to the FSGM attack. On the other hand, ``deep'' networks (akin to ResNet \cite{resnet} or VGG \cite{vgg}) are more robust to FSGM attack, while having less accuracy. This phenomenon is illustrated with examples in Figs.~\ref{fig:accurate_phases} and \ref{fig:robust_phases}, respectively. Furthermore, the skip connection of skipping the entire block's computation appears to be critical in obtaining a network that is robust to adversarial attacks; see Fig.~\ref{fig:robustness_pcp} and \ref{fig:robustness_pcp2}.

\subsubsection{An Application to Multi-view Car Alignment}
In addition to object classification, dense image prediction (e.g. object alignment, human body pose estimation and semantic segmentation, etc.) is another class of problems that is of great importance to computer vision. Dense image prediction assigns a class label to each pixel in the query images, as opposed to one label to the entire image in case of classification. In this section, we apply \ourmethod{} to the problem of multi-view car key-points alignment.

\begin{figure}[!htbp]
	\centering
	\begin{subfigure}[t]{.48\textwidth}
		\centering
		\includegraphics[width=0.98\textwidth]{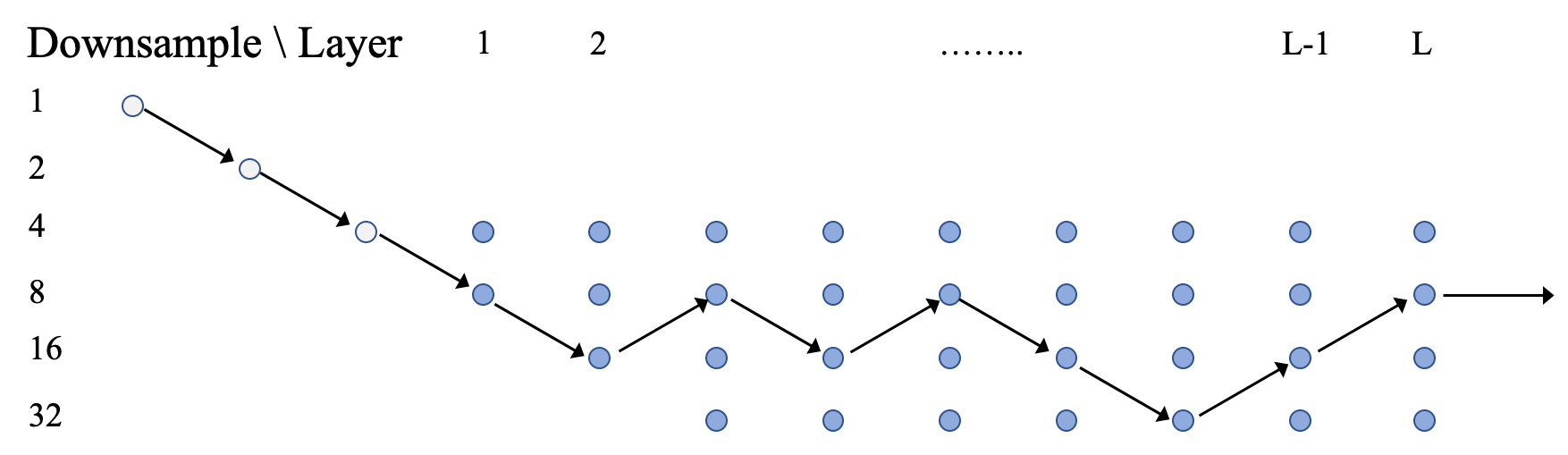}
	\end{subfigure}
	\caption{Spatial-resolution-change path of the \ourmethod{}-C0 architecture. Each circle encodes a residual module \cite{resnet}. Circles colored in white are always executed at the beginning. The arrows and blues circles are parts of the search decisions. The total number of layers (L) are set to be nine.
	\label{fig:nsganet_cmucars}}
\end{figure}

We use the CMU-Car dataset originally introduced in \cite{naresh2013correlation}. The dataset contains around 10,000 car images in different orientations, environments, and occlusion situations. In this case, we search for the path of image resolution changes, similar to \cite{liu2019auto}. The node-level structure is kept fixed, using the basic residual unit \cite{resnet}. The performance of architectures in this case is calculated using the root mean square (RMS) error between the predicted heatmap and ground truth for each key-point, more details are available in \cite{naresh2013correlation}. We use FLOPs as the second objective for architecture complexity measurement. The obtained architectures are named as \ourmethod{}-C0 and -C1. The obtained results are provided in Table~\ref{tab:nsganet_cmucars} and the visualization of the architectures is provided in Fig.~\ref{fig:nsganet_cmucars}.

\begin{table}[!htbp]
\caption{Preliminary results on the CMU-Car alignment \cite{naresh2013correlation}. Notably, our proposed algorithm is able to find architectures with competitive performance while having 2x less parameters when compared to human-designed architecture \cite{newell2016stacked}. \label{tab:nsganet_cmucars}}
\centering
\resizebox{0.38\textwidth}{!}{%
\begin{tabular}{lccc}
    \toprule
    \multicolumn{1}{c}{Architectures} & \multicolumn{1}{c}{Params.} & \multicolumn{1}{c}{FLOPs} & \multicolumn{1}{c}{Regression} \\
     & \multicolumn{1}{c}{(M)} & \multicolumn{1}{c}{(M)} & \multicolumn{1}{c}{Error (\%)} \\\midrule
    Hourglass\cite{newell2016stacked} & 3.38 & 3613 & 7.80 \\
    \ourmethod{}-C0 & 1.53 & 2584 & 8.66 \\
    \ourmethod{}-C1 & 1.61 & 2663 & 8.64 \\
    \bottomrule
    \end{tabular}
    }
\end{table}

{
\subsubsection{Ablation Study on Exploitation Operator}
{
Recall that we use Bayesian Network (BN) as a probabilistic model to estimate the distribution of the Pareto set (of architectures). In this section, we first explain the connection of our proposed BN-based distribution estimation operator to the existing works \cite{5415586,7929324,8533425} of large-scale multi-objective optimization algorithms for general numerical problems. The common theme behind these works is to learn the correlation among decision variables to reduce the dimension either through grouping (optimize a subset of variables at a time) or embedding (projection to lower-dimensional space). In our work, we exploit the problem information (i.e., network architectures are variants of directed acyclic graphs) explicitly in the form of a BN to learn the correlations (i.e., BN edge weights) among architectural variables. The learned BN is then used (as a probabilistic model) to generate the remaining variables given the observed variables, as a form of dimension reduction. Thus, in this work, we take advantage of learning algorithms to capture the properties of good solutions to deal with the large dimensionality of the problem. In short, our approach in NAS application and general-purpose EMO algorithms shares a similar concept of dimensional reduction to handle large-scale problems.}

Secondly, we study the effectiveness of the proposed BN-based exploitation operator. The experimental setup follows a two objective NAS optimization to maximize top-1 validation accuracy on the FashionMNIST dataset \cite{xiao2017fashion} and minimize \#FLOPs simultaneously. We study five different settings of the proposed exploitation operator, namely:
\begin{enumerate}
    \item No exploitation
    \item Exploitation activate after 1/3 computation budget spent. 
    \item Exploitation activate after 1/2 computation budget spent. 
    \item Exploitation activate after 2/3 computation budget spent. 
    \item Exploitation activate after 3/4 computation budget spent.
\end{enumerate}
We use the same population size of 40 and a maximum number of 30 generations for each of the considered settings. And we repeat 11 runs with different random seeds to capture the variance from different initial population. The obtained results are provided in Fig.~\ref{fig:ablation_exploitation}.

\begin{figure}[hbt]
    \centering
    \includegraphics[width=0.48\textwidth]{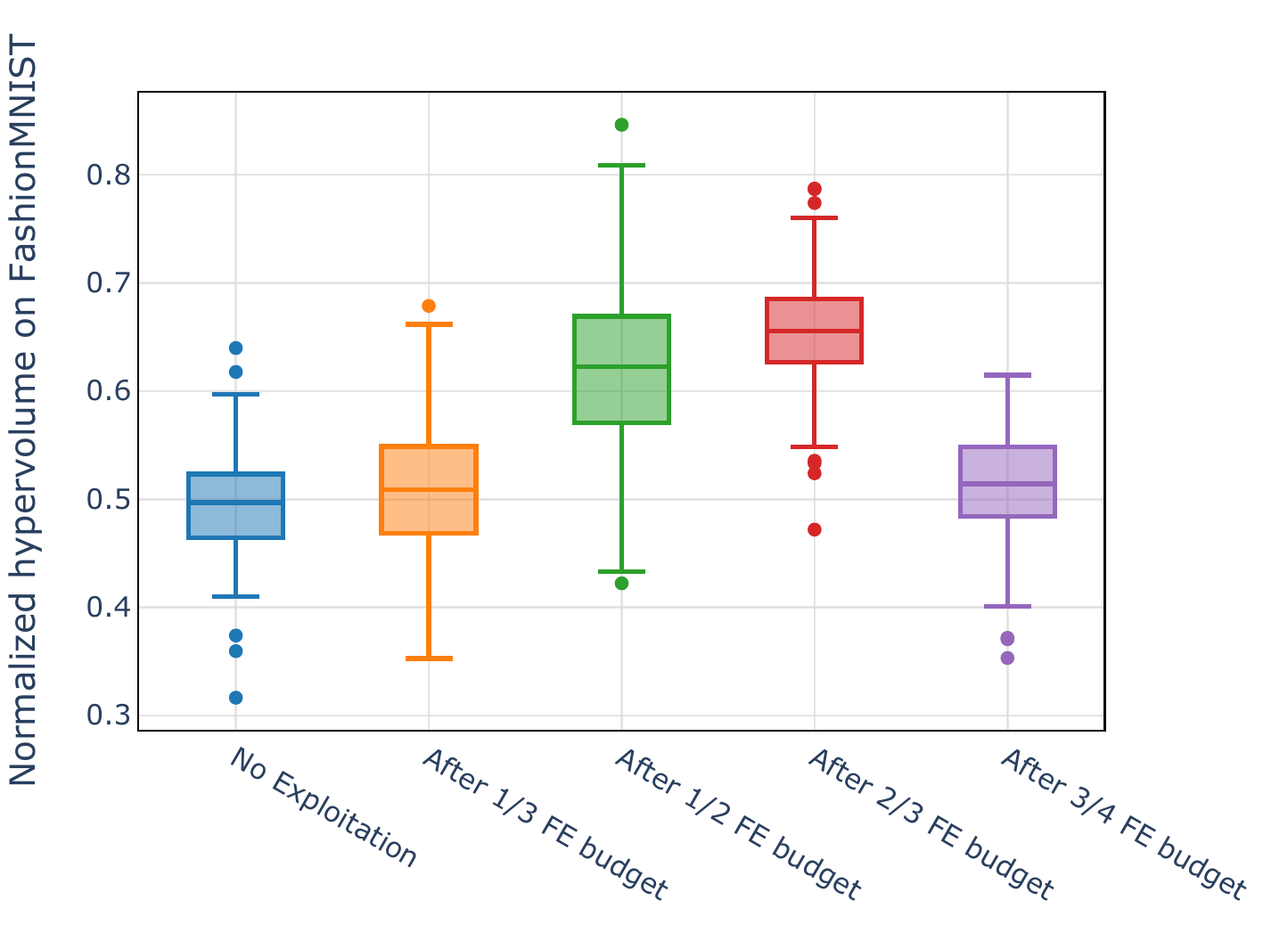}
    \caption{Ablation study on effectiveness of our proposed exploitation operator under different settings.\label{fig:ablation_exploitation}}
\end{figure}

Empirically, we observe that our proposed exploitation operator provides a noticeable improvement to the overall algorithm's performance, measured by hypervolume. However, the margin of improvement quickly diminishes as we activate the operator too early (i.e. before 1/3 of the computation budget spent) or too close to the total budget (i.e. after 3/4 of the computation budget spent). 
}

{
\subsection{Hypervolume Calculation\label{sec:supple_hypervolume}}
For the two-objective experiments presented in Section~IV of the main paper, the reference point used in computing the hypervolume metric is $(100, 1,000)$, where $100$ is the worst error rate in percentage, and $1,000$ is the highest \#FLOPs, in millions, of any architecture that our search space can encode. We then normalize the hypervolume by the rectangular area formed by the reference point and the ideal point---i.e. $(0,0)$. 
}

%% file: arxiv.bbl
\begin{thebibliography}{10}
\providecommand{\url}[1]{#1}
\csname url@samestyle\endcsname
\providecommand{\newblock}{\relax}
\providecommand{\bibinfo}[2]{#2}
\providecommand{\BIBentrySTDinterwordspacing}{\spaceskip=0pt\relax}
\providecommand{\BIBentryALTinterwordstretchfactor}{4}
\providecommand{\BIBentryALTinterwordspacing}{\spaceskip=\fontdimen2\font plus
\BIBentryALTinterwordstretchfactor\fontdimen3\font minus
  \fontdimen4\font\relax}
\providecommand{\BIBforeignlanguage}[2]{{%
\expandafter\ifx\csname l@#1\endcsname\relax
\typeout{** WARNING: IEEEtran.bst: No hyphenation pattern has been}%
\typeout{** loaded for the language `#1'. Using the pattern for}%
\typeout{** the default language instead.}%
\else
\language=\csname l@#1\endcsname
\fi
#2}}
\providecommand{\BIBdecl}{\relax}
\BIBdecl

\bibitem{googlenet}
C.~Szegedy, W.~Liu, Y.~Jia, P.~Sermanet, S.~Reed, D.~Anguelov, D.~Erhan,
  V.~Vanhoucke, and A.~Rabinovich, ``Going deeper with convolutions,'' in
  \emph{IEEE Conference on Computer Vision and Pattern Recognition (CVPR)},
  2015.

\bibitem{resnet}
K.~He, X.~Zhang, S.~Ren, and J.~Sun, ``Deep residual learning for image
  recognition,'' in \emph{IEEE Conference on Computer Vision and Pattern
  Recognition (CVPR)}, 2016.

\bibitem{densenet}
G.~Huang, Z.~Liu, L.~Van Der~Maaten, and K.~Q. Weinberger, ``Densely connected
  convolutional networks,'' in \emph{IEEE Conference on Computer Vision and
  Pattern Recognition (CVPR)}, 2017.

\bibitem{zhang2018shufflenet}
X.~Zhang, X.~Zhou, M.~Lin, and J.~Sun, ``Shufflenet: An extremely efficient
  convolutional neural network for mobile devices,'' in \emph{IEEE Conference
  on Computer Vision and Pattern Recognition (CVPR)}, 2018.

\bibitem{sandler2018mobilenetv2}
M.~Sandler, A.~Howard, M.~Zhu, A.~Zhmoginov, and L.-C. Chen, ``{MobileNetV2}:
  Inverted residuals and linear bottlenecks,'' in \emph{IEEE Conference on
  Computer Vision and Pattern Recognition (CVPR)}, 2018.

\bibitem{lbcnn}
F.~Juefei-Xu, V.~N. Boddeti, and M.~Savvides, ``Local binary convolutional
  neural networks,'' in \emph{IEEE Conference on Computer Vision and Pattern
  Recognition (CVPR)}, 2017.

\bibitem{zoph2016}
B.~Zoph and Q.~V. Le, ``{Neural Architecture Search with Reinforcement
  Learning},'' in \emph{International Conference on Learning Representations
  (ICLR)}, 2016.

\bibitem{nasnet2018}
B.~Zoph, V.~Vasudevan, J.~Shlens, and Q.~V. Le, ``Learning transferable
  architectures for scalable image recognition,'' in \emph{IEEE Conference on
  Computer Vision and Pattern Recognition (CVPR)}, 2018.

\bibitem{liu2018darts}
H.~Liu, K.~Simonyan, and Y.~Yang, ``{DARTS}: Differentiable architecture
  search,'' in \emph{International Conference on Learning Representations
  (ICLR)}, 2019.

\bibitem{xie2018snas}
S.~Xie, H.~Zheng, C.~Liu, and L.~Lin, ``{SNAS}: stochastic neural architecture
  search,'' in \emph{International Conference on Learning Representations
  (ICLR)}, 2019.

\bibitem{Dong_2019_CVPR}
X.~Dong and Y.~Yang, ``Searching for a robust neural architecture in four {GPU}
  hours,'' in \emph{IEEE Conference on Computer Vision and Pattern Recognition
  (CVPR)}, 2019.

\bibitem{wu2019fbnet}
B.~Wu, X.~Dai, P.~Zhang, Y.~Wang, F.~Sun, Y.~Wu, Y.~Tian, P.~Vajda, Y.~Jia, and
  K.~Keutzer, ``{FBNet}: Hardware-aware efficient convnet design via
  differentiable neural architecture search,'' in \emph{IEEE Conference on
  Computer Vision and Pattern Recognition (CVPR)}, 2019.

\bibitem{tan2019efficientnet}
M.~Tan and Q.~V. Le, ``Efficientnet: Rethinking model scaling for convolutional
  neural networks,'' in \emph{International Conference on Machine Learning
  (ICML)}, 2019.

\bibitem{genetic-cnn}
L.~{Xie} and A.~{Yuille}, ``Genetic {CNN},'' in \emph{International Conference
  on Computer Vision (ICCV)}, 2017.

\bibitem{real2017largescale}
E.~Real, S.~Moore, A.~Selle, S.~Saxena, Y.~L. Suematsu, J.~Tan, Q.~V. Le, and
  A.~Kurakin, ``Large-scale evolution of image classifiers,'' in
  \emph{International Conference on Machine Learning (ICML)}, 2017.

\bibitem{liu2018hierarchical}
H.~Liu, K.~Simonyan, O.~Vinyals, C.~Fernando, and K.~Kavukcuoglu,
  ``Hierarchical representations for efficient architecture search,'' in
  \emph{International Conference on Learning Representations (ICLR)}, 2018.

\bibitem{real2019regularized}
E.~Real, A.~Aggarwal, Y.~Huang, and Q.~V. Le, ``Regularized evolution for image
  classifier architecture search,'' in \emph{AAAI Conference on Artificial
  Intelligence}, 2019.

\bibitem{ae-cnn}
Y.~{Sun}, B.~{Xue}, M.~{Zhang}, and G.~G. {Yen}, ``Completely automated {CNN}
  architecture design based on blocks,'' \emph{IEEE Transactions on Neural
  Networks and Learning Systems}, vol.~31, no.~4, pp. 1242--1254, 2020.

\bibitem{ae-cnn-e2epp}
Y.~{Sun}, H.~{Wang}, B.~{Xue}, Y.~{Jin}, G.~G. {Yen}, and M.~{Zhang},
  ``Surrogate-assisted evolutionary deep learning using an end-to-end random
  forest-based performance predictor,'' \emph{IEEE Transactions on Evolutionary
  Computation}, vol.~24, no.~2, pp. 350--364, 2020.

\bibitem{lu2019nsga}
Z.~Lu, I.~Whalen, V.~Boddeti, Y.~Dhebar, K.~Deb, E.~Goodman, and W.~Banzhaf,
  ``{NSGA-Net}: Neural architecture search using multi-objective genetic
  algorithm,'' in \emph{Genetic and Evolutionary Computation Conference
  (GECCO)}, 2019.

\bibitem{yao1999evolving}
\color{black}X. {Yao}, ``Evolving artificial neural networks,''
  \emph{Proceedings of the IEEE}, vol.~87, no.~9, pp. 1423--1447,
  1999\color{black}.

\bibitem{stanley2002evolving}
K.~O. Stanley and R.~Miikkulainen, ``Evolving neural networks through
  augmenting topologies,'' \emph{Evolutionary Computation}, vol.~10, no.~2, pp.
  99--127, 2002.

\bibitem{alphs}
G.~S. Hornby, ``{ALPS}: the age-layered population structure for reducing the
  problem of premature convergence,'' in \emph{Genetic and Evolutionary
  Computation Conference (GECCO)}, 2006.

\bibitem{cgp-cnn}
M.~Suganuma, S.~Shirakawa, and T.~Nagao, ``A genetic programming approach to
  designing convolutional neural network architectures,'' in \emph{Genetic and
  Evolutionary Computation Conference (GECCO)}, 2017.

\bibitem{cnn-ga}
\color{black}Y. {Sun}, B.~{Xue}, M.~{Zhang}, G.~G. {Yen}, and J.~{Lv},
  ``Automatically designing {CNN} architectures using the genetic algorithm for
  image classification,'' \emph{IEEE Transactions on Cybernetics}, pp. 1--15,
  \color{black}2020.

\bibitem{jin2008pareto}
\color{black}Y. {Jin} and B.~Sendhoff, ``Pareto-based multiobjective machine
  learning: An overview and case studies,'' \emph{IEEE Transactions on Systems,
  Man, and Cybernetics, Part C (Applications and Reviews)}, vol.~38, no.~3, pp.
  397--415, 2008\color{black}.

\bibitem{zhu2019multi}
H.~{Zhu} and Y.~{Jin}, ``Multi-objective evolutionary federated learning,''
  \emph{IEEE Transactions on Neural Networks and Learning Systems}, vol.~31,
  no.~4, pp. 1310--1322, 2020.

\bibitem{kim2017nemo}
Y.-H. Kim, B.~Reddy, S.~Yun, and C.~Seo, ``{NEMO}: Neuro-evolution with
  multiobjective optimization of deep neural network for speed and accuracy,''
  in \emph{International Conference on Machine Learning (ICML) AutoML
  Workshop}, 2017.

\bibitem{deb2002fast}
K.~Deb, A.~Pratap, S.~Agarwal, and T.~Meyarivan, ``A fast and elitist
  multiobjective genetic algorithm: {NSGA-II},'' \emph{IEEE Transactions on
  Evolutionary Computation}, vol.~6, no.~2, pp. 182--197, 2002.

\bibitem{dong2018dpp}
J.-D. Dong, A.-C. Cheng, D.-C. Juan, W.~Wei, and M.~Sun, ``{DPP-Net}:
  Device-aware progressive search for pareto-optimal neural architectures,'' in
  \emph{European Conference on Computer Vision (ECCV)}, 2018.

\bibitem{liu2018progressive}
C.~Liu, B.~Zoph, M.~Neumann, J.~Shlens, W.~Hua, L.-J. Li, L.~Fei-Fei,
  A.~Yuille, J.~Huang, and K.~Murphy, ``Progressive neural architecture
  search,'' in \emph{European Conference on Computer Vision (ECCV)}, 2018.

\bibitem{elsken2018efficient}
T.~Elsken, J.~H. Metzen, and F.~Hutter, ``Efficient multi-objective neural
  architecture search via lamarckian evolution,'' in \emph{International
  Conference on Learning Representations (ICLR)}, 2019.

\bibitem{wei2016morhpisms}
T.~Wei, C.~Wang, Y.~Rui, and C.~W. Chen, ``Network morphism,'' in
  \emph{International Conference on Machine Learning (ICML)}, 2016.

\bibitem{cifar10}
A.~Krizhevsky, G.~Hinton \emph{et~al.}, ``Learning multiple layers of features
  from tiny images,'' Citeseer, Tech. Rep., 2009.

\bibitem{zhong2017blockqnn}
Z.~Zhong, J.~Yan, W.~Wu, J.~Shao, and C.-L. Liu, ``Practical block-wise neural
  network architecture generation,'' in \emph{IEEE Conference on Computer
  Vision and Pattern Recognition (CVPR)}, 2018.

\bibitem{eichfelder2010multiobjective}
G.~Eichfelder, ``Multiobjective bilevel optimization,'' \emph{Mathematical
  Programming}, vol. 123, no.~2, pp. 419--449, 2010.

\bibitem{hu2018squeeze}
J.~Hu, L.~Shen, and G.~Sun, ``Squeeze-and-excitation networks,'' in \emph{IEEE
  Conference on Computer Vision and Pattern Recognition (CVPR)}, 2018.

\bibitem{leung1997degree}
Y.~Leung, Y.~Gao, and Z.-B. Xu, ``Degree of population diversity-a perspective
  on premature convergence in genetic algorithms and its markov chain
  analysis,'' \emph{IEEE Transactions on Neural Networks}, vol.~8, no.~5, pp.
  1165--1176, 1997.

\bibitem{miller1995genetic}
B.~L. Miller, D.~E. Goldberg \emph{et~al.}, ``Genetic algorithms, tournament
  selection, and the effects of noise,'' \emph{Complex systems}, vol.~9, no.~3,
  pp. 193--212, 1995.

\bibitem{deb1995simulated}
K.~Deb and R.~B. Agrawal, ``Simulated binary crossover for continuous search
  space,'' \emph{Complex systems}, vol.~9, no.~2, pp. 115--148, 1995.

\bibitem{pelikan1999boa}
M.~Pelikan, D.~E. Goldberg, and E.~Cant{\'u}-Paz, ``{BOA}: The bayesian
  optimization algorithm,'' in \emph{Genetic and Evolutionary Computation
  Conference (GECCO)}, 1999.

\bibitem{xie2017aggregated}
S.~Xie, R.~Girshick, P.~Doll{\'a}r, Z.~Tu, and K.~He, ``Aggregated residual
  transformations for deep neural networks,'' in \emph{IEEE Conference on
  Computer Vision and Pattern Recognition (CVPR)}, 2017.

\bibitem{cai2018proxylessnas}
H.~Cai, L.~Zhu, and S.~Han, ``Proxyless{NAS}: Direct neural architecture search
  on target task and hardware,'' in \emph{International Conference on Learning
  Representations (ICLR)}, 2019.

\bibitem{pmlr-v80-pham18a}
H.~Pham, M.~Guan, B.~Zoph, Q.~Le, and J.~Dean, ``Efficient neural architecture
  search via parameters sharing,'' in \emph{International Conference on Machine
  Learning (ICML)}, 2018.

\bibitem{li2019random}
L.~Li and A.~Talwalkar, ``Random search and reproducibility for neural
  architecture search,'' \emph{arXiv preprint arXiv:1902.07638}, 2019.

\bibitem{xie2019exploring}
S.~Xie, A.~Kirillov, R.~Girshick, and K.~He, ``Exploring randomly wired neural
  networks for image recognition,'' in \emph{International Conference on
  Computer Vision (ICCV)}, 2019, pp. 1284--1293.

\bibitem{cutout}
T.~DeVries and G.~W. Taylor, ``Improved regularization of convolutional neural
  networks with cutout,'' \emph{arXiv preprint arXiv:1708.04552}, 2017.

\bibitem{loshchilov2016sgdr}
I.~Loshchilov and F.~Hutter, ``Sgdr: Stochastic gradient descent with warm
  restarts,'' in \emph{International Conference on Learning Representations
  (ICLR)}, 2017.

\bibitem{baker2017metaqnn}
B.~Baker, O.~Gupta, N.~Naik, and R.~Raskar, ``Designing neural network
  architectures using reinforcement learning,'' in \emph{International
  Conference on Learning Representations (ICLR)}, 2017.

\bibitem{deb2006innovization}
K.~Deb and A.~Srinivasan, ``Innovization: Innovating design principles through
  optimization,'' in \emph{Genetic and Evolutionary Computation Conference
  (GECCO)}, 2006.

\bibitem{recht2019imagenet}
B.~Recht, R.~Roelofs, L.~Schmidt, and V.~Shankar, ``Do imagenet classifiers
  generalize to imagenet?'' \emph{arXiv preprint arXiv:1902.10811}, 2019.

\bibitem{hendrycks2018benchmarking}
D.~Hendrycks and T.~Dietterich, ``Benchmarking neural network robustness to
  common corruptions and perturbations,'' in \emph{International Conference on
  Learning Representations (ICLR)}, 2019.

\bibitem{goodfellow2014explaining}
I.~J. Goodfellow, J.~Shlens, and C.~Szegedy, ``Explaining and harnessing
  adversarial examples,'' \emph{arXiv preprint arXiv:1412.6572}, 2014.

\bibitem{netzer2011reading}
Y.~Netzer, T.~Wang, A.~Coates, A.~Bissacco, B.~Wu, and A.~Y. Ng, ``Reading
  digits in natural images with unsupervised feature learning,'' in
  \emph{Advances in Neural Information Processing Systems (NeurIPS) Workshop on
  Deep Learning and Unsupervised Feature Learning}, 2011.

\bibitem{xiao2017fashion}
H.~Xiao, K.~Rasul, and R.~Vollgraf, ``Fashion-{MNIST}: A novel image dataset
  for benchmarking machine learning algorithms,'' \emph{arXiv preprint
  arXiv:1708.07747}, 2017.

\bibitem{wang2017chestx}
X.~Wang, Y.~Peng, L.~Lu, Z.~Lu, M.~Bagheri, and R.~M. Summers, ``{ChestX-ray8}:
  Hospital-scale chest x-ray database and benchmarks on weakly-supervised
  classification and localization of common thorax diseases,'' in \emph{IEEE
  Conference on Computer Vision and Pattern Recognition (CVPR)}, 2017.

\bibitem{yao2017learning}
L.~Yao, E.~Poblenz, D.~Dagunts, B.~Covington, D.~Bernard, and K.~Lyman,
  ``Learning to diagnose from scratch by exploiting dependencies among
  labels,'' \emph{arXiv preprint arXiv:1710.10501}, 2017.

\bibitem{rajpurkar2017chexnet}
P.~Rajpurkar, J.~Irvin, K.~Zhu, B.~Yang, H.~Mehta, T.~Duan, D.~Ding, A.~Bagul,
  C.~Langlotz, K.~Shpanskaya \emph{et~al.}, ``{CheXNet}: Radiologist-level
  pneumonia detection on chest x-rays with deep learning,'' \emph{arXiv
  preprint arXiv:1711.05225}, 2017.

\bibitem{blog2017automl}
G.~R. Blog, ``Automl for large scale image classification and object
  detection,'' \emph{Google Research, https://research. googleblog.
  com/2017/11/automl-for-large-scaleimage. html, Blog}, 2017.

\bibitem{Liang-automl}
J.~Liang, E.~Meyerson, B.~Hodjat, D.~Fink, K.~Mutch, and R.~Miikkulainen,
  ``Evolutionary neural automl for deep learning,'' in \emph{Genetic and
  Evolutionary Computation Conference (GECCO)}, 2019.

\bibitem{zhou2016learning}
B.~Zhou, A.~Khosla, A.~Lapedriza, A.~Oliva, and A.~Torralba, ``Learning deep
  features for discriminative localization,'' in \emph{IEEE Conference on
  Computer Vision and Pattern Recognition (CVPR)}, 2016.

\bibitem{watkins1989qlearning}
\BIBentryALTinterwordspacing
C.~J. C.~H. Watkins, ``Learning from delayed rewards,'' Ph.D. dissertation,
  King's College, Cambridge, UK, May 1989. [Online]. Available:
  \url{http://www.cs.rhul.ac.uk/~chrisw/new_thesis.pdf}
\BIBentrySTDinterwordspacing

\bibitem{ptb}
M.~Marcus, G.~Kim, M.~A. Marcinkiewicz, R.~MacIntyre, A.~Bies, M.~Ferguson,
  K.~Katz, and B.~Schasberger, ``The penn treebank: annotating predicate
  argument structure,'' in \emph{Proceedings of the workshop on Human Language
  Technology}.\hskip 1em plus 0.5em minus 0.4em\relax Association for
  Computational Linguistics, 1994, pp. 114--119.

\bibitem{liu2019auto}
C.~Liu, L.-C. Chen, F.~Schroff, H.~Adam, W.~Hua, A.~L. Yuille, and L.~Fei-Fei,
  ``Auto-deeplab: Hierarchical neural architecture search for semantic image
  segmentation,'' in \emph{IEEE Conference on Computer Vision and Pattern
  Recognition (CVPR)}, 2019.

\bibitem{dong2018ppp-net}
J.-D. Dong, A.-C. Cheng, D.-C. Juan, W.~Wei, and M.~Sun, ``{PPP-Net}:
  Platform-aware progressive search for pareto-optimal neural architectures,''
  in \emph{International Conference on Learning Representations (ICLR)}, 2018.

\bibitem{kandasamy2018neural}
K.~Kandasamy, W.~Neiswanger, J.~Schneider, B.~Poczos, and E.~P. Xing, ``Neural
  architecture search with bayesian optimisation and optimal transport,'' in
  \emph{Advances in Neural Information Processing Systems (NeurIPS)}, 2018.

\bibitem{mnasnet}
M.~Tan, B.~Chen, R.~Pang, V.~Vasudevan, M.~Sandler, A.~Howard, and Q.~V. Le,
  ``Mnasnet: Platform-aware neural architecture search for mobile,'' in
  \emph{IEEE Conference on Computer Vision and Pattern Recognition (CVPR)},
  2019.

\bibitem{mobilenetv3}
A.~Howard, M.~Sandler, G.~Chu, L.-C. Chen, B.~Chen, M.~Tan, W.~Wang, Y.~Zhu,
  R.~Pang, V.~Vasudevan, Q.~V. Le, and H.~Adam, ``Searching for mobilenetv3,''
  in \emph{International Conference on Computer Vision (ICCV)}, 2019.

\bibitem{Yu2020Evaluating}
K.~Yu, C.~Sciuto, M.~Jaggi, C.~Musat, and M.~Salzmann, ``Evaluating the search
  phase of neural architecture search,'' in \emph{International Conference on
  Learning Representations (ICLR)}, 2020.

\bibitem{brock2018smash}
A.~Brock, T.~Lim, J.~Ritchie, and N.~Weston, ``{SMASH}: One-shot model
  architecture search through hypernetworks,'' in \emph{International
  Conference on Learning Representations (ICLR)}, 2018.

\bibitem{one-shot}
G.~Bender, P.-J. Kindermans, B.~Zoph, V.~Vasudevan, and Q.~Le, ``Understanding
  and simplifying one-shot architecture search,'' in \emph{International
  Conference on Machine Learning (ICML)}, 2018.

\bibitem{bilevel-cma}
X.~{He}, Y.~{Zhou}, and Z.~{Chen}, ``Evolutionary bilevel optimization based on
  covariance matrix adaptation,'' \emph{IEEE Transactions on Evolutionary
  Computation}, vol.~23, no.~2, pp. 258--272, 2019.

\bibitem{bleaq}
\BIBentryALTinterwordspacing
A.~Sinha, P.~Malo, and K.~Deb, ``Evolutionary algorithm for bilevel
  optimization using approximations of the lower level optimal solution
  mapping,'' \emph{European Journal of Operational Research}, vol. 257, no.~2,
  pp. 395 -- 411, 2017. [Online]. Available:
  \url{http://www.sciencedirect.com/science/article/pii/S0377221716306634}
\BIBentrySTDinterwordspacing

\bibitem{carlini2017towards}
N.~Carlini and D.~Wagner, ``Towards evaluating the robustness of neural
  networks,'' in \emph{IEEE Symposium on Security and Privacy (SP)}, 2017.

\bibitem{vgg}
K.~Simonyan and A.~Zisserman, ``{Very Deep Convolutional Networks for
  Large-scale Image Recognition},'' in \emph{International Conference on
  Learning Representations (ICLR)}, 2015.

\bibitem{naresh2013correlation}
V.~Naresh~Boddeti, T.~Kanade, and B.~Vijaya~Kumar, ``Correlation filters for
  object alignment,'' in \emph{IEEE Conference on Computer Vision and Pattern
  Recognition (CVPR)}, 2013.

\bibitem{newell2016stacked}
A.~Newell, K.~Yang, and J.~Deng, ``Stacked hourglass networks for human pose
  estimation,'' in \emph{European Conference on Computer Vision (ECCV)}, 2016.

\bibitem{5415586}
J.~J. {Durillo}, A.~J. {Nebro}, C.~A.~C. {Coello}, J.~{Garcia-Nieto},
  F.~{Luna}, and E.~{Alba}, ``A study of multiobjective metaheuristics when
  solving parameter scalable problems,'' \emph{IEEE Transactions on
  Evolutionary Computation}, vol.~14, no.~4, pp. 618--635, 2010\color{black}.

\bibitem{7929324}
H.~{Zille}, H.~{Ishibuchi}, S.~{Mostaghim}, and Y.~{Nojima}, ``A framework for
  large-scale multiobjective optimization based on problem transformation,''
  \emph{IEEE Transactions on Evolutionary Computation}, vol.~22, no.~2, pp.
  260--275, 2018\color{black}.

\bibitem{8533425}
W.~{Hong}, K.~{Tang}, A.~{Zhou}, H.~{Ishibuchi}, and X.~{Yao}, ``A scalable
  indicator-based evolutionary algorithm for large-scale multiobjective
  optimization,'' \emph{IEEE Transactions on Evolutionary Computation},
  vol.~23, no.~3, pp. 525--537, 2019\color{black}.

\end{thebibliography}
